%% file: main.tex
\documentclass[10pt,twocolumn,letterpaper]{article}

\usepackage[pagenumbers]{preprint}      %

\input{preamble}
\definecolor{linkblue}{rgb}{0.21,0.49,0.74}
\usepackage{bibunits}  %
\usepackage[pagebackref,breaklinks,colorlinks,allcolors=linkblue,plainpages=false]{hyperref}

\title{\paper: Compact Vision-Language Navigation in Continuous Environments\\via On-Policy Training and Route-Bearing Output}

\author{{\fontsize{11}{13}\selectfont Reshad Ul Karim$^{1}$\quad Syeda Maliha Tabassum$^{1}$\quad Md.\ Khalilur Rhaman$^{1}$\quad Md\ Rashad Al Hasan Rony$^{2}$}\\
{\normalsize $^{1}$BRAC University, Dhaka, Bangladesh\qquad $^{2}$AI Competence Center, Deutsche Telekom AG}\\
{\tt\footnotesize \{reshad.ul.karim, syeda.maliha.tabassum1\}@g.bracu.ac.bd, khalilur@bracu.ac.bd, rashad.research@gmail.com}
}

\begin{document}
\maketitle
\input{sec/0_abstract}
\input{sec/1_intro}
\input{sec/2_related}

\input{sec/3_method}
\input{sec/4_experiments}
\input{sec/5_conclusion}
\clearpage
\input{sec/llm_usage}
{
    \small
    \bibliographystyle{ieeenat_fullname}
    \bibliography{main}
}

\begin{bibunit}[ieeenat_fullname]
\makeatletter
\let\Hy@backout\@gobble
\def\hyper@natlinkstart#1{\hyper@linkstart{cite}{cite.supp.#1}\def\hyper@nat@current{supp.#1}}
\def\hyper@natlinkbreak#1#2{\hyper@linkend#1\hyper@linkstart{cite}{cite.supp.#2}}
\def\hyper@natanchorstart#1{\Hy@raisedlink{\hyper@anchorstart{cite.supp.#1}}}
\def\backref#1{}\def\backrefalt#1#2#3#4{}
\makeatother
\input{sec/X_suppl}
{\small\putbib[main]}
\end{bibunit}

\end{document}

%% file: preamble.tex
\usepackage{soul}
\usepackage{multirow}
\usepackage{array}
\usepackage{tikz}
\usetikzlibrary{arrows.meta,positioning,calc,fit,backgrounds,shapes.geometric,decorations.pathreplacing,patterns}
\usepackage{bm}
\usepackage{microtype}
\DeclareMathSizes{8.6}{8.6}{7}{5}
\DeclareMathSizes{8.9}{8.9}{7}{5}
\DeclareMathSizes{9.8}{9.8}{7}{5}
\makeatletter
\def\ftype@qfig{4}\def\fps@qfig{tbp}\def\ext@qfig{lof}
\newenvironment{qfigure*}[1][tbp]{\@dblfloat{qfig}[#1]\def\@captype{figure}}{\end@dblfloat}
\makeatother

\newif\ifhighlights
\highlightstrue

\definecolor{pal-text}{HTML}{111111}
\definecolor{pal-text-muted}{HTML}{111111}
\definecolor{pal-axis}{HTML}{111111}
\definecolor{pal-legend-edge}{HTML}{111111}
\definecolor{pal-white}{HTML}{FFFFFF}
\definecolor{pal-plot-face}{HTML}{EEEFF6}
\definecolor{pal-grid}{HTML}{B2ABD2}
\definecolor{pal-band}{HTML}{FEECD3}
\definecolor{pal-row-stripe}{HTML}{F5F5F5}
\definecolor{pal-placeholder}{HTML}{F7F7F7}
\definecolor{pal-card-face}{HTML}{F7F7F7}
\definecolor{pal-card-edge}{HTML}{B2ABD2}
\definecolor{pal-instruction-card}{HTML}{FEE0B6}
\definecolor{pal-instruction-card-edge}{HTML}{FDB863}
\definecolor{pal-split-chip}{HTML}{F7F7F7}
\definecolor{pal-badge}{HTML}{542788}
\definecolor{pal-chip}{HTML}{542788}
\definecolor{pal-ours}{HTML}{542788}
\definecolor{pal-ours-secondary}{HTML}{8073AC}
\definecolor{pal-ours-light}{HTML}{B2ABD2}
\definecolor{pal-ours-fill}{HTML}{D8DAEB}
\definecolor{pal-baseline}{HTML}{E08214}
\definecolor{pal-baseline-light}{HTML}{FDB863}
\definecolor{pal-baseline-fill}{HTML}{FEE0B6}
\definecolor{pal-gain-text}{HTML}{542788}
\definecolor{pal-gain-fill}{HTML}{B2ABD2}
\definecolor{pal-loss-text}{HTML}{B35806}
\definecolor{pal-loss-fill}{HTML}{FDB863}
\definecolor{pal-near-zero}{HTML}{F7F7F7}
\definecolor{pal-route-bearing}{HTML}{542788}
\definecolor{pal-image-point}{HTML}{E08214}
\definecolor{pal-orient-then-go}{HTML}{542788}
\definecolor{pal-previous-schedule}{HTML}{E08214}
\definecolor{pal-expert}{HTML}{E08214}
\definecolor{pal-truncated-memory}{HTML}{636363}
\definecolor{pal-imitation-only}{HTML}{969696}
\definecolor{pal-third-variant}{HTML}{636363}
\definecolor{pal-tick}{HTML}{542788}
\definecolor{pal-cross}{HTML}{B35806}
\definecolor{pal-neutral-dark}{HTML}{636363}
\definecolor{pal-neutral-mid}{HTML}{969696}
\definecolor{pal-neutral-light}{HTML}{D9D9D9}
\definecolor{pal-not-applicable}{HTML}{BDBDBD}
\definecolor{pal-connector}{HTML}{BDBDBD}
\definecolor{pal-fov-sector}{HTML}{F0F0F0}
\definecolor{pal-polar-grid}{HTML}{D9D9D9}
\definecolor{pal-seed-1}{HTML}{B2ABD2}
\definecolor{pal-seed-2}{HTML}{8073AC}
\definecolor{pal-seed-3}{HTML}{542788}
\definecolor{pal-published-rgb}{HTML}{8073AC}
\definecolor{pal-published-depth}{HTML}{E08214}
\definecolor{pal-published}{HTML}{8073AC}
\definecolor{pal-teaser-seed}{HTML}{8073AC}
\definecolor{pal-gpu-rtx4090}{HTML}{E08214}
\definecolor{pal-gpu-a100}{HTML}{8073AC}
\definecolor{pal-gpu-h200}{HTML}{B35806}
\definecolor{pal-gpu-not-stated}{HTML}{B2ABD2}
\definecolor{pal-gpu-not-stated-edge}{HTML}{542788}
\definecolor{pal-gpu-ours}{HTML}{542788}
\definecolor{pal-camera-control}{HTML}{BDBDBD}
\definecolor{pal-camera-two-view}{HTML}{737373}
\definecolor{pal-camera-unrestricted}{HTML}{B2ABD2}
\definecolor{pal-camera-support-restricted}{HTML}{542788}
\definecolor{pal-prob-ramp-1}{HTML}{F7F7F7}
\definecolor{pal-prob-ramp-2}{HTML}{D8DAEB}
\definecolor{pal-prob-ramp-3}{HTML}{B2ABD2}
\definecolor{pal-prob-ramp-4}{HTML}{8073AC}
\definecolor{pal-prob-ramp-5}{HTML}{542788}
\definecolor{pal-heat-cmap-1}{HTML}{000004}
\definecolor{pal-heat-cmap-2}{HTML}{51127C}
\definecolor{pal-heat-cmap-3}{HTML}{B73779}
\definecolor{pal-heat-cmap-4}{HTML}{FC8961}
\definecolor{pal-heat-cmap-5}{HTML}{FCFDBF}
\definecolor{pal-heat-peak-cross}{HTML}{FFFFFF}
\definecolor{pal-heat-peak-outline}{HTML}{000000}
\definecolor{pal-ov-panel}{HTML}{F9F9FC}
\definecolor{pal-ov-card-edge}{HTML}{B2ABD2}
\definecolor{pal-ov-learned}{HTML}{D8DAEB}
\definecolor{pal-ov-learned-edge}{HTML}{B2ABD2}
\definecolor{pal-ov-frozen}{HTML}{F7F7F7}
\definecolor{pal-ov-frozen-edge}{HTML}{BDBDBD}
\definecolor{pal-ov-deterministic}{HTML}{FFFFFF}
\definecolor{pal-ov-deterministic-edge}{HTML}{B35806}
\definecolor{pal-ov-chevron}{HTML}{B2ABD2}
\definecolor{pal-ov-chevron-act}{HTML}{FDB863}
\definecolor{pal-ov-title-bar}{HTML}{542788}
\definecolor{pal-ov-title-pill}{HTML}{8073AC}
\definecolor{pal-ov-title-text}{HTML}{FFFFFF}
\definecolor{pal-ov-controller-header}{HTML}{FDB863}
\definecolor{pal-ov-head-heatmap}{HTML}{D8DAEB}
\definecolor{pal-ov-head-heatmap-edge}{HTML}{B2ABD2}
\definecolor{pal-ov-head-visibility}{HTML}{D1CDE4}
\definecolor{pal-ov-head-visibility-edge}{HTML}{B2ABD2}
\definecolor{pal-ov-head-stop}{HTML}{FEE0B6}
\definecolor{pal-ov-head-stop-edge}{HTML}{FDB863}
\definecolor{pal-ov-head-bearing}{HTML}{FEDCB1}
\definecolor{pal-ov-head-bearing-edge}{HTML}{FDB863}
\definecolor{pal-ov-case-card}{HTML}{FEE0B6}
\definecolor{pal-ov-loss-a}{HTML}{D8DAEB}
\definecolor{pal-ov-loss-b}{HTML}{FEE0B6}
\definecolor{pal-ov-settings}{HTML}{FEE0B6}
\definecolor{pal-ov-icon}{HTML}{542788}
\definecolor{pal-ov-icon-light}{HTML}{D8DAEB}
\definecolor{pal-ov-accent}{HTML}{E08214}
\definecolor{pal-ov-accent-light}{HTML}{FDB863}
\definecolor{pal-ov-camera}{HTML}{8073AC}
\definecolor{pal-ov-stop-sign}{HTML}{B35806}
\definecolor{pal-ov-pin}{HTML}{542788}
\definecolor{pal-ov-number-circle}{HTML}{E08214}
\definecolor{pal-ov-contribution-badge}{HTML}{542788}
\definecolor{pal-ov-token}{HTML}{542788}
\definecolor{pal-ov-token-light}{HTML}{D8DAEB}
\definecolor{pal-ov-visibility}{HTML}{8073AC}
\definecolor{pal-ov-stop-current}{HTML}{542788}
\definecolor{pal-ov-stop-other}{HTML}{B2ABD2}
\definecolor{pal-ov-mix-r2r}{HTML}{542788}
\definecolor{pal-ov-mix-rxr}{HTML}{E08214}
\definecolor{pal-ov-mix-envdrop}{HTML}{B2ABD2}
\definecolor{pal-support-view-a}{HTML}{8073AC}
\definecolor{pal-support-view-b}{HTML}{E08214}
\definecolor{pal-support-b-only}{HTML}{B35806}
\definecolor{pal-support-common}{HTML}{542788}
\definecolor{pal-support-observed}{HTML}{EEEFF6}
\definecolor{pal-support-hatch}{HTML}{BDBDBD}
\definecolor{pal-support-axis}{HTML}{525252}
\definecolor{pal-support-text}{HTML}{111111}
\definecolor{pal-support-fill}{HTML}{FFFFFF}

\definecolor{hlcolor}{RGB}{255,236,120}
\sethlcolor{hlcolor}
\ifhighlights
  
\else
  \renewcommand{\hl}[1]{#1}
  
\fi

\newcommand{\paper}{AlphaNav\xspace}

\newcommand{\rtor}{R2R-CE\xspace}
\newcommand{\rxr}{RxR-CE\xspace}
\newcommand{\valunseen}{Val-Unseen\xspace}
\newcommand{\valseen}{Val-Seen\xspace}
\providecommand{\cmark}{\ensuremath{\checkmark}}
\providecommand{\xmark}{\ensuremath{\times}}
\providecommand{\na}{--}
\newlength{\oursw}
\providecommand{\ours}[1]{\settowidth{\oursw}{\textbf{#1}}\makebox[\oursw]{#1}}

\usepackage{colortbl}
\colorlet{tab-band}{pal-ours-fill}     %
\colorlet{tab-ours}{pal-plot-face}     %
\colorlet{tab-stripe}{pal-row-stripe}  %
\newcommand{\best}[1]{\textbf{#1}}                                 %
\newcommand{\second}[1]{\underline{#1}}                            %
\newcommand{\oursrow}{\rowcolor{tab-ours}}                         %
\newcommand{\oursname}[1]{\textcolor{pal-ours}{\textbf{#1}}}       %
\newcommand{\stripe}{\rowcolor{tab-stripe}}                        %
\newcommand{\bandtitle}[1]{\textcolor{pal-ours}{\textbf{#1}}}      %
\newcommand{\bandspan}[2]{\rowcolor{tab-band}\multicolumn{#1}{@{}>{\tintLR}l@{}}{#2}}
\newcommand{\groupband}[2]{\bandspan{#1}{\bandtitle{#2}}\\}
\newlength{\tintbleed}
\makeatletter
\newcommand{\tintL}{\gdef\CT@cell@color{\@tempdimb\tintbleed\global\let\CT@cell@color\relax}}
\newcommand{\tintR}{\gdef\CT@cell@color{\@tempdimc\tintbleed\global\let\CT@cell@color\relax}}
\newcommand{\tintLR}{\gdef\CT@cell@color{\@tempdimb\tintbleed\@tempdimc\tintbleed\global\let\CT@cell@color\relax}}
\makeatother

\InputIfFileExists{results_macros}{}{\newcommand{\pend}[1]{\hl{[pending: #1]}}}
\providecommand{\rCostTrainTyp}{\pend{rCostTrainTyp}}
\newcommand{\figinc}[2]{\IfFileExists{figures/#1.pdf}{\includegraphics[width=#2]{figures/#1}}{\fbox{\parbox{#2}{\centering\hl{[figure pending]}}}}}

\DeclareMathOperator{\softmax}{softmax}

\DeclareMathOperator{\KL}{KL}
\newcommand{\nbins}{N}                    %
\newcommand{\hist}{K}                     %
\newcommand{\img}{\bm{I}}                 %
\newcommand{\kap}{\bm{\kappa}}            %
\newcommand{\bearp}{\bm{p}}               %

%% file: results_macros.tex
\newcommand{\rFinSR}{53.3}
\newcommand{\rFinSRSd}{1.4}
\newcommand{\rFinSRA}{53.6}
\newcommand{\rFinSRB}{51.8}
\newcommand{\rFinSRC}{54.5}
\newcommand{\rFinOSR}{62.2}
\newcommand{\rFinOSRSd}{1.0}
\newcommand{\rFinOSRA}{62.2}
\newcommand{\rFinOSRB}{61.2}
\newcommand{\rFinOSRC}{63.2}
\newcommand{\rFinSPL}{49.3}
\newcommand{\rFinSPLSd}{1.7}
\newcommand{\rFinSPLA}{49.4}
\newcommand{\rFinSPLB}{47.5}
\newcommand{\rFinSPLC}{50.8}
\newcommand{\rFinNE}{5.64}
\newcommand{\rFinNESd}{0.11}
\newcommand{\rFinNEA}{5.63}
\newcommand{\rFinNEB}{5.76}
\newcommand{\rFinNEC}{5.53}
\newcommand{\rMedDecisions}{16}
\newcommand{\rFinSRciA}{[47.7, 58.7]}
\newcommand{\rFinSRciB}{[45.4, 57.9]}
\newcommand{\rFinSRciC}{[47.7, 60.7]}
\newcommand{\rFinSRmeanLo}{47.1}
\newcommand{\rFinSRmeanHi}{58.8}

\newcommand{\rCostTrainMinA}{29.95}
\newcommand{\rCostTrainMinB}{30.55}
\newcommand{\rCostTrainMinC}{31.31}
\newcommand{\rParamsTotal}{2203}
\newcommand{\rParamsTrain}{75.8}
\newcommand{\rVisTokens}{192}
\newcommand{\rFwdMs}{64}
\newcommand{\rDecMs}{65}
\newcommand{\rInferGB}{4.3}
\newcommand{\rTrainGB}{7.9}
\newcommand{\rLoopMs}{85.7}
\newcommand{\rLoopN}{141{,}907}
\newcommand{\rPrimsPerDec}{3.65}
\newcommand{\rMsPerPrim}{17.9}
\newcommand{\rMeanDecisions}{25.8}
\newcommand{\rMoveMean}{0.45}
\newcommand{\rMoveMax}{1.05}
\newcommand{\rMoveZero}{37}
\newcommand{\rStopPassFail}{8.9}
\newcommand{\rStopPassFailStop}{8.5}

\newcommand{\rStopNeverFailStop}{34.0}
\newcommand{\rStopNone}{4.2}
\newcommand{\rStopFarEp}{32.4}
\newcommand{\rStopFarFail}{69.3}
\newcommand{\rStopPrec}{55.6}
\newcommand{\rStopNear}{36.4}

\newcommand{\rStopECE}{0.076}
\newcommand{\rStopAUROC}{0.89}
\newcommand{\rStopAUROCThree}{0.86}
\newcommand{\rStopMeanPNear}{0.54}
\newcommand{\rStopMeanPFar}{0.07}

\newcommand{\rStopDecN}{142{,}458}
\newcommand{\rStopTopN}{5{,}284}
\newcommand{\rStopTopMeanP}{0.955}
\newcommand{\rStopTopNear}{45.4}
\newcommand{\rStopBinEightNear}{25.6}
\newcommand{\rStopBinFiveNear}{19.1}

\newcommand{\rStopRepSREight}{52.9}
\newcommand{\rStopRepSPLEight}{49.8}
\newcommand{\rStopRepNEEight}{5.56}
\newcommand{\rStopRepDecEight}{22.9}
\newcommand{\rStopRepSREightDev}{61.3}
\newcommand{\rStopRepSRSeven}{52.3}
\newcommand{\rStopRepSPLSeven}{49.7}
\newcommand{\rStopRepNESeven}{5.52}
\newcommand{\rStopRepDecSeven}{21.7}
\newcommand{\rStopRepSRSevenDev}{61.4}
\newcommand{\rStopRepSRSix}{51.6}
\newcommand{\rStopRepSPLSix}{49.3}
\newcommand{\rStopRepNESix}{5.54}
\newcommand{\rStopRepDecSix}{20.8}
\newcommand{\rStopRepSRSixDev}{60.7}
\newcommand{\rStopRepSRFive}{50.8}
\newcommand{\rStopRepSPLFive}{48.7}
\newcommand{\rStopRepNEFive}{5.53}
\newcommand{\rStopRepDecFive}{20.1}
\newcommand{\rStopRepSRFiveDev}{60.5}
\newcommand{\rStopRepSRFour}{49.6}
\newcommand{\rStopRepSPLFour}{47.8}
\newcommand{\rStopRepNEFour}{5.56}
\newcommand{\rStopRepDecFour}{19.5}
\newcommand{\rStopRepSRFourDev}{59.8}
\newcommand{\rStopRepSRThree}{48.5}
\newcommand{\rStopRepSPLThree}{46.9}
\newcommand{\rStopRepNEThree}{5.56}
\newcommand{\rStopRepDecThree}{18.8}
\newcommand{\rStopRepSRThreeDev}{58.9}
\newcommand{\rStopRepSRNoCorr}{52.1}
\newcommand{\rStopRepSPLNoCorr}{49.5}
\newcommand{\rStopRepNENoCorr}{5.52}
\newcommand{\rStopRepDecNoCorr}{21.3}
\newcommand{\rStopRepSRNoCorrDev}{61.1}
\newcommand{\rStopRepSRNine}{53.3}
\newcommand{\rStopRepSPLNine}{49.3}
\newcommand{\rStopRepNENine}{5.64}
\newcommand{\rStopRepDecNine}{25.8}
\newcommand{\rStopRepSRNineDev}{60.8}
\newcommand{\rStopBoundBuilding}{54.2}
\newcommand{\rStopBoundEpisode}{61.5}
\newcommand{\rStopBoundBuildingA}{54.7}

\newcommand{\rStopBoundBuildingDev}{64.5}
\newcommand{\rStopSelThrDev}{0.7}
\newcommand{\rStopSelSRDev}{61.4}
\newcommand{\rStopSelUnseen}{52.3}
\newcommand{\rStopPosWeight}{4.64}
\newcommand{\rStopOffset}{1.53}
\newcommand{\rStopNoCorrThr}{0.66}
\newcommand{\rStopCheckEightDev}{300}
\newcommand{\rStopCheckSevenDev}{300}
\newcommand{\rStopECEA}{0.070}
\newcommand{\rStopAUROCA}{0.89}

\newcommand{\rStopVFiveECE}{0.034}
\newcommand{\rStopVFiveAUROC}{0.91}

\newcommand{\rStopVFiveFarEp}{33.3}

\newcommand{\rCalaN}{99{,}061}
\newcommand{\rCalaP}{0.001}
\newcommand{\rCalaTwo}{0.9}
\newcommand{\rCalaThree}{3.3}
\newcommand{\rCalbN}{10{,}783}
\newcommand{\rCalbP}{0.025}
\newcommand{\rCalbTwo}{6.3}
\newcommand{\rCalbThree}{16.3}
\newcommand{\rCalcN}{4{,}277}
\newcommand{\rCalcP}{0.072}
\newcommand{\rCalcTwo}{10.1}
\newcommand{\rCalcThree}{22.1}
\newcommand{\rCaldN}{4{,}672}
\newcommand{\rCaldP}{0.146}
\newcommand{\rCaldTwo}{11.9}
\newcommand{\rCaldThree}{24.1}
\newcommand{\rCaleN}{3{,}007}
\newcommand{\rCaleP}{0.248}
\newcommand{\rCaleTwo}{14.6}
\newcommand{\rCaleThree}{26.3}
\newcommand{\rCalfN}{2{,}374}
\newcommand{\rCalfP}{0.349}
\newcommand{\rCalfTwo}{15.1}
\newcommand{\rCalfThree}{26.2}
\newcommand{\rCalgN}{2{,}200}
\newcommand{\rCalgP}{0.450}
\newcommand{\rCalgTwo}{17.6}
\newcommand{\rCalgThree}{29.5}
\newcommand{\rCalhN}{2{,}077}
\newcommand{\rCalhP}{0.550}
\newcommand{\rCalhTwo}{19.1}
\newcommand{\rCalhThree}{31.3}
\newcommand{\rCaliN}{2{,}285}
\newcommand{\rCaliP}{0.651}
\newcommand{\rCaliTwo}{23.2}
\newcommand{\rCaliThree}{35.4}
\newcommand{\rCaljN}{2{,}637}
\newcommand{\rCaljP}{0.752}
\newcommand{\rCaljTwo}{26.7}
\newcommand{\rCaljThree}{40.1}
\newcommand{\rCalkN}{3{,}801}
\newcommand{\rCalkP}{0.854}
\newcommand{\rCalkTwo}{25.6}
\newcommand{\rCalkThree}{36.9}
\newcommand{\rCallN}{5{,}284}
\newcommand{\rCallP}{0.955}
\newcommand{\rCallTwo}{45.4}
\newcommand{\rCallThree}{55.6}

\newcommand{\rOLFinNinetyDev}{12.9}

\newcommand{\rStopSeedAFarEp}{32.9}

\newcommand{\rCtlCommit}{46.9}
\newcommand{\rCtlCommitMoves}{82.7}
\newcommand{\rCtlCommitColl}{5.5}
\newcommand{\rCtlTurn}{28.9}

\newcommand{\rCtlTwo}{15.2}

\newcommand{\rCtlTwoColl}{23.2}
\newcommand{\rCtlGuard}{5.2}

\newcommand{\rCtlGuardColl}{19.5}
\newcommand{\rCtlStop}{3.7}
\newcommand{\rCtlDecN}{142{,}458}
\newcommand{\rCtlCommitFull}{81.4}
\newcommand{\rCtlShort}{12.9}
\newcommand{\rMemAppend}{52.5}
\newcommand{\rMemStill}{23.1}
\newcommand{\rMemBand}{15.8}

\newcommand{\rDecAmbig}{0.9}
\newcommand{\rDecChanged}{0.0}
\newcommand{\rTokInstMean}{31.4}
\newcommand{\rTokInstMed}{29}
\newcommand{\rTokInstMax}{133}
\newcommand{\rTokImage}{210}
\newcommand{\rTokSeqMean}{241}
\newcommand{\rTokSeqMax}{343}
\newcommand{\rTflopsOurs}{1.31}
\newcommand{\rTflopsOursMax}{1.59}
\newcommand{\rTflopsVision}{0.62}
\newcommand{\rTokPerSec}{3{,}708}
\newcommand{\rTflopsPerSec}{20.1}
\newcommand{\rParVision}{407}
\newcommand{\rParDecoder}{1409}
\newcommand{\rDevInfBaseSR}{58.7}
\newcommand{\rDevInfStopEightSR}{58.3}
\newcommand{\rDevInfStopNineSR}{60.0}
\newcommand{\rDevInfHiddenOneSR}{57.7}
\newcommand{\rDevInfSpacingSR}{62.0}
\newcommand{\rAblDStageOneDev}{49.0}
\newcommand{\rFarShareFail}{71}
\newcommand{\rStopRepSRSevenEvery}{49.0}
\newcommand{\rStopRepSPLSevenEvery}{46.8}
\newcommand{\rStopRepNESevenEvery}{5.66}
\newcommand{\rNormTflopsGAVLN}{8.73}
\newcommand{\rNormCtxNaVILA}{1{,}737}
\newcommand{\rNormGenNaVILA}{10}

\newcommand{\rLitCMTwoSR}{34.3}
\newcommand{\rLitCMTwoSPL}{27.6}
\newcommand{\rLitCMTwoOSR}{41.5}
\newcommand{\rLitCMTwoNE}{7.02}
\newcommand{\rLitLAWSR}{35.0}
\newcommand{\rLitLAWSPL}{31.0}
\newcommand{\rLitLAWOSR}{44.0}
\newcommand{\rLitLAWNE}{6.83}
\newcommand{\rLitWSMGMapSR}{38.9}
\newcommand{\rLitWSMGMapSPL}{34.3}
\newcommand{\rLitWSMGMapOSR}{47.6}
\newcommand{\rLitWSMGMapNE}{6.28}
\newcommand{\rLitCANavSR}{25.3}
\newcommand{\rLitCANavSPL}{10.8}
\newcommand{\rLitCANavOSR}{48.0}
\newcommand{\rLitCANavNE}{7.58}

\newcommand{\rLitAOPlannerSR}{25.5}
\newcommand{\rLitAOPlannerSPL}{16.6}
\newcommand{\rLitAOPlannerOSR}{38.3}
\newcommand{\rLitAOPlannerNE}{6.95}

\newcommand{\rLitInstructNavSR}{31.0}
\newcommand{\rLitInstructNavSPL}{24.0}

\newcommand{\rLitInstructNavNE}{6.89}

\newcommand{\rLitDreamNavSR}{32.8}
\newcommand{\rLitDreamNavSPL}{29.0}
\newcommand{\rLitDreamNavOSR}{41.0}
\newcommand{\rLitDreamNavNE}{7.06}

\newcommand{\rLitNaVILASR}{54.0}
\newcommand{\rLitNaVILASPL}{49.0}

\newcommand{\rLitStreamVLNSR}{52.8}
\newcommand{\rLitStreamVLNSPL}{47.2}

\newcommand{\rLitAuxThinkSR}{54.8}
\newcommand{\rLitAuxThinkSPL}{46.9}

\newcommand{\rLitMonoDreamSPL}{49.1}

\newcommand{\rLatNaVidMs}{1{,}220}

\newcommand{\rLatUniNaVidMs}{200}

\newcommand{\rLatNaVILAMs}{595}

\newcommand{\rLatNaVILAMem}{18.5}

\newcommand{\rLatStreamVLNMs}{270}

\newcommand{\rLatMonoDreamMs}{800}

\newcommand{\rLatGoalTwoPixelMs}{120}

\newcommand{\rLatGAVLNMs}{259}

\newcommand{\rLatEffVLNTwoMs}{159}

\newcommand{\rLatDualVLNMem}{20}

\newcommand{\rLatQwenRobotNavMs}{196}

\newcommand{\rLatCAMPNavMs}{65}

\newcommand{\rLatCAMPNavMem}{4.3}

\newcommand{\rAZeroFSR}{39.4}
\newcommand{\rAZeroFSPL}{37.6}
\newcommand{\rAZeroFOSR}{45.2}
\newcommand{\rAZeroFNE}{6.58}
\newcommand{\rAFourFSR}{36.1}
\newcommand{\rAFourFSPL}{34.7}

\newcommand{\rAFourFNE}{6.86}
\newcommand{\rAOneFSR}{43.0}
\newcommand{\rAOneFSPL}{41.4}
\newcommand{\rAOneFOSR}{49.2}
\newcommand{\rAOneFNE}{6.13}

\newcommand{\rDStageOneSR}{43.9}
\newcommand{\rDStageOneSPL}{41.6}
\newcommand{\rDStageOneOSR}{49.2}
\newcommand{\rDStageOneNE}{6.46}
\newcommand{\rDVOneSR}{47.1}
\newcommand{\rDVOneSPL}{45.2}
\newcommand{\rDVOneOSR}{51.5}
\newcommand{\rDVOneNE}{5.73}
\newcommand{\rDVTwoSR}{49.3}
\newcommand{\rDVTwoSPL}{47.1}
\newcommand{\rDVTwoOSR}{53.7}
\newcommand{\rDVTwoNE}{5.72}
\newcommand{\rXCeilSR}{99.7}

\newcommand{\rXFinPixSR}{28.2}
\newcommand{\rXFinPixSPL}{18.4}

\newcommand{\rXAZeroFPixSR}{22.7}

\newcommand{\rXFinScanSR}{39.7}
\newcommand{\rXFinScanSPL}{35.6}
\newcommand{\rXFinScanOSR}{49.6}
\newcommand{\rXFinScanNE}{5.89}
\newcommand{\rXOrStopSR}{72.5}

\newcommand{\rDbStageOneSR}{42.7}

\newcommand{\rDbVTwoSR}{48.5}

\newcommand{\rDcStageOneSR}{45.8}

\newcommand{\rDcVTwoSR}{49.4}

\newcommand{\rXVFiveSR}{53.2}
\newcommand{\rXVFiveSPL}{50.1}

\newcommand{\rRoneSR}{32.5}
\newcommand{\rRoneSPL}{30.7}
\newcommand{\rRoneOSR}{38.3}
\newcommand{\rRoneNE}{6.70}
\newcommand{\rRtwoSR}{84.6}

\newcommand{\rRthreeSR}{44.0}
\newcommand{\rRthreeSPL}{42.4}
\newcommand{\rRthreeOSR}{49.2}
\newcommand{\rRthreeNE}{6.27}
\newcommand{\rdDoseOne}{\ensuremath{+5.4}}
\newcommand{\rdDoseOneLo}{\ensuremath{+2.1}}
\newcommand{\rdDoseOneHi}{\ensuremath{+8.2}}

\newcommand{\rdDoseV}{\ensuremath{+2.2}}
\newcommand{\rdDoseVLo}{\ensuremath{+0.7}}
\newcommand{\rdDoseVHi}{\ensuremath{+3.7}}

\newcommand{\rdFinAZeroF}{\ensuremath{+13.9}}
\newcommand{\rdFinAZeroFLo}{\ensuremath{+8.7}}
\newcommand{\rdFinAZeroFHi}{\ensuremath{+17.5}}

\newcommand{\rdOrientAll}{\ensuremath{+3.5}}
\newcommand{\rdOrientAllLo}{\ensuremath{+1.1}}
\newcommand{\rdOrientAllHi}{\ensuremath{+6.0}}

\newcommand{\rdRule}{\ensuremath{+4.2}}
\newcommand{\rdRuleLo}{\ensuremath{+1.6}}
\newcommand{\rdRuleHi}{\ensuremath{+7.2}}

\newcommand{\rdOrient}{\ensuremath{+4.6}}
\newcommand{\rdOrientLo}{\ensuremath{+1.2}}
\newcommand{\rdOrientHi}{\ensuremath{+8.7}}

\newcommand{\rdOrientAbs}{4.6}
\newcommand{\rdOrientGuard}{\ensuremath{+2.3}}
\newcommand{\rdOrientGuardLo}{\ensuremath{-0.1}}
\newcommand{\rdOrientGuardHi}{\ensuremath{+4.5}}

\newcommand{\rdTrunc}{\ensuremath{-3.3}}
\newcommand{\rdTruncLo}{\ensuremath{-5.7}}
\newcommand{\rdTruncHi}{\ensuremath{-1.2}}

\newcommand{\rdTruncAbs}{3.3}
\newcommand{\rdReadFin}{\ensuremath{+25.4}}
\newcommand{\rdReadFinLo}{\ensuremath{+20.1}}
\newcommand{\rdReadFinHi}{\ensuremath{+30.9}}

\newcommand{\rdReadAZeroF}{\ensuremath{+16.7}}
\newcommand{\rdReadAZeroFLo}{\ensuremath{+12.8}}
\newcommand{\rdReadAZeroFHi}{\ensuremath{+20.1}}

\newcommand{\rdReadFinScan}{\ensuremath{+13.9}}
\newcommand{\rdReadFinScanLo}{\ensuremath{+8.5}}
\newcommand{\rdReadFinScanHi}{\ensuremath{+19.3}}

\newcommand{\rdOracleStop}{\ensuremath{+18.9}}
\newcommand{\rdOracleStopLo}{\ensuremath{+16.4}}
\newcommand{\rdOracleStopHi}{\ensuremath{+22.1}}

\newcommand{\rdOracleStopAbs}{18.9}
\newcommand{\rdDoseTwo}{\ensuremath{+5.8}}
\newcommand{\rdDoseTwoLo}{\ensuremath{+3.9}}
\newcommand{\rdDoseTwoHi}{\ensuremath{+8.2}}

\newcommand{\rdDoseThree}{\ensuremath{+3.5}}
\newcommand{\rdDoseThreeLo}{\ensuremath{+1.1}}
\newcommand{\rdDoseThreeHi}{\ensuremath{+5.3}}

\newcommand{\rdLenShort}{\ensuremath{+7.8}}
\newcommand{\rdLenShortLo}{\ensuremath{+3.6}}
\newcommand{\rdLenShortHi}{\ensuremath{+11.9}}

\newcommand{\rdLenMid}{\ensuremath{+16.9}}
\newcommand{\rdLenMidLo}{\ensuremath{+11.6}}
\newcommand{\rdLenMidHi}{\ensuremath{+21.3}}

\newcommand{\rdLenLong}{\ensuremath{+16.8}}
\newcommand{\rdLenLongLo}{\ensuremath{+6.4}}
\newcommand{\rdLenLongHi}{\ensuremath{+23.4}}

\newcommand{\rdVFive}{\ensuremath{-0.3}}
\newcommand{\rdVFiveLo}{\ensuremath{-2.6}}
\newcommand{\rdVFiveHi}{\ensuremath{+2.4}}

\newcommand{\rdRxRecipe}{\ensuremath{+5.0}}
\newcommand{\rdRxRecipeLo}{\ensuremath{+0.3}}
\newcommand{\rdRxRecipeHi}{\ensuremath{+8.1}}

\newcommand{\rdRxScan}{\ensuremath{+13.9}}
\newcommand{\rdRxScanLo}{\ensuremath{+11.2}}
\newcommand{\rdRxScanHi}{\ensuremath{+15.8}}

\newcommand{\rdFullA}{\ensuremath{+9.6}}

\newcommand{\rdFullB}{\ensuremath{+9.1}}

\newcommand{\rdFullC}{\ensuremath{+8.6}}

\newcommand{\rdReadScanAhead}{\ensuremath{+15.5}}

\newcommand{\rdReadScanBehind}{\ensuremath{+12.5}}

\newcommand{\rdDvTwoDev}{\ensuremath{+5.6}}
\newcommand{\rdDvTwoDevLo}{\ensuremath{+0.6}}
\newcommand{\rdDvTwoDevHi}{\ensuremath{+10.9}}

\newcommand{\rdDvThreeDev}{\ensuremath{-1.5}}
\newcommand{\rdDvThreeDevLo}{\ensuremath{-4.8}}
\newcommand{\rdDvThreeDevHi}{\ensuremath{+1.9}}

\newcommand{\rdAblDDev}{\ensuremath{+13.7}}
\newcommand{\rdAblDDevLo}{\ensuremath{+9.2}}
\newcommand{\rdAblDDevHi}{\ensuremath{+17.8}}

\newcommand{\rdMemKDev}{\ensuremath{-62.7}}
\newcommand{\rdMemKDevLo}{\ensuremath{-68.3}}
\newcommand{\rdMemKDevHi}{\ensuremath{-56.7}}

\newcommand{\rdHfovSixtyDev}{\ensuremath{+6.4}}
\newcommand{\rdHfovSixtyDevLo}{\ensuremath{+3.6}}
\newcommand{\rdHfovSixtyDevHi}{\ensuremath{+9.5}}

\newcommand{\rdHfovTwentyDev}{\ensuremath{+10.0}}
\newcommand{\rdHfovTwentyDevLo}{\ensuremath{+7.6}}
\newcommand{\rdHfovTwentyDevHi}{\ensuremath{+12.7}}

\newcommand{\rdCamPreDev}{\ensuremath{+5.0}}
\newcommand{\rdCamPreDevLo}{\ensuremath{+1.9}}
\newcommand{\rdCamPreDevHi}{\ensuremath{+8.2}}

\newcommand{\rdCamMethodDev}{\ensuremath{-5.3}}
\newcommand{\rdCamMethodDevLo}{\ensuremath{-8.9}}
\newcommand{\rdCamMethodDevHi}{\ensuremath{-1.8}}

\newcommand{\rdCamAltDev}{\ensuremath{-14.6}}
\newcommand{\rdCamAltDevLo}{\ensuremath{-17.6}}
\newcommand{\rdCamAltDevHi}{\ensuremath{-11.4}}

\newcommand{\rdLoopMemDev}{\ensuremath{+0.3}}
\newcommand{\rdLoopMemDevLo}{\ensuremath{-1.6}}
\newcommand{\rdLoopMemDevHi}{\ensuremath{+2.3}}

\newcommand{\rdDvFiveDev}{\ensuremath{-1.2}}
\newcommand{\rdDvFiveDevLo}{\ensuremath{-5.3}}
\newcommand{\rdDvFiveDevHi}{\ensuremath{+3.2}}

\newcommand{\rdRevReadStageOne}{\ensuremath{+11.4}}
\newcommand{\rdRevReadStageOneLo}{\ensuremath{+4.7}}
\newcommand{\rdRevReadStageOneHi}{\ensuremath{+18.9}}
\newcommand{\rdRevStepsCtl}{\ensuremath{+0.1}}
\newcommand{\rdRevStepsCtlLo}{\ensuremath{-3.1}}
\newcommand{\rdRevStepsCtlHi}{\ensuremath{+3.2}}
\newcommand{\rdRevOnPolicyCtl}{\ensuremath{+5.3}}
\newcommand{\rdRevOnPolicyCtlLo}{\ensuremath{+2.8}}
\newcommand{\rdRevOnPolicyCtlHi}{\ensuremath{+8.2}}
\newcommand{\rdRevRouteLearnedStop}{\ensuremath{+31.0}}
\newcommand{\rdRevRouteLearnedStopLo}{\ensuremath{+26.1}}
\newcommand{\rdRevRouteLearnedStopHi}{\ensuremath{+36.1}}
\newcommand{\rdRevRouteOracleStop}{\ensuremath{+27.2}}
\newcommand{\rdRevRouteOracleStopLo}{\ensuremath{+23.2}}
\newcommand{\rdRevRouteOracleStopHi}{\ensuremath{+32.3}}
\newcommand{\rdRevStopOracleRoute}{\ensuremath{+15.1}}
\newcommand{\rdRevStopOracleRouteLo}{\ensuremath{+10.6}}
\newcommand{\rdRevStopOracleRouteHi}{\ensuremath{+21.4}}

\newcommand{\rHAZeroFSRa}{48.4}

\newcommand{\rHAZeroFSRb}{43.8}

\newcommand{\rHAZeroFSRc}{32.6}

\newcommand{\rHAZeroFSRd}{34.7}

\newcommand{\rHAZeroFRot}{64.4}

\newcommand{\rHAZeroFBehindSR}{33.8}

\newcommand{\rHAOneFSRa}{49.2}

\newcommand{\rHAOneFSRb}{47.6}

\newcommand{\rHAOneFSRc}{39.0}

\newcommand{\rHAOneFSRd}{38.0}

\newcommand{\rHAOneFRot}{67.3}

\newcommand{\rHAOneFBehindSR}{38.4}

\newcommand{\rHAFourFSRa}{48.7}

\newcommand{\rHAFourFSRb}{37.6}

\newcommand{\rHAFourFSRc}{31.2}

\newcommand{\rHAFourFSRd}{29.5}

\newcommand{\rHAFourFRot}{56.2}

\newcommand{\rHFinASRa}{61.9}
\newcommand{\rHFinANa}{417}
\newcommand{\rHFinASRb}{57.6}
\newcommand{\rHFinANb}{420}
\newcommand{\rHFinASRc}{46.9}
\newcommand{\rHFinANc}{426}
\newcommand{\rHFinASRd}{49.5}
\newcommand{\rHFinANd}{576}
\newcommand{\rHFinARot}{69.0}

\newcommand{\rHFinABehindSR}{48.4}

\newcommand{\rHFinBRot}{65.5}

\newcommand{\rHFinCRot}{67.8}

\newcommand{\rHXFinPixSRa}{41.0}

\newcommand{\rHXFinPixSRb}{24.3}

\newcommand{\rHXFinPixSRc}{21.1}

\newcommand{\rHXFinPixSRd}{26.9}

\newcommand{\rHXFinPixRot}{7.6}

\newcommand{\rHXFinScanSRa}{46.5}

\newcommand{\rHXFinScanSRb}{41.9}

\newcommand{\rHXFinScanSRc}{37.1}

\newcommand{\rHXFinScanSRd}{35.1}

\newcommand{\rHXFinScanRot}{50.3}

\newcommand{\rHXFinScanBehindSR}{35.9}

\newcommand{\rHDStageOneBehindSR}{37.7}

\newcommand{\rHDVOneBehindSR}{42.8}

\newcommand{\rHDVTwoBehindSR}{45.3}

\newcommand{\rBehindN}{1{,}002}

\newcommand{\rSceneN}{11}
\newcommand{\rSceneWinN}{9}

\newcommand{\rSAZeroF}{0.44}
\newcommand{\rSEligAZeroF}{389}
\newcommand{\rSDropAZeroF}{8.5}
\newcommand{\rSAOneF}{0.21}

\newcommand{\rSDropAOneF}{4.1}

\newcommand{\rTAZeroFFar}{44.5}

\newcommand{\rTAFourFFar}{49.3}

\newcommand{\rTFinAFar}{32.9}

\newcommand{\rTFinARise}{0.090}

\newcommand{\rTFinCLate}{0.100}

\newcommand{\rMemEightSRDev}{62.7}

\newcommand{\rMemEightSPLDev}{58.4}
\newcommand{\rMemFourSRDev}{18.0}

\newcommand{\rMemFourSPLDev}{13.5}
\newcommand{\rMemZeroSRDev}{0.0}

\newcommand{\rMemZeroSPLDev}{0.0}
\newcommand{\rdMemFourDev}{\ensuremath{-44.7}}
\newcommand{\rDevVTwoSR}{60.8}

\newcommand{\rDevVThreeSR}{59.3}

\newcommand{\rPolicyShareVOne}{47.3}
\newcommand{\rPolicyShareVTwo}{65.0}

\newcommand{\rCamFinalCamNomDev}{62.7}

\newcommand{\rCamFinalCamSixtyDev}{44.7}

\newcommand{\rCamFinalCamTwentyDev}{14.7}

\newcommand{\rCamFinalCamPitchDev}{56.0}

\newcommand{\rCamFZeroNomDev}{53.3}

\newcommand{\rCamFZeroSixtyDev}{40.3}

\newcommand{\rCamFZeroTwentyDev}{35.7}

\newcommand{\rCamFZeroPitchDev}{51.7}

\newcommand{\rCamFOneNomDev}{49.3}

\newcommand{\rCamFOneSixtyDev}{35.7}

\newcommand{\rCamFOneTwentyDev}{4.0}

\newcommand{\rCamFOnePitchDev}{44.3}

\newcommand{\rCamFTwoNomDev}{42.0}

\newcommand{\rCamFTwoSixtyDev}{30.0}

\newcommand{\rCamFTwoTwentyDev}{33.3}

\newcommand{\rCamFTwoPitchDev}{35.7}

\newcommand{\rCamFFourNomDev}{47.0}

\newcommand{\rCamFFourSixtyDev}{32.7}

\newcommand{\rCamFFourTwentyDev}{5.3}

\newcommand{\rCamFFourPitchDev}{45.0}

\newcommand{\rRxScanSR}{27.3}
\newcommand{\rRxScanOSR}{40.1}
\newcommand{\rRxScanSPL}{23.1}
\newcommand{\rRxScanNE}{8.72}
\newcommand{\rRxScanNDTW}{40.3}
\newcommand{\rRxLoopMs}{95}
\newcommand{\rRxTokSeq}{319}
\newcommand{\rRxTokInstr}{109}
\newcommand{\rRxDecEp}{43.9}

\newcommand{\rRxDecMs}{83}
\newcommand{\rRxInferGB}{4.4}
\newcommand{\rRxFinSR}{40.8}
\newcommand{\rRxFinSRSd}{1.4}
\newcommand{\rRxFinSRA}{41.2}
\newcommand{\rRxFinSRB}{39.3}
\newcommand{\rRxFinSRC}{42.0}
\newcommand{\rRxStageOneSR}{36.2}
\newcommand{\rRxFinOSR}{55.2}
\newcommand{\rRxFinOSRSd}{1.1}
\newcommand{\rRxFinOSRA}{55.7}
\newcommand{\rRxFinOSRB}{53.9}
\newcommand{\rRxFinOSRC}{55.9}
\newcommand{\rRxStageOneOSR}{46.6}
\newcommand{\rRxFinSPL}{36.4}
\newcommand{\rRxFinSPLSd}{1.4}
\newcommand{\rRxFinSPLA}{36.7}
\newcommand{\rRxFinSPLB}{34.9}
\newcommand{\rRxFinSPLC}{37.6}
\newcommand{\rRxStageOneSPL}{33.6}
\newcommand{\rRxFinNE}{8.11}
\newcommand{\rRxFinNESd}{0.38}
\newcommand{\rRxFinNEA}{7.95}
\newcommand{\rRxFinNEB}{8.53}
\newcommand{\rRxFinNEC}{7.83}
\newcommand{\rRxStageOneNE}{8.66}
\newcommand{\rRxFinNDTW}{47.7}
\newcommand{\rRxFinNDTWSd}{1.6}
\newcommand{\rRxFinNDTWA}{48.1}
\newcommand{\rRxFinNDTWB}{45.9}
\newcommand{\rRxFinNDTWC}{49.0}
\newcommand{\rRxStageOneNDTW}{48.6}

\newcommand{\rRxFinSRmeanLo}{36.9}
\newcommand{\rRxFinSRmeanHi}{44.6}
\newcommand{\rRxN}{3{,}669}
\newcommand{\rRxUnreach}{99}
\newcommand{\rRxLangUS}{44.7}
\newcommand{\rRxLangIN}{38.9}
\newcommand{\rRxLangUSN}{1{,}223}
\newcommand{\rRxLangINN}{2{,}446}
\newcommand{\rRxWords}{96}
\newcommand{\rRxLitVLNThreeDFFSR}{25.5}
\newcommand{\rRxLitVLNThreeDFFSPL}{18.1}
\newcommand{\rRxLitVLNThreeDFFOSR}{36.7}
\newcommand{\rRxLitVLNThreeDFFNE}{8.79}

\newcommand{\rRxLitMapNavSR}{32.6}
\newcommand{\rRxLitMapNavSPL}{27.7}

\newcommand{\rRxLitMapNavNE}{7.62}
\newcommand{\rRxLitMapNavNDTW}{43.5}
\newcommand{\rRxLitInternVLANOneSR}{53.5}
\newcommand{\rRxLitInternVLANOneSPL}{46.1}

\newcommand{\rRxLitInternVLANOneNE}{5.91}
\newcommand{\rRxLitInternVLANOneNDTW}{65.3}
\newcommand{\rRxLitGAVLNSR}{55.4}
\newcommand{\rRxLitGAVLNSPL}{45.2}
\newcommand{\rRxLitGAVLNOSR}{67.0}
\newcommand{\rRxLitGAVLNNE}{5.88}

\newcommand{\rRxLitUniNaVidSR}{48.7}
\newcommand{\rRxLitUniNaVidSPL}{40.9}
\newcommand{\rRxLitUniNaVidOSR}{55.5}
\newcommand{\rRxLitUniNaVidNE}{6.24}

\newcommand{\rRxLitNaVILASR}{49.3}
\newcommand{\rRxLitNaVILASPL}{44.0}

\newcommand{\rRxLitNaVILANE}{6.77}
\newcommand{\rRxLitNaVILANDTW}{58.8}
\newcommand{\rRxLitStreamVLNSR}{48.6}
\newcommand{\rRxLitStreamVLNSPL}{42.5}

\newcommand{\rRxLitStreamVLNNE}{6.72}
\newcommand{\rRxLitStreamVLNNDTW}{60.2}
\newcommand{\rRxLitStreamVLNScaleSR}{54.4}
\newcommand{\rRxLitStreamVLNScaleSPL}{45.4}

\newcommand{\rRxLitStreamVLNScaleNE}{5.65}
\newcommand{\rRxLitStreamVLNScaleNDTW}{63.7}
\newcommand{\rRxLitAuxThinkSR}{52.2}
\newcommand{\rRxLitAuxThinkSPL}{40.2}
\newcommand{\rRxLitAuxThinkOSR}{61.9}
\newcommand{\rRxLitAuxThinkNE}{6.24}

\newcommand{\rRxLitMonoDreamSR}{49.4}
\newcommand{\rRxLitMonoDreamSPL}{40.9}
\newcommand{\rRxLitMonoDreamOSR}{55.8}
\newcommand{\rRxLitMonoDreamNE}{6.38}

\newcommand{\rRxLitVLNRoneSR}{21.4}
\newcommand{\rRxLitVLNRoneSPL}{15.5}
\newcommand{\rRxLitVLNRoneOSR}{30.6}
\newcommand{\rRxLitVLNRoneNE}{10.20}

\newcommand{\rRxLitGoalTwoPixelSR}{42.7}
\newcommand{\rRxLitGoalTwoPixelSPL}{39.5}

\newcommand{\rRxLitGoalTwoPixelNE}{7.46}
\newcommand{\rRxLitGoalTwoPixelNDTW}{60.7}
\newcommand{\rRxLitEffVLNOneSR}{63.5}
\newcommand{\rRxLitEffVLNOneSPL}{52.1}

\newcommand{\rRxLitEffVLNOneNE}{4.40}
\newcommand{\rRxLitEffVLNOneNDTW}{66.8}
\newcommand{\rRxLitEffVLNTwoSR}{75.6}
\newcommand{\rRxLitEffVLNTwoSPL}{60.0}

\newcommand{\rRxLitEffVLNTwoNE}{3.18}

\newcommand{\rRxLitDualVLNSR}{61.4}
\newcommand{\rRxLitDualVLNSPL}{51.8}

\newcommand{\rRxLitDualVLNNE}{4.58}
\newcommand{\rRxLitDualVLNNDTW}{70.0}
\newcommand{\rRxLitQwenRobotNavSR}{71.3}
\newcommand{\rRxLitQwenRobotNavSPL}{61.5}

\newcommand{\rRxLitQwenRobotNavNE}{4.15}
\newcommand{\rRxLitQwenRobotNavNDTW}{68.6}
\newcommand{\rRxDeltaRtor}{12.5}
\newcommand{\rRxUnreachPct}{2.7}
\newcommand{\rRxGapMonoDream}{8.6}
\newcommand{\rRxGapGoalTwoPixel}{1.9}
\newcommand{\rRxGapVLNR}{19.4}

\newcommand{\rCostTrainTyp}{30.6}

\newcommand{\rGapVLNR}{27.7}

\newcommand{\rGapUni}{6.3}

\newcommand{\rGapNaVid}{15.9}

\newcommand{\rGapMonoDream}{2.5}
\newcommand{\rGapGoalTwoPixel}{0.8}

\newcommand{\rBehindShare}{54.5}
\newcommand{\rBudgetIso}{6.9}
\newcommand{\rBudgetLoop}{5.3}
\newcommand{\rCostAllTyp}{41.8}
\newcommand{\rCostGen}{11.2}

\newcommand{\rNormTflopsNaVILA}{28.0}
\newcommand{\rFlopRatioGAVLN}{6.7}
\newcommand{\rFlopRatioNaVILA}{21.4}
\newcommand{\rStopBoundGainA}{1.1}
\newcommand{\rStopBoundShare}{6.0}
\newcommand{\rStopSelDelta}{\ensuremath{-0.9}}
\newcommand{\rStopCorrGain}{1.2}
\newcommand{\rDecHz}{15}
\newcommand{\rSpeedStream}{4}
\newcommand{\rSpeedNaVILA}{9}
\newcommand{\rSpeedMonoDream}{12}
\newcommand{\rSpeedGoalTwoPixel}{1.8}
\newcommand{\rPrimStream}{68}
\newcommand{\rPrimUni}{50}
\newcommand{\rPrimEff}{40}
\newcommand{\rCamKeepPitchDev}{89}
\newcommand{\rCamKeepSixtyDev}{71}
\newcommand{\rCamKeepTwentyDev}{23}
\newcommand{\rCtlCommitFree}{94.5}
\newcommand{\rMemKeepTen}{92.3}
\newcommand{\rMemKeepTwenty}{91.2}
\newcommand{\rdSpacingNineDev}{\ensuremath{+2.7}}
\newcommand{\rdSpacingUnseen}{\ensuremath{+3.3}}
\newcommand{\rdStopNineUnseen}{\ensuremath{+0.9}}
\newcommand{\rCostRatioLo}{18}
\newcommand{\rCostRatioHi}{92}
\newcommand{\rCostRatioAllLo}{13}
\newcommand{\rCostRatioAllHi}{67}
\newcommand{\rRxReachSR}{41.9}

\newcommand{\rStopFailShare}{41}

%% file: sec/0_abstract.tex
\begin{abstract}
Instruction-following robots need navigation policies that reach high accuracy and remain inexpensive to train and to run on a single graphics processing unit (GPU).
The most accurate vision-language navigation policies for continuous environments rely on language models with 4 to 8 billion parameters, long observation histories, and large auxiliary corpora.
Two properties of the standard training setup limit smaller policies.
Imitation labels cover the states that an expert visits, whereas the running policy reaches states that its own errors produce.
In addition, an image-plane waypoint exists only while the route lies inside the camera view.
We present \textbf{\paper}, a monocular red-green-blue (RGB) policy with 2 billion (2B) parameters that addresses both limitations.
\paper adapts Qwen3-VL-2B with low-rank adapters and processes an eight-frame memory and the current view in one forward pass.
The policy predicts the route direction as a bearing in the body frame of the agent, which remains defined for every heading, and a deterministic controller converts this bearing into motion.
This design allows training to relabel the states that the policy itself visits.
On-policy correction improves success, while bearing decoding outperforms image-plane waypoint scanning across all evaluated start headings with identical weights.
On the \valunseen split, \paper achieves \rFinSR\% success on Room-to-Room in Continuous Environments (\rtor), averaged over three seeds, and \rRxFinSR\% on its Room-Across-Room counterpart (\rxr).
Inference requires \rDecMs\,ms and \rInferGB\,GiB of memory, while training takes \rCostTrainTyp\ GPU-hours on one 12\,GB consumer GPU.
\end{abstract}

%% file: sec/1_intro.tex
\section{Introduction}
\label{sec:intro}

\begin{figure}[t]

  \centering
  \figinc{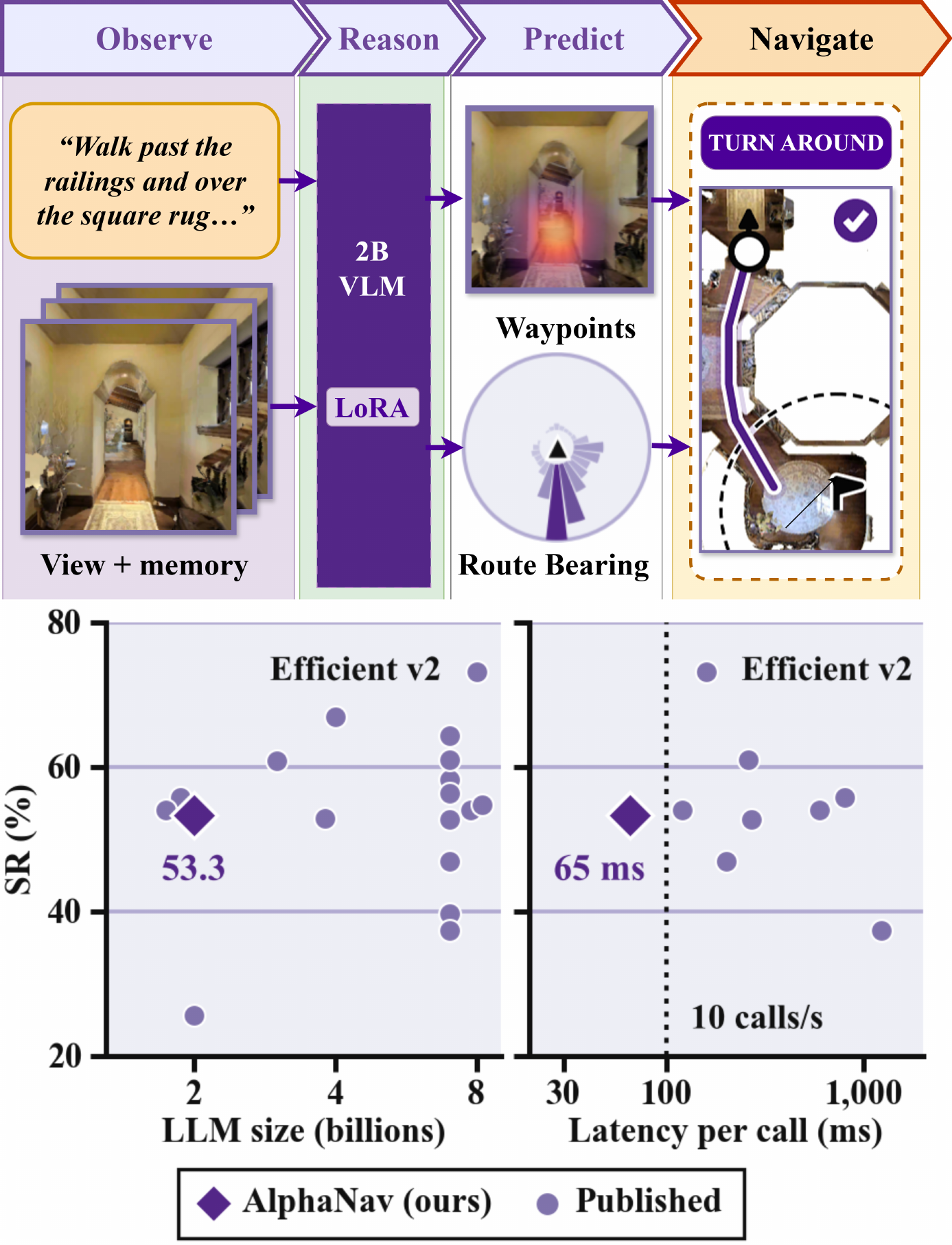}{0.9\columnwidth}
  \caption{\paper on \rtor \valunseen.
  Top: one decision of the vision-language model (VLM).
  Bottom: success rate (SR) against language-model (LLM) size and latency per call; \paper: mean of three seeds; others as reported, without one remotely timed system on the right.
  Dotted line: 10 calls per second.}
  \label{fig:teaser}
  \vspace{-1.0em}
\end{figure}

Instruction-following navigation enables a robot to cross an unfamiliar building by following a route description.
Vision-and-language navigation in continuous environments (VLN-CE) formulates this task with low-level actions, namely 0.25\,m forward steps, $15^\circ$ turns, and a stop action~\cite{krantz2020vlnce}.
On the standard Room-to-Room in Continuous Environments (\rtor) benchmark, video-language models predict these actions from sequences of red-green-blue (RGB) observations.
NaVid~\cite{zhang2024navid} introduced this approach, Uni-NaVid~\cite{zhang2024uninavid} extended it to several navigation tasks, and NaVILA~\cite{cheng2025navila} and StreamVLN~\cite{wei2025streamvln} scaled it further.
These systems are built on language models with 7B-8B parameters and long observation histories, and their training adds web video, question answering, or synthetic navigation episodes to the benchmark data.
StreamVLN and NaVILA report \rLatStreamVLNMs\ and \rLatNaVILAMs\,ms per model call, respectively, on an RTX~4090 GPU, and training these 7B and 8B systems costs 544 to 2{,}816 data-center GPU-hours (details in Table~\ref{tab:cost}).
\begin{figure*}[t]
  \centering
  \vspace{-1.5em}
  \includegraphics[width=0.95\textwidth]{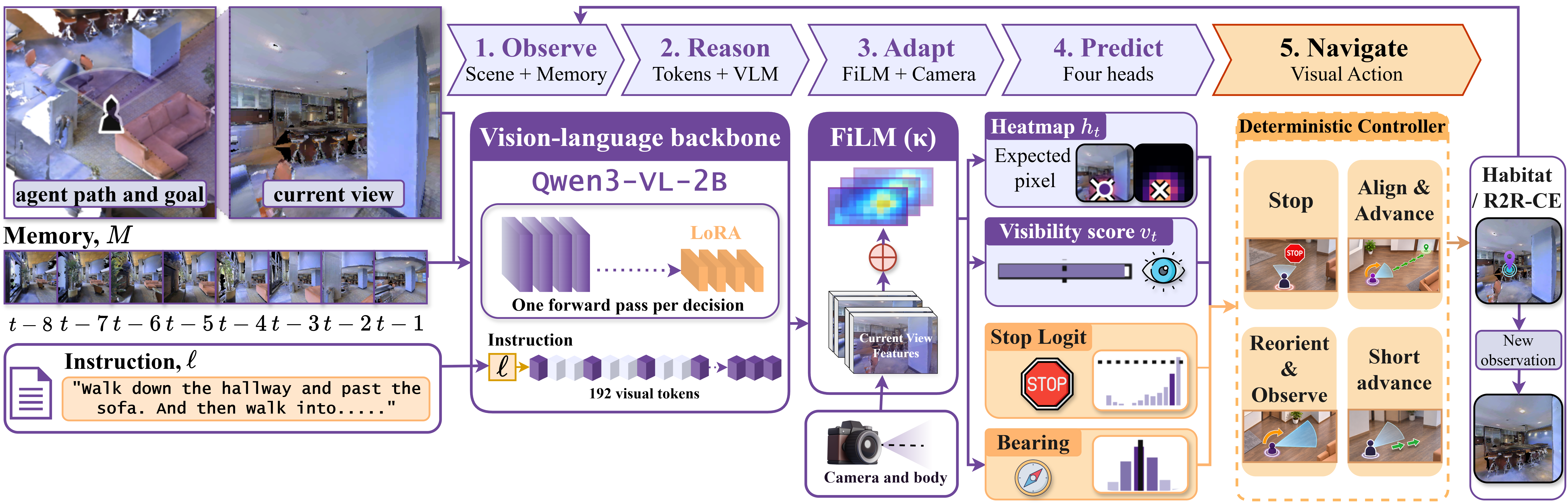}
  \caption{Overview of \paper.
  The instruction $\ell$, an eight-frame memory $M$, and the current frame enter a Qwen3-VL-2B backbone with LoRA adapters as \rVisTokens\ visual tokens.
  FiLM conditions the current-view features on the camera and body parameter vector $\bm\kappa$.
  Four heads predict a waypoint heatmap with its expected pixel, a visibility score, a stop logit, and a 24-bin route-bearing distribution.
  A deterministic controller selects among stopping, aligning and advancing, reorienting and observing, and advancing a short distance before the next observation arrives in the Habitat simulator.}
  \label{fig:overview}
  \vspace{-0.3em}
\end{figure*}

In an instruction-following setup, a deployed robot must act several times per second on its own hardware, so the latency, memory footprint, and training cost of its policy determine where that policy can run and who can afford to train it. Fig.~\ref{fig:teaser} demonstrates \paper among published systems along two axes, language-model size and latency.
It is visible that a 2B-parameter policy is on par in success rate with the 7B StreamVLN and the 8B NaVILA and Aux-Think at the lowest reported latency, whereas the 2B MonoDream~\cite{wang2025monodream} requires \rLatMonoDreamMs\,ms per call and the 2B Goal2Pixel~\cite{goal2pixel2026} needs depth at test time.
We identify two limitations of the standard training setup that stand in the way of compact policies.
The \textit{first} concerns the training data, since imitation covers only the states an expert visits, whereas a running policy drifts into new states after its own errors~\cite{ross2011dagger}. The \textit{second} concerns the action representation, because an image-plane waypoint can express only routes inside the camera view, yet in \rBehindShare\% of the \valunseen episodes the first route step lies more than $90^\circ$ from the start heading.

\paper retains a standard vision-language backbone and instead changes the output representation and the training-state distribution (Fig.~\ref{fig:overview}).
The instruction, eight memory frames, and the current frame form one sequence with \rVisTokens\ visual tokens for Qwen3-VL-2B~\cite{bai2025qwen3vl}, which we fine-tune with low-rank adaptation (LoRA)~\cite{hu2022lora}.
A feature-wise linear modulation (FiLM) layer~\cite{perez2018film} conditions the current-view tokens on known camera and body parameters.
Four heads then predict a waypoint heatmap, a visibility score, a stop probability, and a 24-bin route bearing, which encodes the direction of the route in the body frame and remains defined for every heading.
A deterministic controller converts these predictions into turns and at most four forward moves.
Training first imitates reference paths and then relabels the states of the policy's own rollouts in rounds of on-policy correction (Secs.~\ref{sec:training} and~\ref{sec:suppl_training}).  The core contributions of AlphaNav are as follows:
\begin{enumerate}[leftmargin=*,itemsep=2pt,topsep=2pt]

\item We present \paper, a compact monocular policy that adapts Qwen3-VL-2B with LoRA and processes eight memory frames and the current view in one forward pass.
Over three seeds, \paper reaches \rFinSR\,$\pm$\,\rFinSRSd\% success rate (SR) and \rFinSPL\% success weighted by path length (SPL) on \rtor \valunseen.

\item We train the policy on the states that it visits, and this relabeling raises SR from \rDStageOneSR\% to \rDVTwoSR\%.
With oracle inputs the controller succeeds in \rXCeilSR\% of the episodes, and an oracle stop raises the SR of one trained model from \rFinSRA\% to \rXOrStopSR\%.

\item We pair the image-plane waypoint with a 24-bin body-frame route-bearing head that remains defined for every heading, and we isolate its effect by decoding the same weights both ways.
The bearing raises SR from \rXFinScanSR\% to \rFinSRA\%, and on the baseline recipe the orient-then-go augmentation adds \rdOrientAbs\ points on starts that face away from the route.

\item \paper has the lowest latency of the ten compared systems that report one and trains on one consumer GPU.
One decision requires \rDecMs\,ms and \rInferGB\,GiB of memory, and training requires \rCostTrainTyp\ GPU-hours plus \rCostGen\ GPU-hours of label rendering on one RTX~5070; training alone takes \rCostRatioLo\ to \rCostRatioHi\ times fewer GPU-hours than the 7B and 8B systems report.

\end{enumerate}

%% file: sec/2_related.tex
\section{Related Work}
\label{sec:related}
\paragraph{Navigation in continuous environments.}
VLN-CE~\cite{krantz2020vlnce} extends Room-to-Room (R2R)~\cite{anderson2018vln} to continuous navigation with low-level actions; Room-Across-Room (RxR)~\cite{ku2020rxr} provides longer instructions, while EnvDrop~\cite{tan2019envdrop} uses back-translation.
Depth supports several planning representations: semantic maps in CM2 and WS-MGMap~\cite{georgakis2022cm2,chen2022wsmgmap}, language-aligned waypoints in LAW~\cite{raychaudhuri2021law}, and panoramic topological graphs in ETP-R1 and LCGNav~\cite{ye2025etpr1,peng2026lcgnav}.
Other depth-based methods construct 3D feature fields~\cite{wang2024vln3dff,wang2024g3dlf} or provide VLMs with annotated maps and layered 3D tokens~\cite{zhang2025mapnav,wang2025dynam3d}.
Zero-shot planners prompt large language models with constraint-aware or affordance-oriented plans~\cite{chen2024canav,chen2024aoplanner}, generic instructions~\cite{long2024instructnav}, or predicted trajectories~\cite{wang2025dreamnav}.
RGB-only policies remove depth dependence: NaVid and Uni-NaVid predict actions from RGB histories~\cite{zhang2024navid,zhang2024uninavid}, while NaVILA and StreamVLN use 8B and 7B backbones~\cite{cheng2025navila,wei2025streamvln}.
Aux-Think examines reasoning strategies for data-efficient training~\cite{wang2025auxthink}, while DualVLN and InternVLA-N1 combine slow planners with fast low-level policies~\cite{wei2025dualvln,internvlan1_2025}.
Efficiency is pursued through 2B backbones in VLN-R1 and MonoDream~\cite{qi2025vlnr1,wang2025monodream}, reduced training cost in Efficient-VLN~\cite{zheng2025efficientvln}, and geometry-aware bird's-eye representations in GA-VLN~\cite{yang2026gavln}.
MemVLN addresses memory cost by compressing history into 300 tokens within a 4B model~\cite{wang2026memvln}.
However, some 2B policies require test-time depth~\cite{goal2pixel2026}, incur higher latency~\cite{wang2025monodream}, or achieve lower success~\cite{qi2025vlnr1}, motivating closer attention to training and output design.

\paragraph{Training states, output interfaces, and cameras.}
Beyond model size, training must account for the states encountered during deployment. DAgger addresses the distribution shift of behavior cloning by collecting labels for states encountered by the learner~\cite{ross2011dagger}.
Efficient-VLN uses within-episode correction~\cite{zheng2025efficientvln}, while JOP-VLN and ETP-R1 additionally apply GRPO~\cite{he2026jopvln,ye2025etpr1}.
Output representations determine target availability: panoramic models predict polar waypoints from 12-view RGB-D observations~\cite{krantz2021waypoint,hong2022bridging}, covering every heading.
Goal2Pixel lifts image-plane goals using depth~\cite{goal2pixel2026}, while Qwen-RobotNav predicts robot-centric waypoints that remain defined outside the camera view~\cite{qwenrobotnav2026}.
With one camera, however, image-plane waypoints require a visible route point and cannot supervise routes behind the agent, whether initially or after a wrong turn.
Camera-robust policies canonicalize geometry across platforms~\cite{kim2026canonnav,rehberg2026sensorlayout} or enforce cross-view action consistency~\cite{huang2026crossview}.
Augmentation additionally requires consistency between transformed observations, directional instructions, and supervision.
For mirroring, this includes swapping left/right words in instructions. Within a single 2B RGB policy, \paper conditions on camera parameters, applies only augmentations that preserve exact labels, and represents eight raw $128\times128$ history frames with 128 tokens.

%% file: sec/3_method.tex
\section{Method}
\label{sec:method}

\subsection{Problem formulation}
\label{sec:overview}

At each decision step $t$, the policy receives an instruction $\ell$, a $256\times256$ RGB frame $\img_t$, a memory $\mathcal{M}_t$ of earlier decision frames, and a known vector $\kap$ of camera and body parameters.
The RGB camera is the only exteroceptive sensor, and odometry serves only to maintain the memory.
The policy $\pi_\psi$ maps these inputs to predictions $o_t$, which a deterministic controller $\mathcal{C}$ converts into a sequence $a_t$ of benchmark primitives, namely 0.25\,m forward moves, $15^\circ$ turns, and as shown in Eq.~\ref{eq:decision}.
\begin{equation}
  o_t = \pi_\psi(\ell,\mathcal{M}_t,\img_t,\kap), \qquad a_t = \mathcal{C}(o_t).
  \label{eq:decision}
\end{equation}
All primitives in $a_t$ are executed before the next observation.
The policy grounds the instruction in the image, and the controller turns its predictions into motion with fixed rules.
Two choices shape such a compact policy: what it predicts, so that the controller can act from any state, and on which states it is trained.

\subsection{Route bearing in the body frame}
\label{sec:bearing}

An image-plane waypoint, the pixel $\bm{u}_t$ of the next route point, gives the direction $\theta^{\mathrm{pix}}_t=\arctan\!\big((c_x-u_{t,x})/f_x\big)$, with the horizontal focal length $f_x$ and principal-point coordinate $c_x$, only while that point lies inside the field of view.
When the route is beside or behind the agent, either at the start of an episode or after a wrong turn, the waypoint policy must turn to locate it. However, these states lack waypoint labels during training. \paper therefore also predicts the route bearing, the direction of the route target in the body frame of the agent. For a target at body coordinates $(x_t,y_t)$, with $x$ forward and $y$ to the left, the bearing label $\theta^\star_t=\operatorname{atan2}(y_t,x_t)$ is defined at every heading and at every position farther than 0.3\,m from the target.
The policy predicts the bearing as a distribution $\bearp_t$ over $\nbins=24$ circular bins with centers $\theta_k=-\pi+2\pi k/\nbins$, one bin per $15^\circ$ turn primitive, and the predicted bearing $\hat\theta_t$ is the probability-weighted circular mean of the most likely bin and its two neighbors.
At inference, the controller can therefore turn directly toward a route that is out of view, and it keeps the waypoint as the finer target while the route is in view. In training, the direction label exists at every heading, including at off-route states whose target lies out of view and which therefore lack an image-plane label.
On-policy relabeling can thus supervise the output on which the controller acts at all of these states.

\subsection{Architecture}
\label{sec:input}

\paragraph{Backbone, memory, and camera.}
We adapt Qwen3-VL-2B-Instruct~\cite{bai2025qwen3vl} with LoRA adapters~\cite{hu2022lora} on its language decoder and vision tower, and train only these adapters, the camera conditioning, and the heads.
One sequence holds the instruction, the $\hist=8$ memory frames in temporal order ($128\times128$ pixels, 16 tokens each), and the current frame (64 tokens), so every current-view token attends to the instruction and to the whole memory in one forward pass.
At inference, a new frame is appended to the memory only once the agent has moved 0.5\,m since the last appended frame, and otherwise it replaces the newest entry, so that turning in place does not push earlier spatial context out of the memory.
A FiLM layer~\cite{perez2018film}, initialized as the identity, scales and shifts each current-view feature $\bm{z}_{t,j}$ of the final decoder layer with parameters predicted from $\kap\in\mathbb{R}^8$, which gives $\hat{\bm z}_{t,j}$.
The vector $\kap$ holds the logarithms of the focal lengths, the relative principal point, the camera height and pitch, and the robot width and turning radius, so that the mapping from image positions to directions can follow the camera.
The memory reaches these features through decoder attention, so the heads read only current-view features.

\paragraph{Heads.}
A heatmap $\bm{h}_t$ over the $8\times8$ current-view grid gives the waypoint $\bm{u}_t=\sum_j h_{t,j}\bm{c}_j$ as its expectation over the grid centers $\bm{c}_j$.
The same weights pool the features into $\bm{q}_t=\sum_j h_{t,j}\hat{\bm z}_{t,j}$, from which the route visibility $v_t\in(0,1)$ and the stop logit $\zeta_t$ are predicted, so that both read the region where the policy places the route.
The bearing head reads the pooled and the mean current-view features (Eq.~\ref{eq:bearing}), and the stop-gradient $\operatorname{sg}(\cdot)$ keeps the bearing loss from changing the heatmap, which is therefore trained by the waypoint labels alone.
\begin{equation}
  \bm{r}_t = f_\psi\big([\,\operatorname{sg}(\bm{q}_t)\,;\ \tfrac{1}{64}\textstyle\sum_j \hat{\bm z}_{t,j}\,]\big), \qquad \bearp_t=\softmax(\bm{r}_t).
  \label{eq:bearing}
\end{equation}

\subsection{Controller}
\label{sec:heads}

The controller $\mathcal{C}$ follows one principle: it moves far only when the route is in view and ahead, and otherwise it turns toward the predicted bearing and observes again.
\emph{Stop} is issued when the stop probability exceeds a threshold, after $\ln w$ is subtracted from the logit to remove the shift that the class weight $w$ of the stop loss introduces (Sec.~\ref{sec:suppl_stop}).
\emph{Commit} applies when the route is predicted visible within a cone ahead: the agent turns toward $\bm{u}_t$ and moves 1\,m, so one forward pass covers four primitives.
\emph{Reorient} applies otherwise: the agent turns toward $\hat\theta_t$ and observes again, or moves 0.5\,m if it already faces a route that is predicted out of view (the hidden-route rule).
A \emph{turn-loop guard} adds two forward moves after a run of turn-only decisions.
The image-plane readout commits whenever the route is predicted visible, and because $\theta^{\mathrm{pix}}_t$ cannot point outside the view, it scans instead of reorienting: it turns $90^\circ$ in place and observes again.
Table~\ref{tab:hparams} lists the thresholds, and Sec.~\ref{sec:suppl_cost} reports how often each rule acts.

\subsection{Training}
\label{sec:training}

\paragraph{Labels and objective.}
Reference routes connect the annotated waypoints through shortest navigable paths.
At each state, the route target is the farthest route point within 6\,m that lies along a nearly straight path and is visible and unoccluded, and otherwise the next route point.
All labels derive from this target.
The visibility label $v^\star_t$ is 1 when the target is in view, a Gaussian heatmap $\bm{h}^\star_t$ and the pixel $\bm{u}^\star_t$ mark its projection, and the bearing $\theta^\star_t$ is encoded as a circular Gaussian $\bm{y}_t$ of one-bin width.
The stop label $s^\star_t$ is positive within 2\,m of the goal, inside the 3\,m success radius.
Eq.~\ref{eq:loss} combines the Kullback-Leibler (KL) divergence, an L1 term, the cross-entropy, and the binary cross-entropy (BCE).
\begin{equation}
\begin{split}
  \mathcal{L} ={}& \KL(\bm{h}^\star_t\,\|\,\bm{h}_t) + 0.5\,\|\bm{u}_t-\bm{u}^\star_t\|_1 - \textstyle\sum_k y_{t,k}\log p_{t,k} \\
  &+ 0.25\,\mathrm{BCE}(v_t,v^\star_t) + 0.5\,\mathrm{BCE}_{w}(\zeta_t,s^\star_t).
\end{split}
  \label{eq:loss}
\end{equation}
The waypoint terms apply only to targets in view and the bearing term only to valid bearings, and $\mathrm{BCE}_w$ weights the positive stop labels by the ratio $w$ of negative to positive labels.

\paragraph{Imitation and on-policy correction.}
Stage 1 imitates the reference routes of \rtor, English Room-Across-Room in Continuous Environments (\rxr), and EnvDrop~\cite{tan2019envdrop} in training houses only, with randomized appearance and cameras.
Mirroring reflects the frames and spatial labels and swaps ``left'' and ``right'', which keeps the labels exact for a centered principal point.
Orient-then-go draws a uniform start heading and prepends a turning clause such as ``Turn around.'', which adds starts whose route is out of view.
Imitation covers states near the reference routes, whereas the running policy also reaches states that its own errors produce.
Stage 2 therefore follows DAgger~\cite{ross2011dagger}: the policy drives rollouts on \rtor training episodes, every visited state is relabeled from the live pose with the rule above, and the relabeled states are aggregated with the original data.
For an off-route state, the progress along the reference route advances to the nearest route point within a short window ahead, and the target is chosen from there, so the label leads back to the instructed route rather than along a new shortest path to the goal.
Because the bearing label exists at every heading, the correction also covers states whose route is out of view.
As in Efficient-VLN~\cite{zheng2025efficientvln}, the policy controls decision $i$ of $L$ expected decisions with probability $0.6^{\min(1,\,i/(L-1))}$, and the oracle briefly takes over when the agent strays from the path or is blocked.
Stage 1 trains for 10{,}000 steps on 57{,}513 episodes, stage 2 runs three rounds of 1{,}500 rollouts each, and the trained parameters amount to \rParamsTrain\,M of \rParamsTotal\,M.
Sec.~\ref{sec:suppl_training} and Table~\ref{tab:hparams} give the data, the optimization, and the controller thresholds, and Sec.~\ref{sec:suppl_dagger} the round-level schedule that serves as an ablation baseline.

%% file: sec/4_experiments.tex
\section{Experiments}
\label{sec:experiments}

\begin{table*}[t]
  \vspace{-0.6em}
  \centering
  \footnotesize
  \setlength{\tabcolsep}{4.2pt}
  \resizebox{\textwidth}{!}{%
  \begin{tabular}{@{}>{\tintL}llllcccccc<{\tintR}@{}}
    \toprule
    Method & System type & LLM & Obs. & DAgger & Extra data & NE\,$\downarrow$ & OSR\,$\uparrow$ & SR\,$\uparrow$ & SPL\,$\uparrow$ \\
    \midrule
    VLN-3DFF~\cite{wang2024vln3dff}       & \multirow{6}{*}{3D fields / depth VLM} & \na    & RGB-D & \na    & \na    & 5.95 & 55.8 & 44.9 & 30.4 \\
    g3D-LF~\cite{wang2024g3dlf}           &                                        & \na    & RGB-D & \na    & \cmark & 5.70 & 59.5 & 47.2 & 34.6 \\
    MapNav~\cite{zhang2025mapnav}         &                                        & 7B     & RGB-D & \cmark & \cmark & 4.93 & 53.0 & 39.7 & 37.2 \\
    Dynam3D~\cite{wang2025dynam3d}        &                                        & 3.8B   & RGB-D & \cmark & \cmark & 5.34 & 62.1 & 52.9 & 45.7 \\
    InternVLA-N1~\cite{internvlan1_2025}  &                                        & 7B     & RGB-D & \na    & \cmark & 4.83 & 63.3 & 58.2 & 54.0 \\
    GA-VLN~\cite{yang2026gavln}           &                                        & 7B+1B  & RGB-D & \xmark & \cmark & 4.80 & 67.6 & 61.0 & 55.2 \\
    \midrule
    NaVid~\cite{zhang2024navid}                 & \multirow{15}{*}{RGB video-language} & 7B    & RGB              & \cmark & \cmark & 5.47 & 49.1 & 37.4 & 35.9 \\
    Uni-NaVid~\cite{zhang2024uninavid}          &                                      & 7B    & RGB              & \cmark & \cmark & 5.58 & 53.3 & 47.0 & 42.7 \\
    NaVILA~\cite{cheng2025navila}               &                                      & 8B    & RGB              & \na    & \cmark & 5.22 & 62.5 & 54.0 & 49.0 \\
    StreamVLN~\cite{wei2025streamvln}           &                                      & 7B    & RGB              & \cmark & \cmark & 5.43 & 62.5 & 52.8 & 47.2 \\
    StreamVLN + ScaleVLN~\cite{wei2025streamvln}&                                      & 7B    & RGB              & \cmark & \cmark & 4.90 & 63.6 & 56.4 & 50.2 \\
    Aux-Think~\cite{wang2025auxthink}           &                                      & 8B    & RGB              & \cmark & \cmark & 6.08 & 60.0 & 54.8 & 46.9 \\
    MemVLN~\cite{wang2026memvln}                &                                      & 4B    & RGB              & \xmark & \xmark & 5.19 & 61.2 & 55.0 & 50.5 \\
    VLN-R1~\cite{qi2025vlnr1}                   &                                      & 2B    & RGB              & \xmark & \xmark & 10.2 & 37.5 & 25.6 & 20.5 \\
    MonoDream~\cite{wang2025monodream}          &                                      & 2B    & RGB              & \cmark & \xmark & 5.45 & 61.5 & 55.8 & 49.1 \\
    Goal2Pixel$^{\ddagger}$~\cite{goal2pixel2026}&                                     & 2B    & RGB$^{\ddagger}$ & \xmark & \xmark & 4.85 & 59.9 & 54.1 & 52.5 \\
    Efficient-VLN (v1)~\cite{zheng2025efficientvln} &                                  & 3B+1B & RGB              & \cmark & \na    & 4.36 & 69.0 & 60.8 & 53.7 \\
    Efficient-VLN (v2)~\cite{zheng2025efficientvln} &                                  & 8B    & RGB              & \cmark & \cmark & \best{3.29} & \best{78.2} & \best{73.2} & \best{65.0} \\
    DualVLN~\cite{wei2025dualvln}               &                                      & 7B    & RGB              & \na    & \na    & 4.05 & 70.7 & 64.3 & 58.5 \\
    Qwen-RobotNav~\cite{qwenrobotnav2026}       &                                      & 4B    & RGB              & \xmark & \cmark & 4.22 & 73.6 & 66.9 & 60.5 \\
    JOP-VLN~\cite{he2026jopvln}                 &                                      & 8B    & RGB              & \cmark & \cmark & \second{3.64} & \second{76.4} & \second{69.9} & \second{64.9} \\
    \midrule
    \oursrow \oursname{\paper (ours)} & RGB video-language & \ours{2B} & \ours{RGB}$^{\dagger}$ & \cmark & \xmark &
      \ours{\rFinNE}\,{\scriptsize$\pm$\rFinNESd} & \ours{\rFinOSR}\,{\scriptsize$\pm$\rFinOSRSd} &
      \ours{\rFinSR}\,{\scriptsize$\pm$\rFinSRSd} & \ours{\rFinSPL}\,{\scriptsize$\pm$\rFinSPLSd} \\
    \bottomrule
    \vspace{-1.0em}
  \end{tabular}}
  \caption{Comparison with published systems on \rtor \valunseen.
  LLM: language-model size; Obs.: test-time input; DAgger: training on learner-visited states; Extra data: beyond \rtor, \rxr, and EnvDrop.
  \paper: mean and std.\ over three seeds. $^{\dagger}$Odometry also updates the memory (Sec.~\ref{sec:suppl_odometry}). $^{\ddagger}$Depth turns the image goal into a waypoint. \na: not reported. Bold/underline: best/second.}
  \label{tab:main}
\end{table*}

\paragraph{Setup and metrics.}
We evaluate on \rtor~\cite{krantz2020vlnce}, which ports Room-to-Room~\cite{anderson2018vln} on Matterport3D~\cite{chang2017matterport} to the Habitat simulator~\cite{savva2019habitat}.
Unless stated otherwise, all results refer to the \valunseen split, which contains 1{,}839 episodes in 11 buildings that do not appear in training.
The \valseen split, in contrast, contains 778 episodes in the training buildings.
We select checkpoints on 100 and the inference settings on 300 of these episodes before evaluating \paper on \valunseen (Sec.~\ref{sec:suppl_seen}).
At each step, the agent observes a $256\times256$ RGB frame with a $90^\circ$ horizontal field of view, captured from a height of 1.25\,m.
The benchmark limits each episode to 500 primitive actions (see Sec.~\ref{sec:suppl_eval} for the full protocol).
An episode counts as successful if the agent stops within 3\,m of the goal.
Accordingly, we report the success rate (SR), the success weighted by path length (SPL), the oracle success rate (OSR), and the navigation error (NE) in meters.
We train \paper with three seeds and each ablation model with a single seed.
The only exception is the on-policy schedule, which we replicate on all three seeds (Sec.~\ref{sec:exp_dagger}).
Values in brackets are 95\% bootstrap intervals over buildings, differences between models are paired over episodes, and all differences and means are computed before rounding.
Finally, we measure latency and memory at batch size 1 on a single 12\,GB RTX~5070, the same GPU we use for training (Sec.~\ref{sec:suppl_measure}).

\subsection{Comparison with published systems}
\label{sec:main}

\begin{figure}[!t]
  \centering
  \begin{subfigure}[t]{0.45\columnwidth}
    \centering
    \figinc{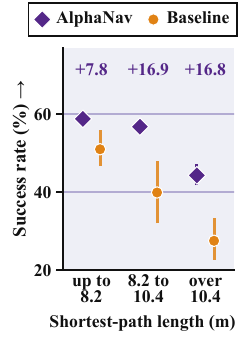}{0.9\linewidth}
    \caption{By route length.}
    \label{fig:strata_a}
  \end{subfigure}
  \hfill
  \begin{subfigure}[t]{0.45\columnwidth}
    \centering
    \figinc{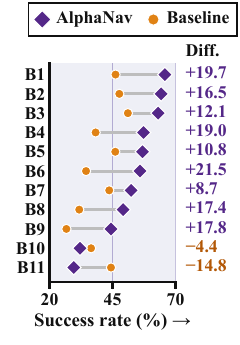}{0.9\linewidth}
    \caption{By building.}
    \label{fig:strata_b}
  \end{subfigure}
  \caption{SR by (a) route-length tercile and (b) building on \rtor \valunseen. Diamonds: \paper; circles: baseline recipe; numbers: gain in points; bars: baseline 95\% CI, \paper\ range over three seeds.}
  \label{fig:strata}
\end{figure}

Using only a single RGB camera, \paper (Table~\ref{tab:main}) matches or exceeds the SR of every map-based, zero-shot (Table~\ref{tab:main_more}), and 3D-feature-field method.
The only exceptions are InternVLA-N1 and GA-VLN, which both rely on depth input and 7B language models.
Among RGB video-language policies, \paper (\rFinSR\% SR, \rFinSPL\% SPL) is on par with the 7B StreamVLN (\rLitStreamVLNSR\% and \rLitStreamVLNSPL\%) and the 8B NaVILA (\rLitNaVILASR\% and \rLitNaVILASPL\%) and Aux-Think (\rLitAuxThinkSR\% and \rLitAuxThinkSPL\%) with less than a third of their parameters, and it trains only on \rtor, \rxr~\cite{ku2020rxr}, and EnvDrop~\cite{tan2019envdrop}.
The 4B MemVLN reaches 55.0\% SR with \rtor and \rxr alone.
\paper outperforms NaVid and Uni-NaVid by \rGapNaVid\ and \rGapUni\ SR points, respectively, and the 2B VLN-R1 by \rGapVLNR\ points. Among the other 2B systems, \paper is on par with MonoDream (\rGapMonoDream\ SR points lower, SPL \rFinSPL\% vs.\ \rLitMonoDreamSPL\%) at \rSpeedMonoDream\ times lower latency, and with Goal2Pixel (\rGapGoalTwoPixel\ points lower), which requires depth at test time and runs at \rSpeedGoalTwoPixel\ times the latency.
Both SRs lie inside the 95\% interval [\rFinSRmeanLo, \rFinSRmeanHi] of our mean over the 11 \valunseen buildings.
The strongest RGB systems, Efficient-VLN, DualVLN, Qwen-RobotNav, and JOP-VLN, reach 61\% to 73\% SR, but they rely on larger backbones, more training data, or reinforcement learning.
The published results are single runs, whereas we report the mean over three seeds, which reach \rFinSRA, \rFinSRB, and \rFinSRC\% SR; the best reaches \rFinSPLC\% SPL. To quantify the effect of our training recipe, we compare \paper with a baseline recipe.
This earlier configuration trains on \rtor alone for 4{,}000 steps, omits the two augmentations, runs only one round of the round-level on-policy schedule (Sec.~\ref{sec:suppl_dagger}), and stops at a threshold of 0.6.
In a paired comparison on all 1{,}839 episodes, \paper gains \rdFinAZeroF\ [\rdFinAZeroFLo, \rdFinAZeroFHi] points over this baseline (from \rAZeroFSR\% to \rFinSR\%).
The gain is positive in every third of the route lengths, with \rdLenShort, \rdLenMid, and \rdLenLong\ points from the shortest to the longest third, and in \rSceneWinN\ of the \rSceneN\ \valunseen buildings (Fig.~\ref{fig:strata}).
\begin{table}[!t]
  \centering
  \footnotesize
  \setlength{\tabcolsep}{3pt}
  \begin{tabular}{@{}>{\tintL}llrrp{1.95cm}<{\tintR}@{}}
    \toprule
    Method & LLM & ms/call & Mem. & Training \\
    \midrule
    NaVid~\cite{zhang2024navid}             & 7B    & \rLatNaVidMs & \na  & 24 A100, $\sim$28\,h \\
    Uni-NaVid~\cite{zhang2024uninavid}      & 7B    & $\sim$\rLatUniNaVidMs$^{d}$ & \na & 40 H800, $\sim$35\,h \\
    NaVILA~\cite{cheng2025navila}           & 8B    & \rLatNaVILAMs$^{a}$ & \rLatNaVILAMem & \na \\
    StreamVLN~\cite{wei2025streamvln}       & 7B    & \rLatStreamVLNMs$^{a}$ & \na  & 1{,}500 A100-h \\
    DualVLN~\cite{wei2025dualvln}           & 7B    & \na & \rLatDualVLNMem & \na \\
    MonoDream~\cite{wang2025monodream}      & 2B    & \rLatMonoDreamMs$^{a}$ & \na  & \na \\
    Goal2Pixel~\cite{goal2pixel2026}        & 2B    & \rLatGoalTwoPixelMs & \na  & 70 H100-h \\
    GA-VLN~\cite{yang2026gavln}             & 7B+1B & \rLatGAVLNMs & \na  & \na \\
    Efficient-VLN v1~\cite{zheng2025efficientvln} & 3B+1B & \na & \na & 282 H800-h \\
    Efficient-VLN v2~\cite{zheng2025efficientvln} & 8B    & \rLatEffVLNTwoMs$^{b}$ & \na & 544 H200-h \\
    Qwen-RobotNav~\cite{qwenrobotnav2026}   & 4B    & \rLatQwenRobotNavMs$^{e}$ & \na & \na \\
    Qwen-RobotNav~\cite{qwenrobotnav2026}   & 8B    & \na & \na & 2{,}816 H100-h \\
    \midrule
    \oursrow \oursname{\paper (ours)} & \ours{2B} & \ours{\rLatCAMPNavMs}$^{c}$ & \ours{\rLatCAMPNavMem} & \rCostTrainTyp\,GPU-h \\
    \bottomrule
  \end{tabular}
  \caption{Cost comparison in ms/call. GPU: $^{a}$RTX~4090, $^{b}$H200, $^{c}$RTX~5070, $^{d}$A100, $^{e}$Unitree Go2 via remote server. Mem. is Memory in GB. \paper numbers are median decision with peak GiB. Unavailable numbers are denoted by '\na'.}
  \label{tab:cost}
\end{table}

\begin{table*}[!t]
  \centering
  \footnotesize
  \setlength{\tabcolsep}{4.5pt}
  \begin{tabular}{@{}>{\tintL}cl ccc ccccc<{\tintR}@{}}
    \toprule
    & & \multicolumn{3}{c}{Components} & \multicolumn{5}{c}{\rtor \valunseen} \\
    \cmidrule(lr){3-5}\cmidrule(l){6-10}
    \# & Model & On-policy share & Readout & Orient-then-go & NE\,$\downarrow$ & OSR\,$\uparrow$ & SR\,$\uparrow$ & SPL\,$\uparrow$ & SR$_{>90^\circ}$\,$\uparrow$ \\
    \midrule
    \bandspan{10}{\bandtitle{(a) On-policy data} \quad (training lineage of \paper, seed 42; rows 1--3: stop threshold 0.7, every frame appended)} \\
    1 & Imitation only            & 0\%                         & bearing & \cmark & \rDStageOneNE& \rDStageOneOSR& \rDStageOneSR& \rDStageOneSPL& \rHDStageOneBehindSR \\
    2 & + round-level schedule     & \rPolicyShareVOne\%         & bearing & \cmark & \rDVOneNE\     & \rDVOneOSR\     & \rDVOneSR\     & \rDVOneSPL\     & \rHDVOneBehindSR \\
    3 & + within-episode schedule  & \rPolicyShareVTwo\%         & bearing & \cmark & \rDVTwoNE\     & \rDVTwoOSR\     & \rDVTwoSR\     & \rDVTwoSPL\     & \rHDVTwoBehindSR \\
    \oursrow 4 & \oursname{+ stop 0.9, memory spacing$^{*}$} & \rPolicyShareVTwo\% & bearing & \cmark & \best{\rFinNEA}& \best{\rFinOSRA}& \best{\rFinSRA}& \best{\rFinSPLA}& \best{\rHFinABehindSR} \\
    \bandspan{10}{\bandtitle{(b) Output interface} \quad (the weights of row 4, decoded two ways)} \\
    \oursrow 4 & \oursname{Route bearing (\paper)} & \rPolicyShareVTwo\%         & bearing & \cmark & \best{\rFinNEA}& \best{\rFinOSRA}& \best{\rFinSRA}& \best{\rFinSPLA}& \best{\rHFinABehindSR} \\
    5 & Image-plane waypoint, scan & \rPolicyShareVTwo\%         & pixel   & \cmark & \rXFinScanNE& \rXFinScanOSR& \rXFinScanSR& \rXFinScanSPL& \rHXFinScanBehindSR \\
    \bandspan{10}{\bandtitle{(c) Augmentation} \quad (baseline recipe: \rtor only, 4{,}000 steps, one round-level round; stop 0.6, memory spacing; one seed each)} \\
    6 & Baseline recipe            & round-level                 & bearing & \xmark      & \rAZeroFNE& \rAZeroFOSR& \rAZeroFSR& \rAZeroFSPL& \rHAZeroFBehindSR \\
    7 & + orient-then-go           & round-level                 & bearing & \cmark      & \best{\rAOneFNE}\  & \best{\rAOneFOSR}\  & \best{\rAOneFSR}\  & \best{\rAOneFSPL}\  & \best{\rHAOneFBehindSR} \\
    \bottomrule
  \end{tabular}
  \caption{Ablations on \rtor \valunseen.
  On-policy share: rollout steps driven by the policy.
  The pixel readout scans in place for routes predicted out of view (Sec.~\ref{sec:heads}).
  SR$_{>90^\circ}$: SR on the \rBehindN\ starts beyond $90^\circ$.
  $^{*}$Selected on \valseen. Bold: best per block.}
  \label{tab:ablation}
\end{table*}

\IfFileExists{figures/fig_qual_cases.pdf}{%
\begin{figure*}[t]
  \centering
  \includegraphics[width=0.9\textwidth,height=2.4in,keepaspectratio]{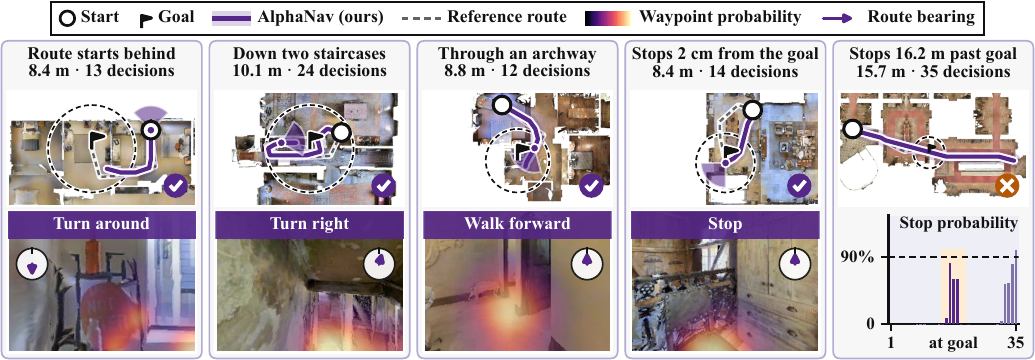}
  \caption{Episodes of \paper (seed 42) on \rtor \valunseen, one per building.
  Successes: chosen by fixed rules (largest start angle, largest height change, a named archway, closest stop) among successes with SPL $\geq 0.85$ that start facing away from the route; failure: a random stop more than 5\,m from the goal.
  Top: path on the floor plan with route length and decisions; wedge: camera view at the shown decision; dashed ring: 3\,m success radius. Bottom: view at one decision with action, waypoint distribution, and route bearing; for the failure, stop probability per decision against the dashed threshold.}
  \label{fig:qualcases}
\end{figure*}}{}

\begin{figure}[!t]
  \centering
  \begin{subfigure}[t]{0.49\columnwidth}
    \centering
    \figinc{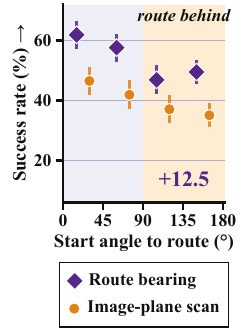}{0.85\linewidth}
    \caption{Output representation.}
    \label{fig:heading_a}
  \end{subfigure}
  \hfill
  \begin{subfigure}[t]{0.49\columnwidth}
    \centering
    \figinc{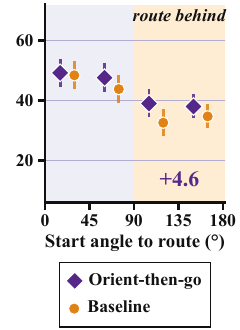}{0.85\linewidth}
    \caption{Orient-then-go training.}
    \label{fig:heading_b}
  \end{subfigure}
  \caption{SR by start-to-route angle on \rtor \valunseen (shaded: $>90^\circ$; bars: 95\% CI; numbers: gain in points on starts beyond $90^\circ$). (a) Route bearing vs.\ image-plane scan, same weights. (b) With vs.\ without orient-then-go.}
  \label{fig:heading}
\end{figure}

\begin{figure*}[t]
  \centering
  \figinc{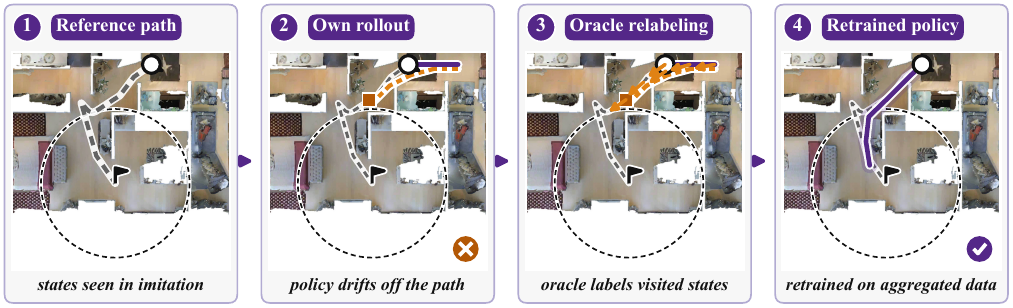}{0.9\textwidth}
  \caption{On-policy correction. (1) Reference path (grey dashed; ring: 3\,m success radius). (2) Rollout after imitation (purple: policy; orange: oracle takeover beyond 1\,m off path; square: end), 3.9\,m from goal. (3) Oracle bearing labels. (4) Later round, 1.0\,m from goal.}
  \label{fig:onpolicy}
\end{figure*}

\subsection{Accuracy and cost}
\label{sec:cost}

Table~\ref{tab:cost} lists the inference and training costs as reported by each paper.
\paper uses \rVisTokens\ visual tokens and \rParamsTrain\,M trainable parameters, and one decision takes \rDecMs\,ms, of which the forward pass accounts for \rFwdMs\,ms.
In closed loop, which additionally includes the controller, the simulator, and rendering, a decision takes \rLoopMs\,ms (Sec.~\ref{sec:suppl_cost}).
In terms of memory, \paper peaks at \rInferGB\,GiB, compared with \rLatNaVILAMem\,GB for NaVILA at 16-bit precision and \rLatDualVLNMem\,GB for DualVLN.
Its latency is the lowest among the ten systems that report one.
Specifically, it is \rSpeedGoalTwoPixel\ times lower than that of Goal2Pixel, the next fastest system, and \rSpeedStream, \rSpeedNaVILA, and \rSpeedMonoDream\ times lower than those of StreamVLN, NaVILA, and MonoDream.
These three systems were measured on an RTX~4090, a faster GPU with 1.5 times the memory bandwidth of our RTX~5070.
Because some systems execute several primitives per call, we also normalize the latency by the number of executed primitives.
Per primitive, \paper requires \rMsPerPrim\,ms, compared with \rPrimStream, \rPrimUni, and \rPrimEff\,ms for StreamVLN, Uni-NaVid, and Efficient-VLN v2, which each predict four actions per call.
To compare the systems independently of the GPU, we further estimate the compute per decision.
One decision of \paper processes \rTokSeqMean\ tokens on average and requires an estimated \rTflopsOurs\ tera floating-point operations (TFLOPs), compared with \rNormTflopsGAVLN\ TFLOPs per call reported for GA-VLN (Table~\ref{tab:norm}).
Training is similarly inexpensive.
It requires \rCostTrainTyp\ GPU-hours on one RTX~5070 with a peak memory of \rTrainGB\,GiB, plus \rCostGen\ GPU-hours for rendering the labels.
Training alone thus takes \rCostRatioLo\ to \rCostRatioHi\ times fewer GPU-hours (\rCostRatioAllLo\ to \rCostRatioAllHi\ times with rendering) than the 544 to 2{,}816 data-center GPU-hours reported by the 7B and 8B systems in Table~\ref{tab:cost} that state their training cost.
The labels are rendered once and shared by the three seeds.
At \rDecMs\,ms per decision, the policy runs at \rDecHz\ decisions per second, above the 10 calls per second marked in Fig.~\ref{fig:teaser}.

\subsection{On-policy data and the controller ceiling}
\label{sec:exp_dagger}

Training on the states that the policy itself visits yields clear gains over the imitation stage (Table~\ref{tab:ablation}a, Figs.~\ref{fig:onpolicy} and~\ref{fig:dose}).
SR rises from \rDStageOneSR\% after imitation to \rDVOneSR\% when the policy drives \rPolicyShareVOne\% of the rollout steps (round-level schedule), and to \rDVTwoSR\% (\rdDoseOne\ [\rdDoseOneLo, \rdDoseOneHi] points) when it drives \rPolicyShareVTwo\% (within-episode schedule).
The within-episode schedule thus gains \rdDoseV\ [\rdDoseVLo, \rdDoseVHi] points over the round-level schedule, which also uses fewer rollouts per round (1{,}000 against 1{,}500).
The gain over imitation replicates on the other two seeds, with \rdDoseTwo\ [\rdDoseTwoLo, \rdDoseTwoHi] and \rdDoseThree\ [\rdDoseThreeLo, \rdDoseThreeHi] points (Table~\ref{tab:seeds_onpolicy}).
On top of this, the inference settings selected on \valseen add \rdRule\ [\rdRuleLo, \rdRuleHi] points (row 4 against row 3); averaged over the three seeds, the memory spacing contributes \rdSpacingUnseen\ points and the stop threshold \rdStopNineUnseen\ points (Table~\ref{tab:stoprep}, Sec.~\ref{sec:suppl_stop}).
To separate the effect of on-policy training from that of the inference settings, we also compare both stages at equal settings on 300 \valseen episodes, where on-policy training raises the SR of seed 42 from \rAblDStageOneDev\% to \rMemEightSRDev\% (\rdAblDDev\ [\rdAblDDevLo, \rdAblDDevHi] points).

To measure how much error stems from the controller, we execute oracle route bearings and stop once within 1.5\,m of the goal; this controller ceiling succeeds in \rXCeilSR\% of the \valunseen episodes.
The remaining failures therefore lie mainly in the policy's predictions, and most of them are stops far from the goal.
In particular, seed 42 stops more than 5\,m from the goal in \rTFinAFar\% of episodes, which accounts for \rFarShareFail\% of its failures.
Replacing only the learned stop with an oracle stop within 1.5\,m of the goal raises the SR of the same navigation from \rFinSRA\% to \rXOrStopSR\% (\rdOracleStop\ [\rdOracleStopLo, \rdOracleStopHi] points), which recovers \rStopFailShare\% of the failures of this seed.
The stop head itself ranks decisions well, with an area under the receiver operating characteristic curve (AUROC) of \rStopAUROC.
In addition, SR stays between \rStopRepSRSix\% and \rStopRepSRNine\% for every threshold from 0.6 to 0.9 (Table~\ref{tab:stoprep}).
The headroom therefore lies not in the threshold but in where the policy stops: over three seeds, \rStopNeverFailStop\% of episodes stop without any decision within 3\,m of the goal, and \rStopPassFail\% pass within 3\,m of the goal but still fail.
Finally, success depends on the memory: on 300 selection episodes with seed 42, keeping only four of the eight memory frames at inference lowers SR from \rMemEightSRDev\% to \rMemFourSRDev\%, and without any memory frames, no episode stops at all (Sec.~\ref{sec:suppl_failure}).

\subsection{Route-bearing output and augmentations}
\label{sec:exp_interface}

To isolate the effect of the output interface, we decode one set of weights in two ways (readouts, Table~\ref{tab:ablation}b).
Both readouts share the training data, the seed, and every weight, so their difference stems from the readout and its turning rule; the on-policy rounds of these weights ran the bearing readout.
Both readouts steer to the waypoint when they commit, and the bearing readout commits only when its decoded bearing also lies within the cone.
Toward a route that is out of view, the route bearing turns directly to the predicted direction, whereas the scan turns $90^\circ$ in place and observes again (Sec.~\ref{sec:heads}).
Decoded as a route bearing, the same weights raise SR from \rXFinScanSR\% to \rFinSRA\% (\rdReadFinScan\ [\rdReadFinScanLo, \rdReadFinScanHi] points).
The bearing leads in all four start-heading bins, by \rdReadScanAhead\ points on starts within $90^\circ$ of the route and by \rdReadScanBehind\ points on starts beyond $90^\circ$ (Fig.~\ref{fig:heading_a} and Sec.~\ref{sec:suppl_interface}).
The gain is even larger on starts that face the route because the route also leaves the view later in the episode: only \rCtlCommit\% of all decisions commit toward a visible route.
On starts beyond $90^\circ$, its first turn faces the route in \rHFinARot\% of the cases, against \rHXFinScanRot\% for the scan, and in \rHFinBRot\% and \rHFinCRot\% for seeds 43 and 44 (Table~\ref{tab:heading}).
On the baseline recipe, the orient-then-go augmentation, which targets starts that face away from the route, adds \rdOrient\ [\rdOrientLo, \rdOrientHi] points on starts beyond $90^\circ$, \rdOrientGuard\ points on the remaining starts, and \rdOrientAll\ [\rdOrientAllLo, \rdOrientAllHi] points over all starts (Fig.~\ref{fig:heading_b}, Table~\ref{tab:ablation}c).
\IfFileExists{figures/fig_qual_cases.pdf}{%
Fig.~\ref{fig:qualcases} shows four successes of \paper (seed 42), each with an SPL of 1.00, and one failure in which the agent passes close to the goal but its stop probability peaks at 0.82, below the threshold of 0.9 (more episodes in Sec.~\ref{sec:suppl_figures}).}{}

\begin{table}[t]
  \centering
  \footnotesize
  \setlength{\tabcolsep}{2.6pt}
  \begin{tabular}{@{}>{\tintL}lcccccr<{\tintR}@{}}
    \toprule
    2B system & NE\,$\downarrow$ & OSR\,$\uparrow$ & SR\,$\uparrow$ & SPL\,$\uparrow$ & nDTW\,$\uparrow$ & ms/call\,$\downarrow$ \\
    \midrule
    MonoDream~\cite{wang2025monodream} & \rRxLitMonoDreamNE & \rRxLitMonoDreamOSR & \rRxLitMonoDreamSR & \rRxLitMonoDreamSPL & \na & \rLatMonoDreamMs$^{a}$ \\
    Goal2Pixel$^{\ddagger}$~\cite{goal2pixel2026} & \rRxLitGoalTwoPixelNE & \na & \rRxLitGoalTwoPixelSR & \rRxLitGoalTwoPixelSPL & \rRxLitGoalTwoPixelNDTW & \rLatGoalTwoPixelMs \\
    VLN-R1~\cite{qi2025vlnr1} & \rRxLitVLNRoneNE & \rRxLitVLNRoneOSR & \rRxLitVLNRoneSR & \rRxLitVLNRoneSPL & \na & \na \\
    \midrule
    \oursrow \oursname{\paper (ours)} & \ours{\rRxFinNE} & \ours{\rRxFinOSR} & \ours{\rRxFinSR} & \ours{\rRxFinSPL} & \ours{\rRxFinNDTW} & \ours{\rRxDecMs}$^{c}$ \\
    \bottomrule
  \end{tabular}
  \caption{2B systems on \rxr \valunseen. \paper: English subset, three-seed mean, ms/call at the median instruction. Footnotes as in Tables~\ref{tab:main} and~\ref{tab:cost}.}
  \label{tab:rxr_main}
\end{table}

\subsection{Results on \texorpdfstring{\rxr}{RxR-CE}}
\label{sec:exp_rxr}

To test whether our findings carry over to longer and more varied instructions, we evaluate the same three models on the \rRxN\ English episodes of \rxr \valunseen~\cite{ku2020rxr,krantz2020vlnce}.
We tune no setting on \rxr, and its English training episodes enter imitation only.
On this benchmark, \paper reaches \rRxFinSR\,$\pm$\,\rRxFinSRSd\% SR [\rRxFinSRmeanLo, \rRxFinSRmeanHi] and \rRxFinSPL\% SPL (Table~\ref{tab:rxr_main}).
This places it \rRxGapVLNR\ points above VLN-R1 and \rRxGapGoalTwoPixel\ and \rRxGapMonoDream\ points below Goal2Pixel and MonoDream, respectively, as reported by these papers (see Table~\ref{tab:rxr} for larger systems).
The gain of the route-bearing output also replicates on \rxr.
On the weights of seed 42, it raises SR from \rRxScanSR\% with the scan readout to \rRxFinSRA\% (\rdRxScan\ [\rdRxScanLo, \rdRxScanHi] points), the same gain as on \rtor (Sec.~\ref{sec:exp_interface}).
Likewise, the imitation-only stage of this seed reaches \rRxStageOneSR\% SR at its own inference settings, \rdRxRecipe\ [\rdRxRecipeLo, \rdRxRecipeHi] points below the final model.
Finally, SR counts the \rRxUnreach\ episodes (\rRxUnreachPct\%) whose goal is unreachable on our navigation mesh as failures.
Excluding these episodes, SR is \rRxReachSR\%.

%% file: sec/5_conclusion.tex
\section{Conclusion and Limitations}
\label{sec:conclusion}

We presented \paper, a compact monocular VLN-CE policy that predicts the route as a body-frame bearing and trains on the states that it visits.
With one RGB camera and a 2B model, \paper reaches \rFinSR\,$\pm$\,\rFinSRSd\% SR and \rFinSPL\% SPL on \rtor over three seeds and \rRxFinSR\% SR on English \rxr.
Each decision requires \rDecMs\,ms and \rInferGB\,GiB of memory, and training requires \rCostTrainTyp\ GPU-hours on one 12\,GB consumer GPU.
On-policy training raises SR from \rDStageOneSR\% to \rDVTwoSR\%, with gains on all three seeds.
The route bearing raises the SR of the same weights from \rXFinScanSR\% to \rFinSRA\% over an image-plane waypoint, in every start-heading bin, with the same gain on \rxr.
With oracle inputs, the controller succeeds in \rXCeilSR\% of episodes, and an oracle stop raises SR from \rFinSRA\% to \rXOrStopSR\%, which recovers \rStopFailShare\% of the failures.

\textit{Limitations and Future work:} Our evaluation follows the standard simulated benchmarks, \rtor and the English subset of \rxr, with one 2B backbone and cameras near the training range (Sec.~\ref{sec:suppl_fov}).
Extending the study to further backbones, to the other languages of \rxr, and to wider cameras is left to future work.
Future work will also improve the stop decision, behind \rStopFailShare\% of the failures, and test the policy on a physical robot.

%% file: sec/llm_usage.tex
\section*{LLM Usage Statement}
The authors developed the research problem, conceptual framework, hypotheses, methodology, theoretical results, and benchmark design independently before using generative AI tools.
OpenAI Codex and Claude subsequently assisted with implementation, routine programming, plotting scripts, supplementary controls, dataset cleaning, and refinement of author-drafted text.
The authors supervised these tools directly, reviewed every assisted output, and verified all citations against the original sources. All reported numerical results were obtained using the authors’ implementation. The code will be made publicly available.

%% file: sec/X_suppl.tex
\clearpage
\setcounter{page}{1}
\renewcommand{\thepage}{S\arabic{page}}
\maketitlesupplementary
\setcounter{section}{0}
\renewcommand{\thesection}{S\arabic{section}}
\def\theHsection{S\arabic{section}}
\makeatletter
\renewcommand\paragraph{\@startsection{paragraph}{4}{\z@}{6pt \@plus 2pt \@minus 1pt}{-1em}{\normalfont\normalsize\bfseries}}
\makeatother
\setlength{\textfloatsep}{8pt plus 2pt minus 1pt}
\linespread{0.965}\selectfont

\section{At a glance}
\label{sec:suppl_glance}

This supplement gives the evidence for each claim of the main paper. Unless stated otherwise, its results use two data splits of \rtor.
The unseen-building evaluation set (\valunseen) holds 1{,}839 episodes in 11 buildings outside the training set, and the main paper reports its results on this set unless it names another one.
The selection set (\valseen) holds 778 held-out episodes in the training buildings, and we use it to select checkpoints and inference settings.
The tables and figures below refer to these splits as unseen buildings and seen buildings, and Sec.~\ref{sec:suppl_rxr} adds the English episodes of \rxr \valunseen.

\section{Qualitative results}
\label{sec:suppl_figures}

This section shows closed-loop episodes on unseen buildings, most of them with a route that starts behind the agent.
Fig.~\ref{fig:qualbest} presents successful episodes whose route starts behind the agent, drawn at random among the successes with an SPL of at least 0.85, with one episode per building.
Fig.~\ref{fig:route} follows one of these routes decision by decision.
The first decision turns the agent around, and forward moves and turns follow until the stop.
Fig.~\ref{fig:qual} then compares \paper with the baseline recipe on starts behind the agent, with one episode per outcome group.
Fig.~\ref{fig:gallery} adds eight episodes across all start headings, drawn with a fixed seed in proportion to the four outcome groups and excluding the episodes of Fig.~\ref{fig:qual}.
Both models succeed in two of these episodes, only \paper succeeds in two, only the baseline succeeds in one, and both models fail in three.

\begin{qfigure*}[t]
  \centering
  \makebox[0.6\textwidth]{\figinc{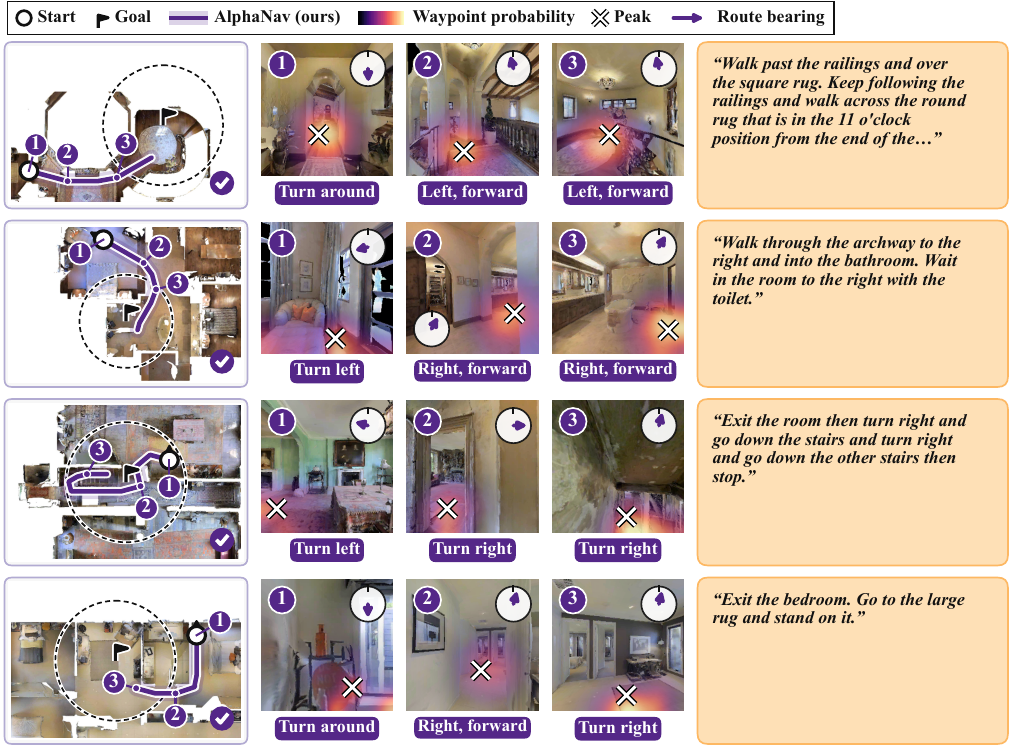}{\textwidth}}
  \caption{Successful closed-loop episodes of \paper on unseen buildings. Each row shows the path on the floor plan with three numbered decisions. At each decision, the frame carries the predicted waypoint distribution as an overlay, its peak as a cross, and the decoded route bearing as an arrow.}
  \label{fig:qualbest}
\end{qfigure*}

\begin{qfigure*}[t]
  \centering
  \figinc{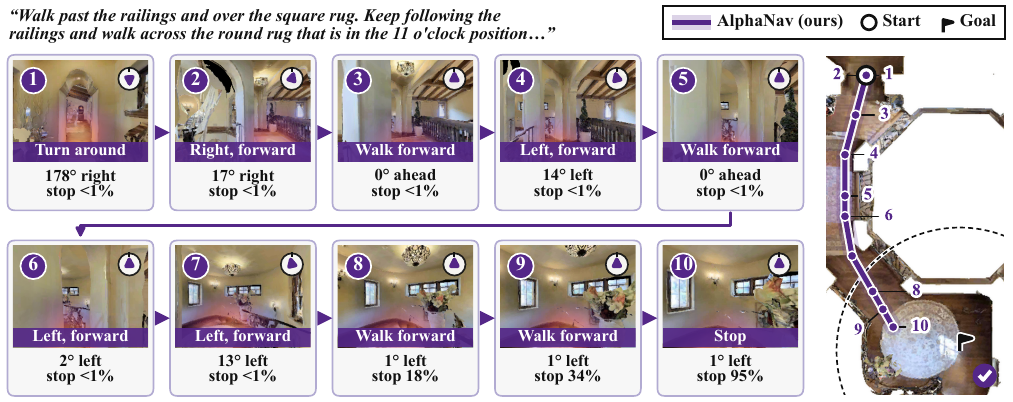}{\textwidth}
  \caption{Decision-by-decision trace of one \paper route on unseen buildings. Each card shows one decision with the central part of the camera view, the action, the decoded route bearing in degrees, and the stop probability. On the compass, straight ahead points up and positive angles point left. The faint overlay marks the predicted waypoint distribution, and the card numbers match the dots on the floor plan.}
  \label{fig:route}
\end{qfigure*}

\begin{qfigure*}[t]
  \centering
  \figinc{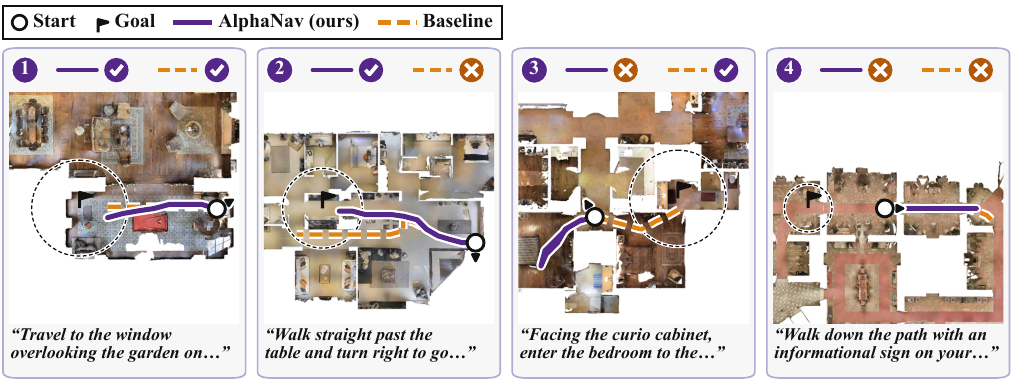}{\textwidth}
  \caption{Closed-loop paths on unseen-building episodes whose route starts behind the agent. The cards show one episode per outcome group: both models succeed, only \paper succeeds, only the baseline recipe succeeds, or both models fail. Rings mark the 3\,m success radius, ticks and crosses mark success and failure, and each card scales to its own path.}
  \label{fig:qual}
\end{qfigure*}

\begin{qfigure*}[t]
  \centering
  \figinc{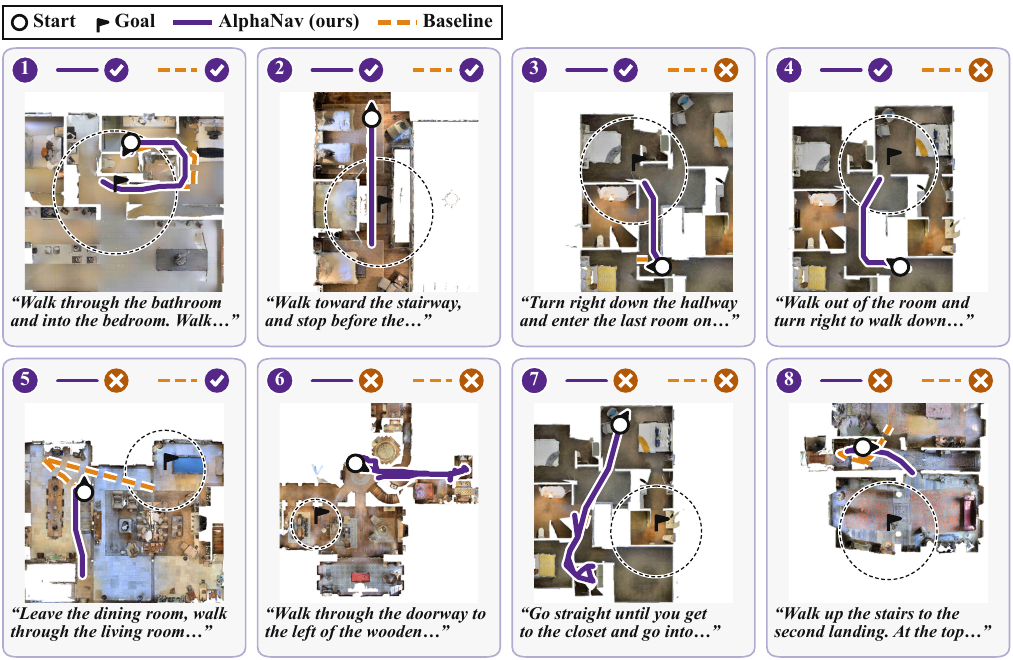}{\textwidth}
  \caption{Further closed-loop paths of \paper and the baseline recipe on unseen buildings. Ticks and crosses mark success and failure.}
  \label{fig:gallery}
\end{qfigure*}

\section{On-policy correction}
\label{sec:suppl_onpolicy}

\begin{figure}[t]
  \centering
  \figinc{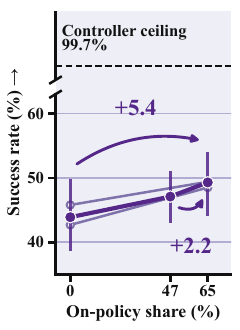}{0.52\columnwidth}
  \caption{Success rate against the on-policy share on unseen buildings. The thick line plots seed 42 after imitation, the round-level schedule, and the within-episode schedule, and the thin lines plot seeds 43 and 44 after imitation and the within-episode schedule. Bars denote 95\% intervals, and arrows mark the paired gains of the within-episode schedule over imitation and over the round-level schedule. The dashed line marks the controller with oracle route bearings and an oracle stop within 1.5\,m of the goal.}
  \label{fig:dose}
\end{figure}

This section reports how success changes with the share of on-policy rollout steps for all three seeds and illustrates the correction on single episodes.
Fig.~\ref{fig:dose} plots the success of the three seeds on unseen buildings against the on-policy share, which is the fraction of rollout steps in which the policy selects the action.
The seed-42 values match those of Table~\ref{tab:ablation}a, and Table~\ref{tab:seeds_onpolicy} lists the SR of each seed before and after correction.
When the inference settings are added, the full recipe raises the SR of the three seeds by \rdFullA, \rdFullB, and \rdFullC\ points over imitation.

\begin{figure}[t]
  \centering
  \figinc{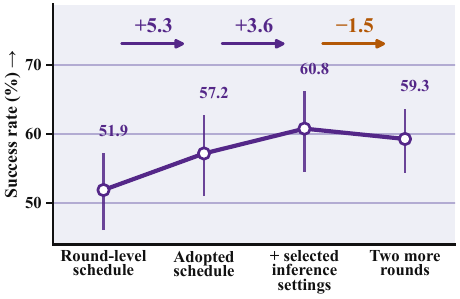}{\columnwidth}
  \caption{Success rate by recipe step on seen buildings. The steps are the round-level schedule, the within-episode schedule (adopted schedule), the same schedule with the selected inference settings, and two further rounds. Arrows mark paired differences between consecutive steps, and bars denote 95\% intervals.}
  \label{fig:ladder}
\end{figure}

\begin{table}[t]
  \centering
  \footnotesize
  \setlength{\tabcolsep}{4pt}
  \begin{tabular}{@{}>{\tintL}lcccc<{\tintR}@{}}
    \toprule
    Seed & Imitation & On-policy & Difference & Final \\
    \midrule
    42 & \rDStageOneSR & \rDVTwoSR & \rdDoseOne\ [\rdDoseOneLo, \rdDoseOneHi] & \rFinSRA \\
    43 & \rDbStageOneSR & \rDbVTwoSR & \rdDoseTwo\ [\rdDoseTwoLo, \rdDoseTwoHi] & \rFinSRB \\
    44 & \rDcStageOneSR & \rDcVTwoSR & \rdDoseThree\ [\rdDoseThreeLo, \rdDoseThreeHi] & \rFinSRC \\
    \bottomrule
  \end{tabular}
  \caption{SR per seed on unseen buildings, in percent. Imitation and On-policy: stop threshold 0.7, every frame appended to the memory (Table~\ref{tab:ablation}a, rows 1 and 3). Final: stop threshold 0.9 and memory spacing.}
  \label{tab:seeds_onpolicy}
\end{table}
On all 778 episodes of the selection set (Fig.~\ref{fig:ladder}), success peaks at \rDevVTwoSR\% with three rounds of the within-episode schedule and the selected inference settings (stop threshold 0.9 and memory frames every 0.5\,m).
Two further rounds then reach \rDevVThreeSR\% (\rdDvThreeDev\ [\rdDvThreeDevLo, \rdDvThreeDevHi] points).

Two figures illustrate how the correction acts on single episodes.
Fig.~\ref{fig:onpolicy} shows one training episode, drawn at random with a fixed seed among the five training episodes whose first rollout needs a correction and whose later rollout stays within 1\,m of the reference path.
Fig.~\ref{fig:mistakes} follows seed 42, with a stop threshold of 0.7, after imitation and after the round-level and within-episode on-policy schedules.
It shows three unseen-building episodes with different outcomes, drawn with a fixed seed.
The within-episode schedule succeeds in the first and third episodes, and imitation alone succeeds in the second.

\begin{qfigure*}[t]
  \centering
  \figinc{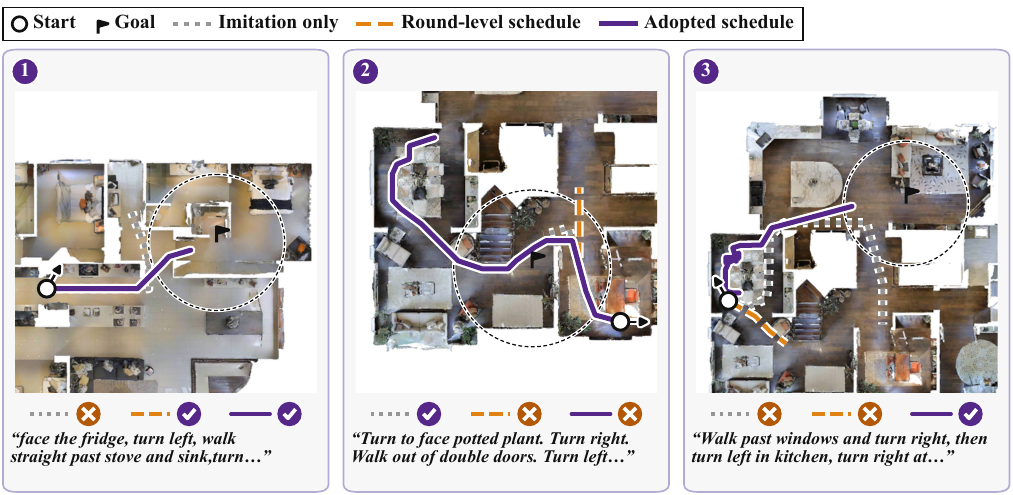}{\textwidth}
  \caption{Closed-loop paths of one \paper seed by training schedule on unseen buildings. The adopted schedule is the within-episode schedule. Ticks and crosses mark success and failure.}
  \label{fig:mistakes}
\end{qfigure*}

\section{Route-bearing output}
\label{sec:suppl_interface}

This section compares the route bearing with two image-plane readouts across start headings and shows why the decoding window does not change the actions.
Besides the scan readout of the main paper, a second image-plane readout turns toward a route out of view with a displacement head, which has a loss weight of zero and therefore receives no direct supervision; it reaches \rXFinPixSR\% SR and \rXFinPixSPL\% SPL on unseen buildings.
Table~\ref{tab:heading} lists the success rate in four bins of the angle between the start heading and the first route step.
It also lists the share of starts beyond $90^\circ$ whose first turn brings the route within $90^\circ$ of the heading.
The route bearing leads the scan readout in all four bins.
Its first turn faces the route in \rHFinARot\% of the starts beyond $90^\circ$, compared with \rHXFinScanRot\% for the scan and \rHXFinPixRot\% for the displacement head.
In most of its decisions, the displacement-head readout predicts that the route is not visible, and it therefore hands these decisions to its unsupervised head.
With the weights of the baseline recipe, the route bearing gains \rdReadAZeroF\ [\rdReadAZeroFLo, \rdReadAZeroFHi] points over the displacement head (\rAZeroFSR\% against \rXAZeroFPixSR\%).
Fig.~\ref{fig:interface} presents both outputs at the first decision of three episodes that start behind the agent.
Fig.~\ref{fig:heat} compares the waypoint distributions that three models predict on the same views.

\paragraph{Bin count and decoding window.}\label{sec:suppl_bins}
The bin count matches the turn primitive of the benchmark, and the decoding window leaves every logged action unchanged.
Every model uses 24 bins of $15^\circ$, one bin per turn primitive, and a decoded bearing therefore turns the agent by whole primitives (Sec.~\ref{sec:heads}).
Positive angles point left, bin $\nbins/2$ points forward, and bin 0 points backward.
A circular mean over the most likely bin and at most its two neighbors stays within half a bin ($7.5^\circ$) of the center of that bin, because the center carries the largest weight.
The two-bin and three-bin means therefore produce the same number of $15^\circ$ turns as the most likely bin alone.
The decoding window can change the controller output only through the $40^\circ$ cone of the four-move commit, namely when the policy predicts a visible route and the most likely bin is centered at $\pm45^\circ$.
On unseen buildings, this condition holds for \rDecAmbig\% of the non-stop decisions of the three seeds.
The three-bin mean moves none of these decisions inside the cone (\rDecChanged\%), which confirms that decoding the most likely bin alone produces the same action at every logged decision.
In the open-loop test of Fig.~\ref{fig:openloop}, the center of the most likely bin lies on average \rOLFinNinetyDev$^\circ$ from the target bearing at the training field of view, which is less than one bin width.

\begin{table*}[t]
  \centering
  \footnotesize
  \setlength{\tabcolsep}{5pt}
  \begin{tabular}{@{}>{\tintL}lccccc<{\tintR}@{}}
    \toprule
    & \multicolumn{4}{c}{SR (\%) by angle between start heading and first route step} & First turn \\
    \cmidrule(lr){2-5}
    Model & 0--45$^\circ$ & 45--90$^\circ$ & 90--135$^\circ$ & 135--180$^\circ$ & within $90^\circ$ (\%) \\
    \midrule
    Episodes & \rHFinANa& \rHFinANb& \rHFinANc& \rHFinANd& -- \\
    \groupband{6}{Same weights (\paper, seed 42), three readouts}
    \oursrow \oursname{Route bearing (\paper)} & \best{\rHFinASRa}& \best{\rHFinASRb}& \best{\rHFinASRc}& \best{\rHFinASRd}& \best{\rHFinARot} \\
    Image-plane waypoint, scan & \second{\rHXFinScanSRa}& \second{\rHXFinScanSRb}& \second{\rHXFinScanSRc}& \second{\rHXFinScanSRd}& \second{\rHXFinScanRot} \\
    Image-plane waypoint, displacement head & \rHXFinPixSRa& \rHXFinPixSRb& \rHXFinPixSRc& \rHXFinPixSRd& \rHXFinPixRot \\
    \groupband{6}{Models trained with and without the augmentation (one seed each, stop threshold 0.6)}
    Baseline recipe            & \rHAZeroFSRa& \second{\rHAZeroFSRb}& \second{\rHAZeroFSRc}& \second{\rHAZeroFSRd}& \second{\rHAZeroFRot} \\
    + orient-then-go           & \best{\rHAOneFSRa}  & \best{\rHAOneFSRb}  & \best{\rHAOneFSRc}  & \best{\rHAOneFSRd}  & \best{\rHAOneFRot} \\
    + truncated-memory training & \second{\rHAFourFSRa}& \rHAFourFSRb& \rHAFourFSRc& \rHAFourFSRd& \rHAFourFRot \\
    \bottomrule
  \end{tabular}
  \caption{Success rate by start heading on unseen buildings. First turn: starts beyond $90^\circ$ whose first rotation brings the route within $90^\circ$. Bold/underline: best/second per column and block.}
  \label{tab:heading}
\end{table*}

\begin{qfigure*}[t]
  \centering
  \makebox[\textwidth]{\figinc{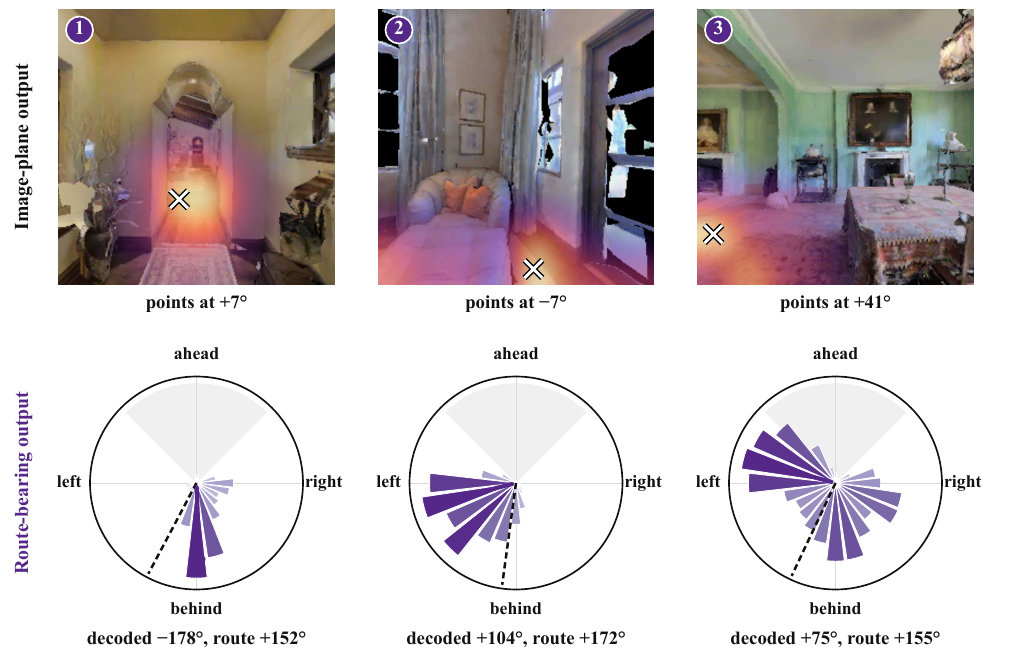}{0.7\textwidth}}
  \caption{The two outputs at the first decision of three unseen-building episodes whose route starts behind the agent. The top row (image-plane output) overlays the predicted waypoint distribution on the camera view, with its peak as a cross and its image-plane direction below. The bottom row plots the 24-bin route-bearing distribution with the field of view as a grey sector and the direction of the first reference-path step as a dashed line, and the label gives the decoded bearing and the route direction.}
  \label{fig:interface}
\end{qfigure*}

\begin{qfigure*}[t]
  \centering
  \figinc{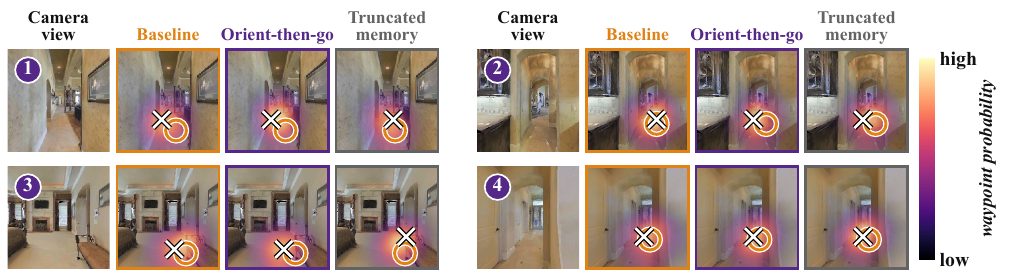}{\textwidth}
  \caption{Waypoint distributions of three models on the same four views of seen buildings. The models are the baseline recipe, the orient-then-go model, and the truncated-memory model. Crosses mark the predicted peak, and rings mark the reference waypoint.}
  \label{fig:heat}
\end{qfigure*}

\section{Label-preserving augmentations}
\label{sec:suppl_twins}

The two augmentations of Sec.~\ref{sec:suppl_aug} keep their labels exact under the following conditions.
The mirror augmentation requires a principal point on the center column, which holds for every generated frame.
It also requires a swap of every directional left and right word, while compass words and the adverbial use of ``left'' or ``right'', as in ``stop right before the door'', keep their form.
In addition, the swap must act as an involution, meaning that a second application restores the original instruction.
An automated check over the 57{,}513 training instructions, taken as the model receives them, finds a directional word in 50{,}496 instructions (87.8\%).
Applying the swap twice restores 50{,}495 of these exactly and leaves the remaining instructions identical.
The one exception is an \rxr instruction that contains the spelling ``leFt'', which returns with a lower-case ``f''.
The orient-then-go augmentation applies to episodes with a uniformly drawn start heading and inserts its clause only when the required rotation is at least $45^\circ$.

\paragraph{Direction words.}\label{sec:suppl_words}
The instruction sensitivity measures how often the first turn follows the direction words of the instruction.
It is the share of first turns that flip when we swap the left and right words of the first clause while the scene stays unchanged, measured on the episodes with a directional word in that clause.
The baseline recipe flips \rSAZeroF\ of its first turns, and the orient-then-go model flips \rSAOneF.
On these \rSEligAZeroF\ episodes, the swap lowers SR by \rSDropAZeroF\ points for the baseline recipe and by \rSDropAOneF\ points for the orient-then-go model, so the direction words change both the first turn and the outcome.

\section{Accuracy and cost}
\label{sec:suppl_main}

This section extends the accuracy comparison of the main paper to further published systems and to RxR-CE, and it analyzes the cost of each decision.

\paragraph{Further published systems.}\label{sec:suppl_more}
Table~\ref{tab:main_more} lists the map-based and zero-shot systems that complete Table~\ref{tab:main}.
Each of these systems stays below 40\% SR, compared with \rFinSR\% for \paper.
Table~\ref{tab:more_recent} lists the data regimes of five systems that differ in memory, training regime, or interface; MemVLN and JOP-VLN also appear in Table~\ref{tab:main}.
The streaming MemVLN (4B) trains on the R2R and RxR trajectories alone and reaches 55.0\% SR.
JOP-VLN (8B) adds EnvDrop, a ScaleVLN subset, DAgger, and group relative policy optimization (GRPO) and reaches 69.9\%.
ETP-R1 and LCGNav plan over a topological graph from panoramic RGB-D input, and the panoramic waypoint models of Krantz~\etal reach 36\% SR with reinforcement learning.

\begin{table}[t]
  \centering
  \footnotesize
  \setlength{\tabcolsep}{3pt}
  \resizebox{\columnwidth}{!}{%
  \begin{tabular}{@{}lllcccc@{}}
    \toprule
    Method & Group & Obs. & NE\,$\downarrow$ & OSR\,$\uparrow$ & SR\,$\uparrow$ & SPL\,$\uparrow$ \\
    \midrule
    CM2~\cite{georgakis2022cm2} & Map / waypoint & RGB-D & \rLitCMTwoNE & \rLitCMTwoOSR & \rLitCMTwoSR & \rLitCMTwoSPL \\
    LAW~\cite{raychaudhuri2021law} & Map / waypoint & RGB-D & \second{\rLitLAWNE} & \rLitLAWOSR & \second{\rLitLAWSR} & \second{\rLitLAWSPL} \\
    WS-MGMap~\cite{chen2022wsmgmap} & Map / waypoint & RGB-D & \best{\rLitWSMGMapNE} & \second{\rLitWSMGMapOSR} & \best{\rLitWSMGMapSR} & \best{\rLitWSMGMapSPL} \\
    CA-Nav~\cite{chen2024canav} & Zero-shot planner & RGB-D & \rLitCANavNE & \best{\rLitCANavOSR} & \rLitCANavSR & \rLitCANavSPL \\
    AO-Planner~\cite{chen2024aoplanner} & Zero-shot planner & RGB-D, 4 views & \rLitAOPlannerNE & \rLitAOPlannerOSR & \rLitAOPlannerSR & \rLitAOPlannerSPL \\
    InstructNav~\cite{long2024instructnav} & Zero-shot planner & RGB-D, multi-view & \rLitInstructNavNE & \na & \rLitInstructNavSR & \rLitInstructNavSPL \\
    DreamNav$^{\S}$~\cite{wang2025dreamnav} & Zero-shot planner & RGB-D & \rLitDreamNavNE & \rLitDreamNavOSR & \rLitDreamNavSR & \rLitDreamNavSPL \\
    \bottomrule
  \end{tabular}}
  \caption{Map-based and zero-shot systems on \rtor \valunseen, from the cited papers. $^{\S}$613 trajectories. Bold/underline: best/second.}
  \label{tab:main_more}
\end{table}

\begin{table*}[t]
  \centering
  \footnotesize
  \setlength{\tabcolsep}{3.5pt}
  \begin{tabular}{@{}>{\tintL}lll>{\raggedright\arraybackslash}p{6.4cm}cccc<{\tintR}@{}}
    \toprule
    Method & LLM & Obs. & Training data and regime & NE\,$\downarrow$ & OSR\,$\uparrow$ & SR\,$\uparrow$ & SPL\,$\uparrow$ \\
    \midrule
    Waypoint models (HPN+DN)~\cite{krantz2021waypoint} & -- & pano.\ RGB-D & \rtor; reinforcement learning (DD-PPO) & 6.31 & 40 & 36 & 34 \\
    MemVLN~\cite{wang2026memvln} & 4B & RGB & R2R and RxR trajectories; imitation with a trajectory-summary task & 5.19 & 61.2 & 55.0 & 50.5 \\
    ETP-R1~\cite{ye2025etpr1} & -- & pano.\ RGB-D & Prevalent and its Gemini-annotated version, RxR-Marky, R2R, RxR; online DAgger, then GRPO & 3.94 & 72 & 65 & 56 \\
    LCGNav on ETP-R1~\cite{peng2026lcgnav} & -- & pano.\ RGB-D & ETP-R1 frozen; a point-cloud plug-in trained for at most 1{,}000 iterations & 3.93 & 71 & 65 & 57 \\
    JOP-VLN~\cite{he2026jopvln} & 8B & RGB & R2R, RxR, EnvDrop, ScaleVLN subset; imitation, DAgger, then GRPO mixed with imitation & 3.64 & 76.4 & 69.9 & 64.9 \\
    \midrule
    \oursrow \oursname{\paper (ours)} & \ours{2B} & \ours{RGB} & \rtor, \rxr, EnvDrop; imitation and on-policy correction & \ours{\rFinNE} & \ours{\rFinOSR} & \ours{\rFinSR} & \ours{\rFinSPL} \\
    \bottomrule
  \end{tabular}
  \caption{Further published systems on \rtor \valunseen, from the cited papers (some report whole percentages). \paper: mean of three seeds. LLM: language-model size (--: none; ETP-R1 uses a 0.5B planner). pano.: panoramic; HPN+DN: heading prediction and distance networks; DD-PPO: decentralized distributed proximal policy optimization.}
  \label{tab:more_recent}
\end{table*}

\paragraph{RxR-CE.}\label{sec:suppl_rxr}
We evaluate the three seeds of \paper in Table~\ref{tab:main} on the \rRxN\ English guide episodes of \rxr \valunseen~\cite{ku2020rxr,krantz2020vlnce}.
English is the language of the training instructions (en-US and en-IN; \rRxWords\ words per instruction on average), and we do not evaluate the Hindi and Telugu episodes.
The settings are those selected on \rtor \valseen, without any change or tuning on \rxr.
English \rxr training episodes are part of the imitation data (Sec.~\ref{sec:suppl_data}), and the on-policy rounds use \rtor alone.
They comprise a $90^\circ$ field of view, 500 primitives per episode, and success within 3\,m by geodesic distance.
We evaluate each seed once, and \paper reaches \rRxFinSR\,$\pm$\,\rRxFinSRSd\% SR (interval [\rRxFinSRmeanLo, \rRxFinSRmeanHi]) and \rRxFinSPL\% SPL (Table~\ref{tab:rxr_seeds}).
This SR lies \rRxDeltaRtor\ points below its SR on \rtor.
Among the 2B systems with published \rxr results (Table~\ref{tab:rxr}), it lies \rRxGapVLNR\ points above VLN-R1 (\rRxLitVLNRoneSR\%) and \rRxGapGoalTwoPixel\ and \rRxGapMonoDream\ points below Goal2Pixel (\rRxLitGoalTwoPixelSR\%) and MonoDream (\rRxLitMonoDreamSR\%).
By locale, SR is \rRxLangUS\% on the \rRxLangUSN\ en-US episodes and \rRxLangIN\% on the \rRxLangINN\ en-IN episodes.
The imitation-only model of seed 42 (stage 1) reaches \rRxStageOneSR\% SR, and the final model lies \rdRxRecipe\ points above it (interval \rdRxRecipeLo\ to \rdRxRecipeHi).
Because the two runs use different inference rules, with stop threshold 0.7 and no memory spacing for the imitation-only model and stop threshold 0.9 and 0.5\,m spacing for the final model, this gain combines the effects of on-policy training and the inference rule.
The weights and the rule of seed 42 also decode with the image-plane readout that scans in place, which reaches \rRxScanSR\% SR and \rRxScanSPL\% SPL.
The route bearing therefore exceeds the scan by \rdRxScan\ [\rdRxScanLo, \rdRxScanHi] points on \rxr, compared with \rdReadFinScan\ points on \rtor.
The instructions average \rRxTokInstr\ tokens, so a decision processes \rRxTokSeq\ tokens on average, compared with \rTokSeqMean\ on \rtor.
At the median English instruction, a decision takes \rRxDecMs\,ms and \rRxInferGB\,GiB, measured as in Sec.~\ref{sec:suppl_measure}, and the closed loop takes \rRxLoopMs\,ms per decision over \rRxDecEp\ decisions per episode on average.
In every run, \rRxUnreach\ episodes (\rRxUnreachPct\%) in one building have a goal that is unreachable on our navigation mesh.
SR, OSR, and SPL count these episodes as not reached, and NE averages the other episodes, on which the SR is \rRxReachSR\%.
We compute normalized dynamic time warping (nDTW), which measures how closely the path follows the reference path, as in VLN-CE, and we reconstruct the path per move from the decision-level positions.
None of the compared papers states which language subset its \rxr table covers, and we list their values as reported.

\begin{table*}[t]
  \centering
  \footnotesize
  \setlength{\tabcolsep}{3.5pt}
  \begin{tabular}{@{}>{\tintL}lllccccc<{\tintR}@{}}
    \toprule
    Method & LLM & Obs. & NE\,$\downarrow$ & OSR\,$\uparrow$ & SR\,$\uparrow$ & SPL\,$\uparrow$ & nDTW\,$\uparrow$ \\
    \midrule
    \groupband{8}{2B language model, RGB}
    MonoDream~\cite{wang2025monodream} & 2B & RGB & \rRxLitMonoDreamNE & \rRxLitMonoDreamOSR & \rRxLitMonoDreamSR & \rRxLitMonoDreamSPL & \na \\
    Goal2Pixel$^{\ddagger}$~\cite{goal2pixel2026} & 2B & RGB$^{\ddagger}$ & \rRxLitGoalTwoPixelNE & \na & \rRxLitGoalTwoPixelSR & \rRxLitGoalTwoPixelSPL & \rRxLitGoalTwoPixelNDTW \\
    \oursrow \oursname{\paper (ours)} & \ours{2B} & \ours{RGB} & \ours{\rRxFinNE} & \ours{\rRxFinOSR} & \ours{\rRxFinSR} & \ours{\rRxFinSPL} & \ours{\rRxFinNDTW} \\
    VLN-R1~\cite{qi2025vlnr1} & 2B & RGB & \rRxLitVLNRoneNE & \rRxLitVLNRoneOSR & \rRxLitVLNRoneSR & \rRxLitVLNRoneSPL & \na \\
    \groupband{8}{3B to 8B language model, RGB}
    Efficient-VLN (v2)~\cite{zheng2025efficientvln} & 8B & RGB & \rRxLitEffVLNTwoNE & \na & \rRxLitEffVLNTwoSR & \rRxLitEffVLNTwoSPL & \na \\
    Qwen-RobotNav~\cite{qwenrobotnav2026} & 4B & RGB & \rRxLitQwenRobotNavNE & \na & \rRxLitQwenRobotNavSR & \rRxLitQwenRobotNavSPL & \rRxLitQwenRobotNavNDTW \\
    Efficient-VLN (v1)~\cite{zheng2025efficientvln} & 3B+1B & RGB & \rRxLitEffVLNOneNE & \na & \rRxLitEffVLNOneSR & \rRxLitEffVLNOneSPL & \rRxLitEffVLNOneNDTW \\
    DualVLN~\cite{wei2025dualvln} & 7B & RGB & \rRxLitDualVLNNE & \na & \rRxLitDualVLNSR & \rRxLitDualVLNSPL & \rRxLitDualVLNNDTW \\
    StreamVLN + ScaleVLN~\cite{wei2025streamvln} & 7B & RGB & \rRxLitStreamVLNScaleNE & \na & \rRxLitStreamVLNScaleSR & \rRxLitStreamVLNScaleSPL & \rRxLitStreamVLNScaleNDTW \\
    Aux-Think~\cite{wang2025auxthink} & 8B & RGB & \rRxLitAuxThinkNE & \rRxLitAuxThinkOSR & \rRxLitAuxThinkSR & \rRxLitAuxThinkSPL & \na \\
    NaVILA~\cite{cheng2025navila} & 8B & RGB & \rRxLitNaVILANE & \na & \rRxLitNaVILASR & \rRxLitNaVILASPL & \rRxLitNaVILANDTW \\
    Uni-NaVid~\cite{zhang2024uninavid} & 7B & RGB & \rRxLitUniNaVidNE & \rRxLitUniNaVidOSR & \rRxLitUniNaVidSR & \rRxLitUniNaVidSPL & \na \\
    StreamVLN~\cite{wei2025streamvln} & 7B & RGB & \rRxLitStreamVLNNE & \na & \rRxLitStreamVLNSR & \rRxLitStreamVLNSPL & \rRxLitStreamVLNNDTW \\
    \groupband{8}{Depth input}
    GA-VLN~\cite{yang2026gavln} & 7B+1B & RGB-D & \rRxLitGAVLNNE & \rRxLitGAVLNOSR & \rRxLitGAVLNSR & \rRxLitGAVLNSPL & \na \\
    InternVLA-N1~\cite{internvlan1_2025} & 7B & RGB-D & \rRxLitInternVLANOneNE & \na & \rRxLitInternVLANOneSR & \rRxLitInternVLANOneSPL & \rRxLitInternVLANOneNDTW \\
    MapNav~\cite{zhang2025mapnav} & 7B & RGB-D & \rRxLitMapNavNE & \na & \rRxLitMapNavSR & \rRxLitMapNavSPL & \rRxLitMapNavNDTW \\
    VLN-3DFF~\cite{wang2024vln3dff} & -- & RGB-D & \rRxLitVLNThreeDFFNE & \rRxLitVLNThreeDFFOSR & \rRxLitVLNThreeDFFSR & \rRxLitVLNThreeDFFSPL & \na \\
    \bottomrule
  \end{tabular}
  \caption{Systems on \rxr \valunseen, from the cited papers as reported (\na: not reported). \paper: English subset, mean of three seeds. LLM: language-model size. $^{\ddagger}$Depth turns the image goal into a waypoint. Groups are ordered by SR.}
  \label{tab:rxr}
\end{table*}

\begin{table}[t]
  \centering
  \footnotesize
  \setlength{\tabcolsep}{3pt}
  \begin{tabular}{@{}>{\tintL}lccccc<{\tintR}@{}}
    \toprule
    Model & NE\,$\downarrow$ & OSR\,$\uparrow$ & SR\,$\uparrow$ & SPL\,$\uparrow$ & nDTW\,$\uparrow$ \\
    \midrule
    Seed 42 & \rRxFinNEA& \rRxFinOSRA& \rRxFinSRA& \rRxFinSPLA& \rRxFinNDTWA \\
    Seed 43 & \rRxFinNEB& \rRxFinOSRB& \rRxFinSRB& \rRxFinSPLB& \rRxFinNDTWB \\
    Seed 44 & \rRxFinNEC& \rRxFinOSRC& \rRxFinSRC& \rRxFinSPLC& \rRxFinNDTWC \\
    \midrule
    \oursrow \oursname{Mean} & \rRxFinNE & \rRxFinOSR & \rRxFinSR & \rRxFinSPL & \rRxFinNDTW \\
    \oursrow \oursname{Std.} & \rRxFinNESd & \rRxFinOSRSd & \rRxFinSRSd & \rRxFinSPLSd & \rRxFinNDTWSd \\
    \midrule
    Imitation only (42) & \rRxStageOneNE & \rRxStageOneOSR & \rRxStageOneSR & \rRxStageOneSPL & \rRxStageOneNDTW \\
    Scan readout (42) & \rRxScanNE & \rRxScanOSR & \rRxScanSR & \rRxScanSPL & \rRxScanNDTW \\
    \bottomrule
  \end{tabular}
  \caption{Results per seed on \rxr \valunseen, English subset. NE in meters, others in percent. Std.: sample standard deviation. Imitation only: stage 1 of seed 42. Scan readout: seed 42 decoded as an image-plane waypoint that scans.}
  \label{tab:rxr_seeds}
\end{table}

\paragraph{Closed loop and motion per decision.}\label{sec:suppl_cost}
The closed-loop cost depends on how many decisions an episode takes and on how far each decision moves the agent.
Over the three seeds on unseen buildings, an episode takes \rMeanDecisions\ decisions on average, with a median of \rMedDecisions.
A decision executes \rPrimsPerDec\ policy primitives on average, which amounts to \rMsPerPrim\,ms of isolated decision time per executed primitive.
Including the controller, simulator steps, and rendering, the loop takes \rLoopMs\,ms per decision, measured over the \rLoopN\ decisions of the episodes that stay in the building of the previous episode.
A decision moves the agent \rMoveMean\,m horizontally on average and at most \rMoveMax\,m, and \rMoveZero\% of the decisions leave its position unchanged.
At 1\,m/s, the mean motion takes \rBudgetIso\ times the isolated decision latency and \rBudgetLoop\ times the closed-loop time per decision.

\paragraph{Controller cases.}
The controller cases show how often each controller rule acts and how often its forward moves collide.
Of the \rCtlDecN\ policy decisions of the three seeds, \rCtlCommit\% commit four moves toward a route that is predicted visible within the $40^\circ$ cone, and \rCtlTurn\% only turn.
In addition, \rCtlTwo\% take two moves because the agent already faces a route that is predicted out of view (the hidden-route rule), \rCtlGuard\% consist of moves of the turn-loop guard, and \rCtlStop\% stop.
The commits account for \rCtlCommitMoves\% of all forward moves, with one forward pass each.
Of the commits, \rCtlCommitFull\% execute all four moves, and a collision ends the others.
Of the commit moves, \rCtlCommitFree\% are collision-free.
A forward move collides in \rCtlCommitColl\% of the commit moves, compared with \rCtlTwoColl\% of the moves of the hidden-route rule and \rCtlGuardColl\% of the guard moves.
The controller takes both of these shorter move types when no visible route lies within the cone.
The long commit therefore occurs where the route is in view, and the short steps occur where the route is out of view.
In \rCtlShort\% of all forward moves, the agent travels less than 0.23\,m because it slides along an obstacle, and the next decision processes a new frame at most \rMoveMax\,m later.
With oracle route bearings and an oracle stop within 1.5\,m, the same commits reach \rXCeilSR\% SR (Sec.~\ref{sec:exp_dagger}).
On the 300 selection episodes of the inference batch, with seed 42, stop threshold 0.7, and every frame appended, the hidden-route rule reaches \rDevInfBaseSR\% SR with two moves and \rDevInfHiddenOneSR\% with one move.

\paragraph{Cost normalized by tokens and FLOPs.}\label{sec:suppl_norm}
Table~\ref{tab:norm} lists the quantities that describe the cost of one call independently of the GPU, in particular tokens and floating-point operations (FLOPs).
On the 1{,}839 unseen-building instructions in bfloat16, a decision of \paper processes \rTokSeqMean\ tokens on average.
These tokens comprise \rTokImage\ image-block tokens, namely \rVisTokens\ visual tokens with their start and end markers, and \rTokInstMean\ instruction tokens, with a median of \rTokInstMed\ and a maximum of \rTokInstMax.
We estimate the compute analytically as $2\times$ parameters $\times$ tokens per module.
The estimate counts \rParVision\,M vision-tower parameters over 768 image patches before the $2\times2$ merge and \rParDecoder\,M decoder parameters over the sequence.
It yields \rTflopsOurs\ TFLOPs per decision on average, including \rTflopsVision\ TFLOPs in the vision tower, and \rTflopsOursMax\ TFLOPs for the longest instruction.
This estimate excludes attention over the sequence and the heads.
At the measured decision latency, this load corresponds to \rTokPerSec\ tokens and an estimated \rTflopsPerSec\ TFLOPs per second on the RTX~5070.
GA-VLN reports \rNormTflopsGAVLN\ theoretical TFLOPs per inference, which is \rFlopRatioGAVLN\ times our estimate.
The same rule applied to the 8B NaVILA, with its \rNormCtxNaVILA\ context tokens and \rNormGenNaVILA\ generated tokens, yields \rNormTflopsNaVILA\ TFLOPs, which is \rFlopRatioNaVILA\ times our estimate.
The other papers report only frames or tokens per frame.

\begin{table*}[t]
  \centering
  \footnotesize
  \setlength{\tabcolsep}{3.2pt}
  \begin{tabular}{@{}>{\tintL}ll>{\raggedright\arraybackslash}p{5.6cm}cllr<{\tintR}@{}}
    \toprule
    Method & LLM & Tokens per call (as stated) & TFLOPs per call & Precision & GPU & ms/call (actions) \\
    \midrule
    NaVid~\cite{zhang2024navid} & 7B & 64 for the current frame, 4 per history frame & -- & -- & -- & \rLatNaVidMs\ (1) \\
    Uni-NaVid~\cite{zhang2024uninavid} & 7B & 64 current, 4 per short-term frame (64-frame buffer), 1 per long-term frame & -- & -- & A100 & $\sim$\rLatUniNaVidMs\ (4) \\
    NaVILA~\cite{cheng2025navila} & 8B & \rNormCtxNaVILA\ context, \rNormGenNaVILA\ generated &\rNormTflopsNaVILA$^{b}$ & FP16 & RTX~4090 & \rLatNaVILAMs\ (1) \\
    StreamVLN~\cite{wei2025streamvln} & 7B & memory of 8 frames $\times$ 196, window of 8 turns & -- & -- & RTX~4090 & \rLatStreamVLNMs\ (4) \\
    MonoDream~\cite{wang2025monodream} & 2B & 8 history frames and the current frame & -- & -- & RTX~4090 & \rLatMonoDreamMs\ (step) \\
    Goal2Pixel~\cite{goal2pixel2026} & 2B & up to 8 history images and the current frame & -- & -- & -- & \rLatGoalTwoPixelMs\ (pixel goal) \\
    GA-VLN~\cite{yang2026gavln} & 7B+1B & BEV of up to 8 history frames and the current view & \rNormTflopsGAVLN &-- & -- & \rLatGAVLNMs\ (--) \\
    Efficient-VLN v2~\cite{zheng2025efficientvln} & 8B & 16 frames at $512\times512$ & -- & FP16 & H200 & \rLatEffVLNTwoMs\ (4) \\
    Qwen-RobotNav~\cite{qwenrobotnav2026} & 4B & visual budget of 2{,}048--4{,}096 in training & -- & FP8$^{c}$ & --$^{c}$ & \rLatQwenRobotNavMs\ (8 waypoints) \\
    \midrule
    \oursrow \oursname{\paper (ours)} & \ours{2B} & \rTokSeqMean\ on average (\rVisTokens\ visual), at most \rTokSeqMax & \rTflopsOurs$^{a}$ & bfloat16 & RTX~5070 & \rDecMs\ (\rPrimsPerDec) \\
    \bottomrule
  \end{tabular}
  \caption{One policy call in tokens and FLOPs, as each paper states them (--: not stated). Parentheses: primitives or actions per call; for \paper, mean executed primitives. BEV: bird's-eye view. $^{a}$Analytic estimate, $2\times$ parameters $\times$ tokens per module. $^{b}$Rule $^{a}$ with 8B parameters and $\rNormCtxNaVILA+\rNormGenNaVILA$ tokens. $^{c}$FP8 on an onboard Jetson Thor; the timed remote server's GPU is not stated.}
  \label{tab:norm}
\end{table*}

\section{Robustness}
\label{sec:suppl_robust}

This section tests the robustness of the policy to the training seed, to route length and building, and to new cameras.
The SR of the three seeds has a standard deviation of \rFinSRSd\ points, and the policy runs on new cameras without retraining.
On seen buildings, the policy retains \rCamKeepPitchDev\% of its SR at a pitch of $-30^\circ$ and \rCamKeepSixtyDev\% at a $60^\circ$ field of view, each $20^\circ$ outside its training range.
At $120^\circ$, which lies $20^\circ$ beyond the widest training camera, the policy retains \rCamKeepTwentyDev\%.

\begin{figure}[t]
  \centering
  \begin{subfigure}[t]{0.49\columnwidth}
    \centering
    \figinc{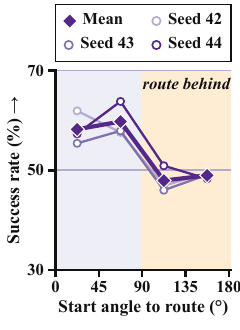}{\linewidth}
    \caption{Success rate by start heading.}
    \label{fig:headingseeds_a}
  \end{subfigure}
  \hfill
  \begin{subfigure}[t]{0.49\columnwidth}
    \centering
    \figinc{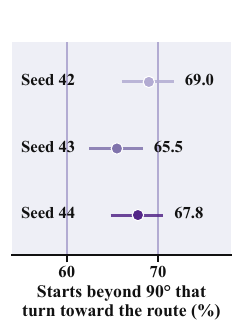}{\linewidth}
    \caption{First turn toward the route.}
    \label{fig:headingseeds_b}
  \end{subfigure}
  \caption{Start heading per seed on unseen buildings. Panel (b) plots the share of starts beyond $90^\circ$ whose first rotation brings the route within $90^\circ$ of the heading, with 95\% Wilson intervals.}
  \label{fig:headingseeds}
\end{figure}

\paragraph{Seeds.}\label{sec:suppl_seeds}
Each seed trains stage 1, the initial imitation training, and the on-policy rounds from its own initialization.
We evaluate each seed once on all 1{,}839 unseen-building episodes.
The seeds reach \rFinSRA, \rFinSRB, and \rFinSRC\% SR with a standard deviation of \rFinSRSd\ points (Table~\ref{tab:perseed}).
Every seed performs worse on starts beyond $90^\circ$ than on the other starts (Fig.~\ref{fig:headingseeds}).
For the three seeds, the first turn faces the route in \rHFinARot\%, \rHFinBRot\%, and \rHFinCRot\% of the starts beyond $90^\circ$.

\begin{table}[t]
  \centering
  \footnotesize
  \setlength{\tabcolsep}{3pt}
  \begin{tabular}{@{}>{\tintL}lcccccc<{\tintR}@{}}
    \toprule
    Seed & NE\,$\downarrow$ & OSR\,$\uparrow$ & SR\,$\uparrow$ & SR 95\% interval & SPL\,$\uparrow$ & GPU-h \\
    \midrule
    42 & \rFinNEA& \rFinOSRA& \rFinSRA& \rFinSRciA& \rFinSPLA& \rCostTrainMinA \\
    43 & \rFinNEB& \rFinOSRB& \rFinSRB& \rFinSRciB& \rFinSPLB& \rCostTrainMinB \\
    44 & \rFinNEC& \rFinOSRC& \rFinSRC& \rFinSRciC& \rFinSPLC& \rCostTrainMinC \\
    \midrule
    \oursrow \oursname{Mean} & \rFinNE & \rFinOSR & \rFinSR & [\rFinSRmeanLo, \rFinSRmeanHi] & \rFinSPL & \rCostTrainTyp \\
    \oursrow \oursname{Std.} & \rFinNESd & \rFinOSRSd & \rFinSRSd & -- & \rFinSPLSd & -- \\
    \bottomrule
  \end{tabular}
  \caption{Results per seed on unseen buildings. NE in meters, others in percent. GPU-h: uninterrupted training run on one RTX~5070. Std.: sample standard deviation.}
  \label{tab:perseed}
\end{table}

\paragraph{Route length and buildings.}
We also split the comparison with the baseline recipe by route length and by building.
Against the baseline recipe, \paper gains \rdLenShort\ [\rdLenShortLo, \rdLenShortHi], \rdLenMid\ [\rdLenMidLo, \rdLenMidHi], and \rdLenLong\ [\rdLenLongLo, \rdLenLongHi] points on the short, middle, and long thirds of the routes and leads in \rSceneWinN\ of the \rSceneN\ buildings (Fig.~\ref{fig:strata}).

\begin{figure}[t]
  \centering
  \figinc{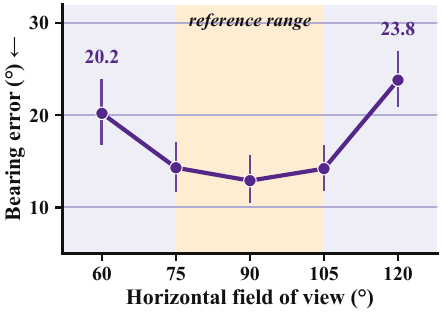}{3.0in}
  \caption{Open-loop route-bearing error of \paper against the field of view on seen buildings, with 95\% intervals. The shaded band marks the reference range from $75^\circ$ to $105^\circ$.}
  \label{fig:openloop}
\end{figure}

\paragraph{Field of view.}\label{sec:suppl_fov}
The policy receives the vector $\kap$ of the camera on which it runs.
A new camera with known intrinsics, height, and pitch therefore runs with unchanged weights.
The first row of Table~\ref{tab:camera} lists the closed-loop SR of \paper at four cameras on the 300 selection episodes.
It reaches \rCamFinalCamNomDev\% with the training camera and \rCamFinalCamPitchDev\% at a pitch of $-30^\circ$, against a training range from $-10^\circ$ to $0^\circ$.
It reaches \rCamFinalCamSixtyDev\% at a $60^\circ$ field of view, against a training range from $80^\circ$ to $100^\circ$, and \rCamFinalCamTwentyDev\% at $120^\circ$, the camera with the largest open-loop error.
In the open-loop test of Fig.~\ref{fig:openloop}, the policy predicts the route bearing on 300 decision poses of seen buildings, re-rendered at fields of view from $60^\circ$ to $120^\circ$.
The bearing error is lowest for fields of view between $75^\circ$ and $105^\circ$ and grows by \rdHfovSixtyDev\ [\rdHfovSixtyDevLo, \rdHfovSixtyDevHi] degrees at $60^\circ$ and \rdHfovTwentyDev\ [\rdHfovTwentyDevLo, \rdHfovTwentyDevHi] degrees at $120^\circ$.

\paragraph{Camera fine-tuning.}\label{sec:suppl_camera_negative}\label{sec:support}
We also test whether a short fine-tuning on pairs of camera renders improves the policy on new cameras.
Four arms fine-tune the trained policy for 1{,}500 steps at a learning rate of $2\times10^{-5}$, and every arm keeps one forward pass per decision.
The control fine-tunes on single renders, the two-view imitation arm imitates both renders of each pose, and two further arms add a Jensen-Shannon consistency term between the two route-bearing distributions, on all bins (unrestricted) or on the bins that both cameras observe (common support).
The pairs use fields of view from $70^\circ$ to $100^\circ$, heights from 1.0 to 1.5\,m, and pitch from $-10^\circ$ to $0^\circ$.
The control and the unrestricted arm raise the SR at $120^\circ$ from \rCamFinalCamTwentyDev\% to \rCamFZeroTwentyDev\% and \rCamFTwoTwentyDev\%, and every arm loses success at the other three cameras (Table~\ref{tab:camera}).
\paper therefore uses no camera fine-tuning.
Related methods condition on camera parameters with feature-wise linear modulation (FiLM)~\cite{perez2018film}, canonicalize geometry~\cite{kim2026canonnav}, use spherical range images with validity masks~\cite{rehberg2026sensorlayout}, or enforce cross-view action consistency~\cite{huang2026crossview}.

\begin{table}[t]
  \centering
  \footnotesize
  \setlength{\tabcolsep}{3pt}
  \begin{tabular}{@{}>{\tintL}lcccc<{\tintR}@{}}
    \toprule
    Arm & Nominal & hfov $60^\circ$ & hfov $120^\circ$ & pitch $-30^\circ$ \\
    \midrule
    \oursrow \oursname{\paper} & \rCamFinalCamNomDev & \rCamFinalCamSixtyDev & \rCamFinalCamTwentyDev & \rCamFinalCamPitchDev \\
    \midrule
    No pairs (control) & \best{\rCamFZeroNomDev}& \best{\rCamFZeroSixtyDev}& \best{\rCamFZeroTwentyDev}& \best{\rCamFZeroPitchDev} \\
    Two-view imitation & \second{\rCamFOneNomDev}& \second{\rCamFOneSixtyDev}& \rCamFOneTwentyDev& \rCamFOnePitchDev \\
    Unrestricted & \rCamFTwoNomDev& \rCamFTwoSixtyDev& \second{\rCamFTwoTwentyDev}& \rCamFTwoPitchDev \\
    Common support & \rCamFFourNomDev& \rCamFFourSixtyDev& \rCamFFourTwentyDev& \second{\rCamFFourPitchDev} \\
    \bottomrule
  \end{tabular}
  \caption{SR (\%) of the camera fine-tuning arms on seen buildings. First row: the starting checkpoint. Bold/underline: best/second among the arms.}
  \label{tab:camera}
\end{table}

\paragraph{Odometry and actuation.}\label{sec:suppl_odometry}
The memory rule relies on odometry through one comparison per decision, which tests whether the agent lies at least 0.5\,m from the position of the last appended frame.
Compared with appending every frame, this rule raises SR on the 300 selection episodes by \rdSpacingNineDev\ points at stop threshold 0.9 (\rMemEightSRDev\% against \rDevInfStopNineSR\%) and from \rDevInfBaseSR\% to \rDevInfSpacingSR\% at stop threshold 0.7, at the same cost per decision.
To test how sensitive the rule is to odometry errors, we replay the logged positions of the three seeds on unseen buildings.
In this replay, \rMemAppend\% of the decisions append a frame, \rMemStill\% lie less than 0.05\,m from the last appended position because of turns in place and blocked moves, and \rMemBand\% lie between 0.4 and 0.6\,m.
Scaling the measured distance by a factor between 0.9 and 1.1 leaves the outcome of the test unchanged for at least \rMemKeepTen\% of the logged decisions, and a factor between 0.8 and 1.2 leaves it unchanged for at least \rMemKeepTwenty\%.
An odometer that always reports motion reduces the rule to appending every frame, which is the setting measured above.
For actuation, the simulator executes turns and moves exactly except at obstacles, where \rCtlShort\% of the forward moves end short through sliding contact (Sec.~\ref{sec:suppl_cost}).

\section{Failure analysis}
\label{sec:suppl_failure}

This section analyzes the failures of the policy, most of which come from the stop decision.
The stop head ranks decisions with an AUROC of \rStopAUROC, and SR on unseen buildings stays between \rStopRepSRSix\% and \rStopRepSRNine\% for every threshold from 0.6 to 0.9.
Over three seeds, \rStopNeverFailStop\% of the episodes stop without any decision within 3\,m of the goal, and \rStopPassFail\% pass within 3\,m and fail (\rStopPassFailStop\% after a stop elsewhere, the others without a stop).
An oracle stop lifts seed 42 from \rFinSRA\% to \rXOrStopSR\% SR (Sec.~\ref{sec:exp_dagger}).

\begin{figure}[t]
  \centering
  \figinc{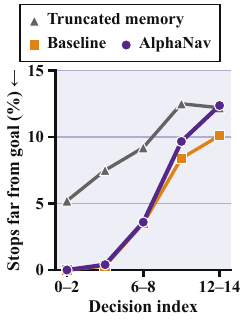}{0.5\columnwidth}
  \caption{Stops far from the goal by decision index on unseen buildings. The vertical axis is the share of the decisions at each index that stop more than 5\,m from the goal, and the \paper curve is the mean of three seeds.}
  \label{fig:stop}
\end{figure}

\paragraph{Stops far from the goal.}
Most failures of \paper are stops more than 5\,m from the goal (Sec.~\ref{sec:exp_dagger}).
Their rate grows with the decision index in every model of Fig.~\ref{fig:stop} and reaches \pgfmathparse{100*\rTFinCLate}\pgfmathprintnumber[fixed,fixed zerofill,precision=1]{\pgfmathresult}\% at decisions 9 to 11 for seed 44.

\paragraph{Stop decision.}\label{sec:suppl_stop}
The stop head receives the heatmap-pooled current-view state.
Training labels a decision as positive within a geodesic distance of 2\,m from the goal and uses the measured class weight $w=\rStopPosWeight$.
At inference, we correct the logit by $\ln w=\rStopOffset$ and compare the resulting probability with 0.9 (Sec.~\ref{sec:heads}).
Over the three seeds on unseen buildings, \rStopNone\% of the episodes do not stop at all, \rStopNeverFailStop\% stop without any decision within 3\,m of the goal, \rStopPassFailStop\% stop elsewhere after a decision within 3\,m, and \rStopFarEp\% stop more than 5\,m from the goal.
These far stops account for \rStopFarFail\% of the failures.
Of all stops, \rStopPrec\% lie within 3\,m and \rStopNear\% lie within 1.5\,m of the goal.

Table~\ref{tab:stopcal} reports the calibration of the stop probability, that is, how closely the predicted probability matches the observed rate of decisions near the goal.
Over the \rStopDecN\ policy decisions, the corrected probability separates decisions within 2\,m of the goal from the others with an AUROC of \rStopAUROC, or \rStopAUROCThree\ for 3\,m.
Its expected calibration error (ECE), which averages the gap between the predicted probability and the observed rate of positive labels over probability bins, is \rStopECE.
The probability averages \rStopMeanPNear\ within 2\,m and \rStopMeanPFar\ beyond 5\,m.
Above 0.5, the probability is overconfident.
Only \rStopBinFiveNear\% of the decisions between 0.5 and 0.6, \rStopBinEightNear\% of those between 0.8 and 0.9, and \rStopTopNear\% of the \rStopTopN\ decisions above 0.9 lie within 2\,m.
These \rStopTopN\ decisions are the stops, and their probability averages \rStopTopMeanP.

Table~\ref{tab:stoprep} replays the logged episodes at lower thresholds.
At a lower threshold, an episode stops at its first decision whose corrected stop probability exceeds the new value.
Up to this decision, the logged trajectory equals the trajectory of a run at that threshold, and the replay is therefore exact.
As a check, replaying the run of the inference batch with stop threshold 0.9 at 0.8 and 0.7 reproduces the measured runs at these thresholds in \rStopCheckEightDev\ and \rStopCheckSevenDev\ of their 300 episodes.
On unseen buildings, every lower threshold lowers SR, from \rStopRepSRNine\% at 0.9 to \rStopRepSREight\% at 0.8 and \rStopRepSRThree\% at 0.3.
Applying 0.9 to the uncorrected probability corresponds to a corrected threshold of \rStopNoCorrThr\ and yields \rStopRepSRNoCorr\% SR.
The logit correction therefore adds \rStopCorrGain\ points.
We fixed the threshold of 0.9 together with the memory spacing on 300 selection episodes, where it reaches \rMemEightSRDev\% against \rDevInfSpacingSR\% at 0.7 (Sec.~\ref{sec:suppl_seen}), before the unseen-building evaluation.
On all 778 selection episodes, a threshold of \rStopSelThrDev\ achieves the highest replayed SR (\rStopSelSRDev\% against \rStopRepSRNineDev\% at 0.9), and on unseen buildings this threshold yields \rStopSelUnseen\% (\rStopSelDelta\ points).

We also compute two offline bounds on the SR that a better choice of threshold can reach.
The best threshold per building, chosen on the same episodes, reaches \rStopBoundBuilding\% SR on unseen buildings.
For seed 42, this bound reaches \rStopBoundBuildingA\%, which lies \rStopBoundGainA\ points above the learned stop and covers \rStopBoundShare\% of the \rdOracleStopAbs-point gain of the oracle stop.
The best threshold per episode reaches \rStopBoundEpisode\%.
A lower threshold can only stop earlier.
However, \rStopNeverFailStop\% of the episodes already stop before any decision within 3\,m, and the oracle stop of Sec.~\ref{sec:exp_dagger} gains by letting these episodes continue.

Finally, we test a stop-focused on-policy recipe.
This recipe uses a focal stop loss with $\gamma=2$, which down-weights decisions that the head already classifies well, together with a larger policy share ($\alpha=0.85$ in Eq.~\ref{eq:schedule}) and uniform start headings in half of the rollouts.
It trains from the stage-1 checkpoint of seed 42 and uses a stop threshold of 0.4, selected on the 300 selection episodes.
The recipe halves the calibration error (\rStopVFiveECE\ against \rStopECEA\ for seed 42) and raises the AUROC from \rStopAUROCA\ to \rStopVFiveAUROC\ at a comparable SR.
In its single evaluation on unseen buildings, the recipe reaches \rXVFiveSR\% SR and \rXVFiveSPL\% SPL (\rdVFive\ [\rdVFiveLo, \rdVFiveHi] points against seed 42).
On all 778 selection episodes, it differs from the adopted recipe by \rdDvFiveDev\ [\rdDvFiveDevLo, \rdDvFiveDevHi] points (Table~\ref{tab:registered}).
The recipe stops beyond 5\,m in \rStopVFiveFarEp\% of its episodes, compared with \rStopSeedAFarEp\% for seed 42.
The better-calibrated probability therefore stops far from the goal equally often.

\begin{table}[t]
  \centering
  \footnotesize
  \setlength{\tabcolsep}{4pt}
  \begin{tabular}{@{}>{\tintL}lrccc<{\tintR}@{}}
    \toprule
    Stop probability & Decisions & Mean & Within 2\,m (\%) & Within 3\,m (\%) \\
    \midrule
    0--0.01 & \rCalaN & \rCalaP & \rCalaTwo & \rCalaThree \\
    0.01--0.05 & \rCalbN & \rCalbP & \rCalbTwo & \rCalbThree \\
    0.05--0.1 & \rCalcN & \rCalcP & \rCalcTwo & \rCalcThree \\
    0.1--0.2 & \rCaldN & \rCaldP & \rCaldTwo & \rCaldThree \\
    0.2--0.3 & \rCaleN & \rCaleP & \rCaleTwo & \rCaleThree \\
    0.3--0.4 & \rCalfN & \rCalfP & \rCalfTwo & \rCalfThree \\
    0.4--0.5 & \rCalgN & \rCalgP & \rCalgTwo & \rCalgThree \\
    0.5--0.6 & \rCalhN & \rCalhP & \rCalhTwo & \rCalhThree \\
    0.6--0.7 & \rCaliN & \rCaliP & \rCaliTwo & \rCaliThree \\
    0.7--0.8 & \rCaljN & \rCaljP & \rCaljTwo & \rCaljThree \\
    0.8--0.9 & \rCalkN & \rCalkP & \rCalkTwo & \rCalkThree \\
    0.9--1 (stops) & \rCallN & \rCallP & \rCallTwo & \rCallThree \\
    \bottomrule
  \end{tabular}
  \caption{Reliability of the corrected stop probability on unseen buildings, over all decisions of three seeds (geodesic distances).}
  \label{tab:stopcal}
\end{table}

\begin{table}[t]
  \centering
  \footnotesize
  \setlength{\tabcolsep}{3.5pt}
  \begin{tabular}{@{}>{\tintL}lccccc<{\tintR}@{}}
    \toprule
    & \multicolumn{4}{c}{Unseen buildings (3 seeds)} & Seen \\
    \cmidrule(lr){2-5}\cmidrule(l){6-6}
    Stop threshold & SR\,$\uparrow$ & SPL\,$\uparrow$ & NE\,$\downarrow$ & Decisions & SR\,$\uparrow$ \\
    \midrule
    \oursrow \oursname{0.9 (\paper)} & \best{\rStopRepSRNine} & \rStopRepSPLNine & \rStopRepNENine & \rStopRepDecNine & \rStopRepSRNineDev \\
    0.8 & \rStopRepSREight & \best{\rStopRepSPLEight} & \rStopRepNEEight & \rStopRepDecEight & \rStopRepSREightDev \\
    0.7 & \rStopRepSRSeven & \rStopRepSPLSeven & \best{\rStopRepNESeven} & \rStopRepDecSeven & \best{\rStopRepSRSevenDev} \\
    \rStopNoCorrThr\ (0.9 uncorrected) & \rStopRepSRNoCorr & \rStopRepSPLNoCorr & \best{\rStopRepNENoCorr} & \rStopRepDecNoCorr & \rStopRepSRNoCorrDev \\
    0.6 & \rStopRepSRSix & \rStopRepSPLSix & \rStopRepNESix & \rStopRepDecSix & \rStopRepSRSixDev \\
    0.5 & \rStopRepSRFive & \rStopRepSPLFive & \rStopRepNEFive & \rStopRepDecFive & \rStopRepSRFiveDev \\
    0.4 & \rStopRepSRFour & \rStopRepSPLFour & \rStopRepNEFour & \rStopRepDecFour & \rStopRepSRFourDev \\
    0.3 & \rStopRepSRThree & \rStopRepSPLThree & \rStopRepNEThree & \rStopRepDecThree & \rStopRepSRThreeDev \\
    \midrule
    Best per building$^\dagger$ & \rStopBoundBuilding & -- & -- & -- & \rStopBoundBuildingDev \\
    Best per episode$^\dagger$ & \rStopBoundEpisode & -- & -- & -- & -- \\
    \midrule
    0.7, every frame$^{*}$ & \rStopRepSRSevenEvery & \rStopRepSPLSevenEvery & \rStopRepNESevenEvery & -- & -- \\
    \bottomrule
  \end{tabular}
  \caption{Exact replay at lower stop thresholds. Seen: 778 selection episodes, seed 42. $^\dagger$Offline bounds chosen on the same episodes. $^{*}$Measured runs without memory spacing (Table~\ref{tab:seeds_onpolicy}); with the 0.7 row, they give the memory-spacing gain. Bold: best threshold per column (bounds and $^{*}$ excluded).}
  \label{tab:stoprep}
\end{table}

\paragraph{Memory frames at inference.}
Because training always uses eight memory frames, keeping fewer frames at inference measures how strongly the stop decision depends on its trained input sequence.
On 300 selection episodes with seed 42 of \paper, success falls from \rMemEightSRDev\% with all eight frames to \rMemFourSRDev\% with four frames and \rMemZeroSRDev\% with none, and without memory frames no episode stops (Table~\ref{tab:memory}).

\paragraph{Truncated-memory training.}\label{sec:suppl_truncation}
With one seed per model and a stop threshold of 0.6, truncating the memory during training lowers SR by \rdTruncAbs\ points and raises far stops from \rTAZeroFFar\% to \rTAFourFFar\% of the episodes (Table~\ref{tab:truncation}).
This result suggests that the far stops depend on more than elapsed time.

\paragraph{Visited-place memory.}
We also test a training-free memory that replaces a planned move returning within 1\,m of a place visited at least 3\,m of travel earlier.
This memory changes success on seen buildings by \rdLoopMemDev\ [\rdLoopMemDevLo, \rdLoopMemDevHi] points (Table~\ref{tab:registered}).

\begin{table}[t]
  \centering
  \footnotesize
  \setlength{\tabcolsep}{3pt}
  \begin{tabular}{@{}>{\tintL}lccc<{\tintR}@{}}
    \toprule
    Memory frames at inference & SR\,$\uparrow$ & SPL\,$\uparrow$ & $\Delta$SR [95\% interval] \\
    \midrule
    \oursrow \oursname{All 8 frames (\paper)} & \best{\rMemEightSRDev}& \best{\rMemEightSPLDev}& -- \\
    4 of 8 frames & \rMemFourSRDev\  & \rMemFourSPLDev\  & \rdMemFourDev \\
    No memory frame & \rMemZeroSRDev\  & \rMemZeroSPLDev\  & \rdMemKDev\ [\rdMemKDevLo, \rdMemKDevHi] \\
    \bottomrule
  \end{tabular}
  \caption{Memory frames kept at inference on seen buildings. $\Delta$SR: change against all eight frames.}
  \label{tab:memory}
\end{table}

\begin{table}[t]
  \centering
  \footnotesize
  \setlength{\tabcolsep}{3.5pt}
  \begin{tabular}{@{}lcccc@{}}
    \toprule
    Model & NE\,$\downarrow$ & SR\,$\uparrow$ & SPL\,$\uparrow$ & Far stops (\%) \\
    \midrule
    Baseline recipe & \rAZeroFNE\ & \rAZeroFSR\ & \rAZeroFSPL\ & \rTAZeroFFar \\
    + truncated-memory training & \rAFourFNE\ & \rAFourFSR\ & \rAFourFSPL\ & \rTAFourFFar \\
    \bottomrule
  \end{tabular}
  \caption{Truncated-memory training on unseen buildings. Far stops: stops beyond 5\,m from the goal.}
  \label{tab:truncation}
\end{table}

\section{Additional controls}
\label{sec:suppl_controls}

Three further runs on all 1{,}839 unseen-building episodes with seed 42 test alternative explanations of the main results (Table~\ref{tab:controls}).
We ran them after the main experiments, and none of them selects a setting.

\paragraph{Readout without on-policy rounds.}
The on-policy rounds collect their states with the route-bearing controller, so the scan readout of Table~\ref{tab:ablation}b decodes weights that were corrected under the bearing readout.
Decoding the imitation-only weights of the same seed, which never ran the bearing controller, removes this effect.
At the settings of Table~\ref{tab:ablation}a (stop threshold 0.7, every frame appended), the route bearing reaches \rDStageOneSR\% SR and the scan \rRoneSR\%, a gain of \rdRevReadStageOne\ [\rdRevReadStageOneLo, \rdRevReadStageOneHi] points, compared with \rdReadFinScan\ points on the final weights.
Most of the readout gain therefore holds without on-policy correction.

\paragraph{Equal training steps.}
The three on-policy rounds add 3{,}600 fine-tuning steps to stage 1.
An imitation control repeats the same pipeline from the same stage-1 checkpoint, with the same rounds, rollouts, steps, learning rate, and data mix, but the expert drives every rollout step, so the rounds add only reference-route states.
At the settings of Table~\ref{tab:ablation}a, it reaches \rRthreeSR\% SR, against \rDStageOneSR\% after imitation and \rDVTwoSR\% after on-policy correction.
The extra steps alone change SR by \rdRevStepsCtl\ [\rdRevStepsCtlLo, \rdRevStepsCtlHi] points, and on-policy correction gains \rdRevOnPolicyCtl\ [\rdRevOnPolicyCtlLo, \rdRevOnPolicyCtlHi] points over the control, so the gain of Sec.~\ref{sec:exp_dagger} comes from the visited states and not from the added steps.

\paragraph{Route and stop errors.}
The third run replaces the route outputs of \paper (waypoint, visibility, and bearing) by the oracle outputs of the controller ceiling at every decision and keeps the learned stop.
The oracle route follows the shortest path from the live pose to the goal, not the instructed route.
Together with the controller ceiling (oracle route and stop) and the oracle stop of Sec.~\ref{sec:exp_dagger}, it completes the split of Table~\ref{tab:controls}b at the final settings.
The oracle route raises SR by \rdRevRouteLearnedStop\ [\rdRevRouteLearnedStopLo, \rdRevRouteLearnedStopHi] points with the learned stop and by \rdRevRouteOracleStop\ [\rdRevRouteOracleStopLo, \rdRevRouteOracleStopHi] points with the oracle stop.
The oracle stop raises SR by \rdOracleStop\ [\rdOracleStopLo, \rdOracleStopHi] points with the learned route and by \rdRevStopOracleRoute\ [\rdRevStopOracleRouteLo, \rdRevStopOracleRouteHi] points with the oracle route.
Route errors therefore cost more success than stop errors, although the stop still accounts for \rStopFailShare\% of the failures of the learned policy.

\begin{table}[t]
  \centering
  \footnotesize
  \setlength{\tabcolsep}{3.5pt}
  \begin{tabular}{@{}>{\tintL}lcccc<{\tintR}@{}}
    \toprule
    \multicolumn{5}{@{}l}{(a) Controls, stop threshold 0.7, every frame appended} \\
    \midrule
    Run & SR\,$\uparrow$ & OSR\,$\uparrow$ & SPL\,$\uparrow$ & NE\,$\downarrow$ \\
    \midrule
    \groupband{5}{Imitation-only weights, two readouts}
    Route bearing & \rDStageOneSR & \rDStageOneOSR & \rDStageOneSPL & \rDStageOneNE \\
    Image-plane scan & \rRoneSR & \rRoneOSR & \rRoneSPL & \rRoneNE \\
    \groupband{5}{Equal training steps, route bearing}
    Imitation (stage 1) & \rDStageOneSR & \rDStageOneOSR & \rDStageOneSPL & \rDStageOneNE \\
    + 3{,}600 expert-driven steps & \rRthreeSR & \rRthreeOSR & \rRthreeSPL & \rRthreeNE \\
    + on-policy correction & \rDVTwoSR & \rDVTwoOSR & \rDVTwoSPL & \rDVTwoNE \\
    \midrule
    \multicolumn{5}{@{}l}{(b) SR of route and stop, final settings} \\
    \midrule
    & \multicolumn{2}{c}{Learned stop} & \multicolumn{2}{c}{Oracle stop} \\
    \midrule
    Learned route & \multicolumn{2}{c}{\rFinSRA} & \multicolumn{2}{c}{\rXOrStopSR} \\
    Oracle route & \multicolumn{2}{c}{\rRtwoSR} & \multicolumn{2}{c}{\rXCeilSR} \\
    \bottomrule
  \end{tabular}
  \caption{Additional controls on unseen buildings, seed 42, in percent except NE (meters). Oracle route: the controller ceiling's route outputs along the live shortest path to the goal; oracle stop: stop within 1.5\,m of the goal.}
  \label{tab:controls}
\end{table}

\section{Evaluation and measurement protocol}
\label{sec:suppl_protocol}

This section states how we evaluate the models, select the inference settings, measure latency and memory, and report the controlled comparisons.

\paragraph{Evaluation.}\label{sec:suppl_eval}
We evaluate every model once on all 1{,}839 unseen-building episodes with the settings of Sec.~\ref{sec:experiments}.
The agent has a radius of 0.1\,m and one $256\times256$ RGB camera with a $90^\circ$ horizontal field of view at a height of 1.25\,m and a pitch of $0^\circ$.
It acts with 0.25\,m forward moves, $15^\circ$ turns, and stop, slides along obstacles, and ends the remaining moves of a decision after a collision, and an episode ends after 500 primitives.
SR, OSR, SPL, and NE follow the VLN-CE definitions with geodesic distances and a success radius of 3\,m.
Unless noted otherwise, intervals are 95\% bootstrap intervals that resample whole buildings, differences between models are paired over episodes, and all differences and means are computed before rounding.

\paragraph{Selection of the inference settings.}\label{sec:suppl_seen}
We choose the stop threshold and the memory spacing on the selection set.
Each setting gains at least 1.0 SR point on 300 selection episodes.
The combination of a stop threshold of 0.9 and memory frames every 0.5\,m gains at least 1.5 points on the 478 other selection episodes.
The base rule of seed 42 after the on-policy rounds uses stop threshold 0.7, appends every frame, and reaches \rDevInfBaseSR\% SR.
From this base, a stop threshold of 0.8 yields \rDevInfStopEightSR\%, a stop threshold of 0.9 yields \rDevInfStopNineSR\%, and the memory spacing yields \rDevInfSpacingSR\%.

\paragraph{Latency and memory.}\label{sec:suppl_measure}
We measure latency on one 12\,GB RTX~5070 in bfloat16 at batch size 1.
The timed span covers frame preprocessing, assembly of the eight-frame memory, and one forward pass, and excludes the controller and the simulator.
Eight decisions fill the memory, and five decisions warm up the GPU.
We then report the median of 50 decisions with Compute Unified Device Architecture (CUDA) synchronization, on one fixed random $256\times256$ frame with an instruction of median length.
Each decision consists of one forward pass over a fixed number of tokens, and the heads take its last-layer states as input.
We report memory as the peak memory in GiB that PyTorch allocates during the timed decisions, including the weights.
The closed-loop time of Sec.~\ref{sec:suppl_cost} is the wall time of the three evaluation runs, including the controller, simulator, rendering, and logging.
We sum this time over the episodes that stay in the building of the previous episode and divide it by their decisions.

\paragraph{Measured effects.}\label{sec:suppl_registered}
Table~\ref{tab:registered} lists every controlled comparison of the paper with its measured effect and 95\% interval.
The on-policy schedule gains \rdDoseOne\ points over imitation and \rdDoseV\ points over the round-level schedule.
The route bearing gains \rdReadFinScan\ points over the scan, and the orient-then-go augmentation gains \rdOrient\ points on starts that face away from the route.
Two further on-policy rounds change success by \rdDvThreeDev\ points, and the schedule therefore ends after three rounds.

\begin{table*}[t]
  \centering
  \footnotesize
  \setlength{\tabcolsep}{3pt}
  \renewcommand{\arraystretch}{1.15}
  \begin{tabular}{@{}>{\tintL}ll>{\raggedright\arraybackslash}p{7.4cm}cc<{\tintR}@{}}
    \toprule
    Claim & Test & Comparison & Buildings & Effect [95\% interval] \\
    \midrule
    Accuracy & Mean SR & mean SR over three seeds & unseen & \rFinSR \\
    \stripe On-policy & Gain over imitation & within-episode schedule $-$ imitation & unseen & \rdDoseOne\ [\rdDoseOneLo, \rdDoseOneHi] \\
    On-policy & Gain over round-level & within-episode $-$ round-level schedule & unseen & \rdDoseV\ [\rdDoseVLo, \rdDoseVHi] \\
    \stripe On-policy & Equal inference settings & within-episode schedule $-$ imitation, both at stop 0.9 with memory spacing, 300 episodes, seed 42 & seen & \rdAblDDev\ [\rdAblDDevLo, \rdAblDDevHi] \\
    On-policy & Schedule selection & within-episode $-$ round-level, 478 other episodes & seen & \rdDvTwoDev\ [\rdDvTwoDevLo, \rdDvTwoDevHi] \\
    \stripe On-policy & Further rounds & two more rounds $-$ three rounds & seen & \rdDvThreeDev\ [\rdDvThreeDevLo, \rdDvThreeDevHi] \\
    On-policy & Stop-focused recipe & focal stop loss, $\alpha=0.85$, uniform starts $-$ adopted recipe, 778 episodes & seen & \rdDvFiveDev\ [\rdDvFiveDevLo, \rdDvFiveDevHi] \\
    \stripe On-policy & Oracle stop & oracle stop within 1.5\,m $-$ learned stop, \paper weights & unseen & \rdOracleStop\ [\rdOracleStopLo, \rdOracleStopHi] \\
    Output & Readout & route bearing $-$ image-plane waypoint with scan, \paper weights & unseen & \rdReadFinScan\ [\rdReadFinScanLo, \rdReadFinScanHi] \\
    \stripe Output & Readout & route bearing $-$ image-plane waypoint with displacement head, \paper weights & unseen & \rdReadFin\ [\rdReadFinLo, \rdReadFinHi] \\
    Output & Readout, \rxr & route bearing $-$ image-plane waypoint with scan, \paper weights, seed 42, \rxr English & unseen & \rdRxScan\ [\rdRxScanLo, \rdRxScanHi] \\
    \stripe Output & First turn & starts beyond $90^\circ$ whose first turn brings the route within $90^\circ$, seeds 42 / 43 / 44 (\%) & unseen & \rHFinARot\ / \rHFinBRot\ / \rHFinCRot \\
    Augmentation & Orient-then-go & orient-then-go $-$ baseline, starts that face away & unseen & \rdOrient\ [\rdOrientLo, \rdOrientHi] \\
    \stripe Augmentation & Orient-then-go & orient-then-go $-$ baseline, other starts & unseen & \rdOrientGuard\ [\rdOrientGuardLo, \rdOrientGuardHi] \\
    Augmentation & Direction words & instruction sensitivity, baseline / orient-then-go & unseen & \rSAZeroF\ / \rSAOneF \\
    \stripe Failures & Far-stop rise & far-stop rate at decisions 9--11 minus 0--2, seed 42 (\%) & unseen & \pgfmathparse{100*\rTFinARise}\pgfmathprintnumber[fixed,fixed zerofill,precision=1,showpos]{\pgfmathresult} \\
    Failures & Truncation & truncated memory $-$ baseline, SR & unseen & \rdTrunc\ [\rdTruncLo, \rdTruncHi] \\
    \stripe Failures & Memory frames & no memory frame $-$ all eight & seen & \rdMemKDev\ [\rdMemKDevLo, \rdMemKDevHi] \\
    Failures & Visited places & visited-place memory $-$ \paper, seed 42 & seen & \rdLoopMemDev\ [\rdLoopMemDevLo, \rdLoopMemDevHi] \\
    \stripe Camera & Unrestricted & unrestricted $-$ two-view imitation, pooled SR & seen & \rdCamPreDev\ [\rdCamPreDevLo, \rdCamPreDevHi] \\
    Camera & Common support & common support $-$ unrestricted, pooled SR & seen & \rdCamMethodDev\ [\rdCamMethodDevLo, \rdCamMethodDevHi] \\
    \stripe Camera & Two-view imitation & two-view imitation $-$ no pairs, pooled SR & seen & \rdCamAltDev\ [\rdCamAltDevLo, \rdCamAltDevHi] \\
    Camera & Field of view & error at $60^\circ$ $-$ error at $75^\circ$--$105^\circ$ & seen & \rdHfovSixtyDev\ [\rdHfovSixtyDevLo, \rdHfovSixtyDevHi] \\
    \stripe Camera & Field of view & error at $120^\circ$ $-$ error at $75^\circ$--$105^\circ$ & seen & \rdHfovTwentyDev\ [\rdHfovTwentyDevLo, \rdHfovTwentyDevHi] \\
    \bottomrule
  \end{tabular}
  \caption{Measured effects of the controlled comparisons, in SR points except field of view (degrees), first turn and far-stop rise (percent), and sensitivity (probability). Pooled SR: mean SR over the three held-out cameras.}
  \label{tab:registered}
\end{table*}

\section{Implementation details}
\label{sec:suppl_impl}

This section describes the configuration, inputs, supervision, training procedure, training data, and compute of the evaluated policy.

\paragraph{Configuration.}\label{sec:suppl_config}
Table~\ref{tab:hparams} lists the configuration of the evaluated system as specified in the configuration files of the final recipe.

\paragraph{Inputs and supervision.}\label{sec:suppl_inputs}
At deployment, the policy receives the instruction, the current $256\times256$ RGB frame, the eight memory frames, and the camera vector $\kap$.
This vector contains the logarithms of the focal lengths and the principal point relative to the image size, the camera height and pitch, and the robot width and turning radius.
The memory rule uses the displacement since the last appended frame, which a robot obtains from odometry.
During training, the memory holds the eight previous decision frames, without the spacing rule.
When fewer than eight frames are available, copies of the oldest frame fill the remaining slots, so the sequence length stays fixed.
The controller uses only the policy outputs, and rendered depth serves only to build the training labels.

The training logs show which terms supervise the outputs.
At every logged step of stage 1 and of the on-policy rounds, they contain five weighted terms, namely the waypoint-heatmap Kullback-Leibler (KL) divergence (1.0), the pixel L1 loss (0.5), the route-bearing cross-entropy (1.0), the visibility loss (0.25), and the stop loss (0.5).
The logged total equals their weighted sum, and the displacement and absolute-heading heads have weight zero.
The heatmap and pixel terms therefore supervise the waypoint to which the image-plane readouts of Tables~\ref{tab:ablation}b and~\ref{tab:heading} steer, and the route-bearing term supervises the bearing readout.

\paragraph{Reference-route supervision.}\label{sec:suppl_training}
We build the supervision targets from reference routes, which connect the annotated waypoints through shortest navigable paths on the navigation mesh of each training house.
At each step, we select the farthest route point within 6\,m if it lies along a nearly straight path and is visible and unoccluded in rendered depth.
If no such point exists, we use the next route point.
Visible targets receive heatmap and pixel labels, and targets at least 0.3\,m away receive bearing labels $\theta^\star_t$.

\paragraph{Training objective.}
Each term of the objective (Eq.~\ref{eq:loss}) is averaged over the frames that it supervises.
The waypoint heatmap $\bm{h}^\star_t$ is a Gaussian centered on the grid position of the projected target.
The bearing target $\bm{y}_t$ is a circular Gaussian with a standard deviation of one bin, centered on the bin of the target bearing.
The target-pixel losses apply only to targets that are visible in the image, and the bearing loss applies only to valid directions.
A stop label is positive within a geodesic distance of 2\,m of the goal, and the class weight $w$ equals the ratio of negative to positive stop labels.

\paragraph{Label-preserving augmentations.}\label{sec:suppl_aug}
Two augmentations transform the inputs together with their labels.
The mirror augmentation reflects the current and memory frames horizontally and maps the pixel column $u$ to $W-u$, where $W$ is the image width.
It exchanges the directional left and right words of the instruction and keeps the compass terms.
It transforms the bearing angle and bin according to Eq.~\ref{eq:mirror}.
\begin{equation}
  \theta_k \mapsto -\theta_k, \qquad k \mapsto (\nbins-k) \bmod \nbins.
  \label{eq:mirror}
\end{equation}
Two exchanges restore all but one of the 50{,}496 instructions that contain directional words.
The labels stay exact when the principal point is horizontally centered and the instruction keeps its meaning (Sec.~\ref{sec:suppl_twins}).
The orient-then-go augmentation samples the start heading uniformly.
It prefixes the instruction with ``Turn left.'', ``Turn right.'', or ``Turn around.'', which guides the agent back to the heading that the original instruction assumes.
The clause is omitted for rotations below $45^\circ$.
Half of the examples are mirrored, and all transformations are applied in place.
We ablate orient-then-go (Table~\ref{tab:ablation}c) but not mirroring, which every model except the baseline recipe uses.

\paragraph{Initial imitation training.}
Stage 1, the initial imitation training, trains the policy with AdamW for 10{,}000 steps.
It samples \rtor training episodes, English \rxr training episodes, and 30{,}000 EnvDrop episodes with weights 0.5, 0.3, and 0.2, respectively.
All episodes are restricted to training houses~\cite{tan2019envdrop}.
Stage 1 applies photometric jitter and randomizes the camera parameters in 25\% of episodes, with fields of view between $80^\circ$ and $100^\circ$ (Table~\ref{tab:hparams}).
We select the checkpoint by a closed-loop evaluation on 100 \valseen episodes every 2{,}500 steps.

\paragraph{On-policy correction.}\label{sec:suppl_dagger}
On-policy correction extends the supervision to states that the policy reaches after its own errors.
We run the current policy on \rtor training episodes and label the visited states with the reference-route targets defined above.
Following DAgger (dataset aggregation)~\cite{ross2011dagger}, we aggregate these corrections with the original data and sample half of each batch from them.
The probability that the policy controls decision $i$ follows the Efficient-VLN schedule~\cite{zheng2025efficientvln}:
\begin{equation}
  P_{\mathrm{policy}}(i) = \alpha^{\,p_i}, \qquad p_i=\min\!\big(1,\ i/(L-1)\big), \quad \alpha=0.6.
  \label{eq:schedule}
\end{equation}
Here, $L$ denotes the expected number of decisions.
The probability of policy control is therefore close to 1 early in the episode and falls to 0.6 at its expected end.
If the agent is more than 1\,m from the reference path or is blocked by an obstacle after a policy decision, the oracle takes control for five decisions.
Control then returns to the policy.
With these takeovers, the policy drives \rPolicyShareVTwo\% of the rollout steps, which is the on-policy share of this schedule.
The ablation of Table~\ref{tab:ablation}a compares it with a round-level schedule, in which the oracle acts with probability $0.5^{r+1}$ at the start of round $r$, a probability that then rises toward the end of the episode.
This schedule collects 1{,}000 rollouts per round, uses the same takeovers, and lets the policy drive \rPolicyShareVOne\% of the rollout steps.
We run three correction rounds, each with 1{,}500 rollouts followed by 1{,}200 fine-tuning steps at a learning rate of $5\times10^{-5}$.
Both augmentations stay active during correction.

\begin{table}[t]
  \centering
  \footnotesize
  \setlength{\tabcolsep}{3pt}
  \begin{tabular}{@{}ll@{}}
    \toprule
    Item & Value \\
    \midrule
    \groupband{2}{Model}
    Backbone & Qwen3-VL-2B-Instruct, base weights frozen \\
    LoRA & rank 32, scaling 64, bfloat16 \\
    Visual tokens & $8{\times}16$ memory (128\,px), 64 current (256\,px) \\
    Camera vector $\kap$ & eight entries; FiLM hidden size 128 \\
    Head hidden size & 512; $\nbins=24$ bearing bins \\
    \groupband{2}{Training}
    Stage-1 steps & 10{,}000, batch 32 ($2\times16$ accumulation) \\
    Optimizer & AdamW $(0.9,0.95)$, decay 0.01, clip 1.0 \\
    Learning rate & $10^{-4}$, 500 warm-up steps, cosine to $10\%$ \\
    Loss weights & 1.0 heatmap, 0.5 pixel, 1.0 bearing, \\
    & 0.25 visibility, 0.5 stop (measured weight), \\
    & 0 displacement head (unsupervised) \\
    Bearing target & Gaussian, $\sigma=1$ bin \\
    Data draw weights & 0.5 \rtor, 0.3 \rxr English, 0.2 EnvDrop \\
    Camera randomization & 25\% of episodes; hfov $80^\circ$--$100^\circ$, \\
    & height 1.0--1.5\,m, pitch $-10^\circ$--$0^\circ$ \\
    Mirror, rotation clause & 0.5 of items; 20\% of uniform starts \\
    On-policy rounds & 3 rounds of 1{,}500 rollouts, \\
    & 1{,}200 steps each, learning rate $5\times10^{-5}$ \\
    Schedule & $P_{\mathrm{policy}}=0.6^{p}$, state-level oracle labels \\
    Probe & 100 selection episodes every 2{,}500 steps \\
    \groupband{2}{Inference}
    Controller & stop $>0.9$ (corrected); visibility $>0.5$; \\
    & cone $40^\circ$; 4 commit moves; 2 hidden moves \\
    & within $7.5^\circ$; turn-loop guard after 4 turns \\
    Memory & 8 frames, 0.5\,m spacing at inference \\
    \bottomrule
  \end{tabular}
  \caption{Configuration of the evaluated system. hfov: horizontal field of view.}
  \label{tab:hparams}
\end{table}

\paragraph{Training data.}\label{sec:suppl_data}
Stage 1 draws on 10{,}352 \rtor episodes with 162{,}493 records, 18{,}490 English \rxr episodes with 421{,}164 records, and 28{,}671 EnvDrop episodes with 502{,}544 records.
In total, it uses 57{,}513 episodes and 1{,}086{,}201 records.
Before the expert follower generates the trajectories, we remove the \rxr and EnvDrop episodes whose start and goal both lie within 0.5\,m of those of an evaluation or selection route.
This filter removes 15 and 32 episodes, respectively.
The expert follower completes the episodes counted above within its decision limit, out of 10{,}819 \rtor, 19{,}981 \rxr, and 30{,}000 sampled EnvDrop episodes.
We use the \rtor training split as released, and its six episodes on two routes within 0.5\,m of a selection route lie outside the generated set.
The baseline recipe of Table~\ref{tab:ablation}c trains on the \rtor part alone for 4{,}000 steps without the two augmentations.

\paragraph{Training compute.}
On one RTX~5070, an uninterrupted training run of stage 1 and three on-policy rounds takes \rCostTrainTyp\ GPU-hours on average (Table~\ref{tab:perseed}), at a peak memory of \rTrainGB\,GiB.
Rendering the labels takes \rCostGen\ GPU-hours once, which brings the total to \rCostAllTyp\ GPU-hours.

\paragraph{Data, models, and licenses.}\label{sec:suppl_licences}
{\footnotesize We use the Matterport3D scenes under the Matterport \href{https://kaldir.vc.cit.tum.de/matterport/MP_TOS.pdf}{terms of use} for non-commercial academic purposes, and these terms also cover our figure renderings and our LoRA adapters, which the terms classify as derived information.
\par\smallskip
{\scriptsize\setlength{\tabcolsep}{2pt}\renewcommand{\arraystretch}{0.95}%
\begin{tabular}{@{}>{\raggedright\arraybackslash}p{0.9in}>{\raggedright\arraybackslash}p{0.5in}>{\raggedright\arraybackslash}p{1.3in}>{\raggedright\arraybackslash}p{0.38in}@{}}
  \toprule
  Resource & Used for & License or terms & Cite \\
  \midrule
  Matterport3D & scenes & Matterport terms: non-commercial academic use & \cite{chang2017matterport} \\
  \rtor & train, eval & Matterport terms; CC BY-NC-SA 3.0 US & \cite{krantz2020vlnce,anderson2018vln} \\
  \rxr English & training & RxR: CC BY 4.0; port as \rtor & \cite{ku2020rxr,krantz2020vlnce} \\
  EnvDrop & training & no data license stated; port as \rtor & \cite{tan2019envdrop} \\
  Habitat-Sim & simulator & MIT & \cite{savva2019habitat} \\
  Qwen3-VL-2B-Instruct, PEFT & backbone, LoRA & Apache-2.0 & \cite{bai2025qwen3vl,hu2022lora,mangrulkar2022peft} \\
  \bottomrule
\end{tabular}\par}
\smallskip
\noindent We use the episodes as released in the VLN-CE data after the sampling above and use Qwen3-VL-2B-Instruct under its license.
We do not redistribute any scene or collect any personal data, and we quote the numbers of other systems from their papers.\par}